\documentclass[11pt, dvipsnames, DIV=11, sfdefaults=false,headings=standardclasses]{scrartcl} 
\usepackage{hyperref}
\usepackage{url}

\usepackage[english]{babel}
 
\usepackage[utf8x]{inputenc}
\usepackage[T1]{fontenc}
\usepackage{natbib}
\usepackage{subcaption}
\usepackage{booktabs}
\usepackage{multirow}
\usepackage{url}
\usepackage{soul}
\usepackage{tikz,}
\usepackage{xcolor}
\usetikzlibrary{calc,quotes,decorations.pathreplacing,decorations.text}

\usepackage[scaled=0.86]{helvet}
\usepackage{newtxtext}
\usepackage{newtxmath}

\usepackage{paralist}
\usepackage[stable]{footmisc}
\usepackage{adjustbox}

\usepackage{pifont}
\newcommand{\cmark}{\ding{51}}

\usepackage{ifthen}
\newcommand{\CC}[1][]{$\text{C\hspace{-.25ex}}^{_{_{_{++}}}}
\ifthenelse{\equal{#1}{}}{}{\text{\hspace{-.625ex}#1}}$}

\usepackage{bm}
\usepackage{bbm}
\usepackage{amsmath}
\usepackage{thmtools}	
\usepackage[euro]{isonums}
\usepackage{nicefrac}
\usepackage{dsfont}
\usepackage{makecell}

\usepackage{booktabs}
\usepackage{multirow}
\usepackage{siunitx}
\usepackage{caption}
\usepackage{xspace}
\usepackage{array}
\usepackage{natbib}    

\newcolumntype{L}[1]{>{\raggedright\arraybackslash}p{#1}} 
\newcolumntype{C}[1]{>{\centering\arraybackslash}p{#1}}   
\newcolumntype{R}[1]{>{\raggedleft\arraybackslash}p{#1}}  
\newcolumntype{d}{S}                                      

\newcommand{\tabledefaults}{%
  \small
  \setlength{\tabcolsep}{6pt}%
  \renewcommand{\arraystretch}{1.15}%
}

\newcommand{\dataset}[1]{\textsc{#1}}
\newcommand{\solver}[1]{\textsc{#1}}

\usepackage[mathic=true]{mathtools}
\usepackage{fixmath}
\usepackage{siunitx}
\usepackage{dirtytalk}

\usepackage{mleftright}
\usepackage{stmaryrd}
\usepackage{xfrac}
\usepackage{algorithm}
\usepackage{algorithmicx}
\usepackage[noend]{algpseudocode}

\let\originalleft\mleft
\let\originalright\mright
\renewcommand{\mleft}{\mathopen{}\mathclose\bgroup\originalleft}
\renewcommand{\mright}{\aftergroup\egroup\originalright}

\usepackage{pifont}

\usepackage{enumitem}
\usepackage{listings}
\usepackage{graphicx}

\definecolor{codegreen}{rgb}{0,0.6,0}
\definecolor{codegray}{rgb}{0.5,0.5,0.5}
\definecolor{codepurple}{rgb}{0.58,0,0.82}
\definecolor{backcolour}{rgb}{0.95,0.95,0.92}

\lstdefinestyle{mystyle}{
    backgroundcolor=\color{backcolour},   
    commentstyle=\color{codegreen},
    keywordstyle=\color{magenta},
    numberstyle=\tiny\color{codegray},
    stringstyle=\color{codepurple},
    basicstyle=\ttfamily\footnotesize,
    breakatwhitespace=false,         
    breaklines=true,                 
    captionpos=b,                    
    keepspaces=true,                 
    numbers=left,                    
    numbersep=5pt,                  
    showspaces=false,                
    showstringspaces=false,
    showtabs=false,                  
    tabsize=2
}

\setlist[enumerate]{itemsep=0.2ex, topsep=0.5\topsep}
\setlist[description]{itemsep=0.2ex, topsep=0.5\topsep}
\setlist[itemize]{itemsep=0.2ex, topsep=0.5\topsep}

\makeatletter
\def\thmt@refnamewithcomma #1#2#3,#4,#5\@nil{%
\@xa\def\csname\thmt@envname #1utorefname\endcsname{#3}%
\ifcsname #2refname\endcsname
\csname #2refname\expandafter\endcsname\expandafter{\thmt@envname}{#3}{#4}%
\fi
}
\makeatother

\usepackage[capitalise,noabbrev]{cleveref}   

\usepackage[protrusion=true,expansion=false, activate={true,nocompatibility},final,kerning=true,spacing=true]{microtype}
\microtypecontext{spacing=nonfrench}
\usepackage{ellipsis}

\usepackage[auth-lg]{authblk}

\usepackage{todonotes}
\usepackage{anyfontsize}
\usepackage{yfonts}

\title{\huge \normalfont\textbf{GraphBench: Next-generation graph learning benchmarking}}

\author[1]{Timo Stoll\footnote{\hspace{5pt}First author, order decided randomly.}}
\author[1]{Chendi Qian$^{*}$}
\author[2]{Ben Finkelshtein$^{*}$}
\author[3]{Ali Parviz$^{*}$}
\author[1]{Darius Weber$^{*}$}
\author[4]{Fabrizio Frasca$^{*}$}
\author[1]{Hadar Shavit$^{*}$}
\author[1]{Antoine Siraudin$^{*}$}
\author[5,6]{Arman Mielke$^{*}$}
\author[1]{Erik Müller$^{*}$}
\author[1]{Marie Anastacio}
\author[7]{Maya Bechler-Speicher}
\author[2,8]{Michael Bronstein}
\author[9]{Mikhail Galkin}
\author[1]{Holger Hoos}
\author[6]{Mathias Niepert}
\author[9]{Bryan Perozzi}
\author[10]{Jan Tönshoff}
\author[1]{Christopher Morris\footnote{\hspace{5pt}Main contributing senior author.}}

\affil[1]{RWTH Aachen University}
\affil[2]{University of Oxford}
\affil[3]{Mila – Quebec AI Institute}
\affil[4]{Technion - Israel Institute of Technology}
\affil[5]{ETAS Research}
\affil[6]{University of Stuttgart}
\affil[7]{Meta}
\affil[8]{AITHYRA}
\affil[9]{Google Research}
\affil[10]{Microsoft Research}

\date{\vspace{-30pt}}

\recalctypearea

\newcommand{\Rb}{\mathbb{R}}

\newcommand{\gb}{\textsc{GraphBench}}

\newcommand{\new}[1]{\emph{#1}}

\renewcommand{\vec}[1]{\mathbold{#1}}

\DeclareFontFamily{U}{mathx}{\hyphenchar\font45}
\DeclareFontShape{U}{mathx}{m}{n}{<-> mathx10}{}
\DeclareSymbolFont{mathx}{U}{mathx}{m}{n}

\newcommand{\meanstd}[2]{#1{\scriptsize $\pm$ #2}}

\begin{document}

\maketitle

\begin{abstract}
	Machine learning on graphs has made substantial progress across domains such as molecular property prediction and chip design. Yet benchmarking practices remain fragmented, often relying on narrow, task-specific datasets and inconsistent evaluation protocols, hindering reproducibility and broader progress. With the recent popularity of graph foundation models, these weaknesses have become apparent, as existing benchmarks are insufficient for thorough evaluation. To address these challenges, we introduce \gb{}, a comprehensive benchmark suite spanning diverse real-world domains and task settings, including node-level, edge-level, graph-level, and generative tasks. \gb{} provides standardized evaluation protocols, including consistent dataset splits and metrics for assessing out-of-distribution generalization across selected tasks, as well as a unified hyperparameter-tuning framework. We further evaluate \gb{} with recent message-passing neural networks and graph transformer models, establishing principled baselines for future research. See \url{www.graphbench.io} for further details.
\end{abstract}

Machine learning on graphs is central to many applications, including drug design~\citep{Won+2023}, recommender systems~\citep{Yin+2018a}, chip design~\citep{zheng2025deepgate}, combinatorial optimization~\citep{Cap+2021}, and LLM trustworthiness~\citep{frasca2026neural}. Yet benchmarking remains fragmented, i.e., task-specific datasets, inconsistent protocols, narrow domain coverage, and misaligned metrics limit reproducibility and meaningful comparison~\citep{Spe+2025}. In addition, existing benchmarks often overrepresent 2D molecular graphs while underrepresenting other impactful domains such as relational databases, chip design, and combinatorial optimization. Many datasets poorly reflect real-world structural complexity, and reliance on single metrics can encourage overfitting rather than robust generalization. These issues are especially limiting for evaluating graph foundation models (GFMs), which require broad, realistic, and challenging benchmarks.

\paragraph{Shortcomings of current graph learning benchmarks}~
Existing benchmarks have advanced the field but remain limited in scope, scale, and evaluation coverage. \textsc{TUDatasets}~\citep{Mor+2020} offers many graph-level tasks, but most datasets are small, molecular, and lack unified metrics. The \emph{Open Graph Benchmark}~\citep{hu2020ogb} and its extensions~\citep{hu2021ogblsc} provide larger datasets, yet focus mainly on molecules and citation graphs, are partially saturated, support only limited model comparisons, and lack generative tasks. Other benchmarks~\citep{dwivedi2020benchmarking,dwivedi2022long} target narrow prediction regimes, small-scale settings, restricted model sizes, and limited out-of-distribution evaluation. Graph generation benchmarks are particularly underdeveloped~\citep{Spe+2025}, often relying on low-practical-value molecular or synthetic datasets and overlooking large or structurally constrained graph generation. Recent efforts such as dataset quality assessment~\citep{Cou+2025} and \textsc{GraphLand}~\citep{Baz+2025} broaden coverage, but scalability and application gaps remain. As GFMs expand across tasks and domains~\citep{xia2024anygraph, bechlerspeicher2026billionscalegraphfoundationmodels}, \gb{} addresses these gaps by providing a unified suite of challenging, real-world domains for GFM evaluation.

\paragraph{Present work}~ We introduce \gb{}, a next-generation benchmarking suite addressing key shortcomings in graph learning evaluation. \gb{} spans diverse domains---%
social networks,
electronic circuits,
chip design,
combinatorial optimization,
SAT solver prediction,
algorithmic reasoning,
and weather forecasting%
---and supports node-, edge-, graph-level, and generative tasks. It provides standardized splits, domain-specific metrics, dataset-generation instructions, and hyperparameter-tuning scripts, along with out-of-distribution evaluation on selected datasets to assess generalization. We benchmark modern MPNNs and GTs, offering strong baselines and insights into architectural trade-offs. \emph{By unifying heterogeneous tasks under one framework, \gb{} enables reproducible, robust, and impactful graph learning research, paving the way for graph foundation models.}

\section{Overview of GraphBench}

\begin{figure*}[t!]
	\vspace{-0.2ex}
	\centering
	\includegraphics[width=0.7\textwidth, trim={3mm 0 0 3mm}, clip]{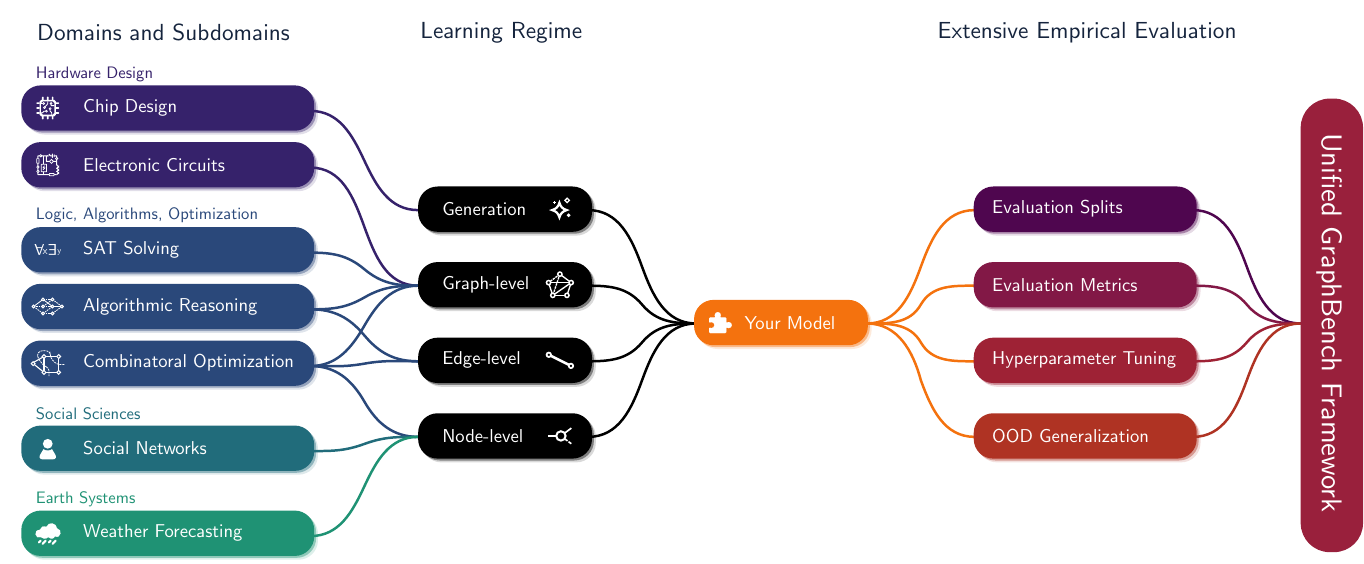}
	\caption{\textbf{Overview of the {GraphBench} framework.} \gb{} integrates datasets from diverse domains---ranging from social sciences and hardware design to logic, optimization, and earth systems---into different \emph{learning regimes} (node-, edge-,  graph-level prediction and generative tasks). \gb{} offers \emph{extensive empirical evaluation} via standardized splits, metrics, hyperparameter tuning scripts, and out-of-distribution (OOD) generalization tests, all within a \emph{unified evaluation pipeline}.
		\label{fig:overview}}
	\vspace{-15pt}
\end{figure*}

The goal of \gb{} is to support the next generation of graph learning research by addressing key shortcomings of existing benchmarks. Here, we provide an overview of the design and features of \gb{}. \autoref{fig:overview} shows an overview of the framework. \gb{} is designed around the following core principles.

\begin{enumerate}[leftmargin=*,topsep=2pt, itemsep=2pt, parsep=1pt]
	\item \textbf{Diverse tasks and domains}~ \gb{} supports node-, edge-, and graph-level prediction, as well as generative tasks, across domains such as social networks, chip and analog circuit design, algorithm performance prediction, and weather forecasting. We describe each domain in \cref{sec:datasets_main}.

	\item \textbf{Real-world impact}~ Unlike benchmarks dominated by small-scale citation or molecular graphs, \gb{} targets modern real-world applications, including large-scale chip design, weather forecasting, and combinatorial optimization, better reflecting industrial and emerging research needs.

	\item \textbf{OOD generalization}~ \gb{} evaluates out-of-distribution generalization through domain-specific temporal shifts or problem-size shifts, moving beyond random train--test splits. Dataset-specific challenges and OOD settings are detailed in the corresponding subsections.

	\item \textbf{Task-relevant evaluation}~ \gb{} provides standardized splits, domain-specific metrics beyond accuracy and mean-squared error, and state-of-the-art hyperparameter tuning scripts. Evaluation and tuning procedures are described in \cref{sec:datasets_main}, \cref{Section:ExpSetup}, and \cref{sec:hpo}.

	\item \textbf{Data generation pipelines}~ For all domains, \gb{} documents the full dataset generation process in \Cref{sec:data_details}, supporting replication, extension, and future expansion of the benchmark suite.
\end{enumerate}
\vspace{-10pt}

\begin{table}[ht!]
	\centering
	\caption{Comparison of graph benchmark datasets across task levels, real-world domains, and OOD evaluation.}
	\label{tab:benchmarks}
	\resizebox{0.5\columnwidth}{!}{%
		\begin{tabular}{lccccc}
			\toprule
			\textbf{Benchmark}
			                                          & \textbf{Node-level}
			                                          & \textbf{Edge-level}
			                                          & \textbf{Graph-level}
			                                          & \textbf{Real-world Domains}
			                                          & \textbf{OOD}                                                                    \\
			\midrule
			OGB  \cite{hu2020ogb, hu2021ogblsc}       & \checkmark                  & \checkmark & \checkmark & \checkmark &            \\
			LRGB  \cite{dwivedi2022long}              & \checkmark                  & \checkmark & \checkmark & \checkmark &            \\
			GraphLand \cite{Baz+2025}                 & \checkmark                  &            & \checkmark & \checkmark &            \\
			\midrule
			CLRS \cite{Velickovic2022CLRSbenchmark}   &                             &            & \checkmark &            &            \\
			SALSA-CLRS \cite{minder2023salsaclrs}     &                             &            & \checkmark &            & \checkmark \\
			Weatherbench2 \cite{rasp2024weatherbench} & \checkmark                  &            &            & \checkmark & \checkmark \\
			\midrule
			\textbf{GraphBench}                       & \checkmark                  & \checkmark & \checkmark & \checkmark & \checkmark \\
			\bottomrule
		\end{tabular}%
	}
	\label{table:comp_benchmarks}
\end{table}

\paragraph{Benchmarks and evaluation}~ To support fair and reproducible comparisons, \gb{} provides a standardized benchmarking protocol for each dataset: a clearly defined prediction or generation task, realistic domain-specific splits, standardized input features and labels, and evaluation scripts with task-specific metrics such as RMSE, ROC-AUC, and closed gap. State-of-the-art hyperparameter tuning scripts further reduce preprocessing effort and enable consistent comparison under realistic conditions.

\paragraph{Baselines and experimental evaluation}~ We evaluate a range of graph learning architectures, including MPNNs, graph transformers, and recent strong baselines. To assess the importance of graph structure, we include MLP and DeepSet~\cite{zaheer2017deep} baselines, which ignore connectivity. We also include GIN as a standard MPNN baseline, additional GNNs for electronic circuit datasets, the recent GNNPlus models~\cite{luo2025GNNPlus}, and a graph transformer based on \citet{stoll2025generalizable}. For weather forecasting, we provide a specialized graph-transformer model inspired by \cite{lam2023learning}. Our results highlight key challenges for current methods, including OOD generalization and efficient training on graph-structured data.

\paragraph{Software and accessibility}~ \gb{} is released as an open-source \textsc{Python} package for easy adoption and extension. It includes streamlined data loaders, \textsc{PyTorch Geometric}-based data objects, predefined splits, and hyperparameter tuning scripts. Its modular design enables researchers to prototype and benchmark new graph learning models under a unified evaluation protocol, while supporting extension to new domains, datasets, and baselines, as illustrated in \Cref{app:extending_graphbench}. Source code is provided in the supplementary material.

\section{Overview of GraphBench's datasets}
\label{sec:datasets_main}
\gb{} encompasses four domains, each further divided into seven subdomains, yielding a total of 38 datasets. We \emph{highlight four subdomains} and describe them in more detail, with exemplary datasets highlighted.
\emph{In-depth descriptions of the remaining subdomains and datasets, along with discussions of related work for each, can be found in \autoref{sec:data_details}.}

\subsection{Social sciences}
\label{subsec:social}

This \gb{} domain models human interactions and societal processes as graphs, with nodes representing individuals or entities and edges representing relationships or communications. It supports tasks such as link prediction, community detection, and temporal forecasting, enabling analysis of real-world social systems.

\subsubsection{Social networks: Predicting engagements on BlueSky}

Social networks---graphs with users as nodes and interactions as edges---are rich benchmarks for graph learning~\citep{newman2010networks}. They contain hubs, peripheral nodes, communities, evolving directed edges, and temporal dependencies driven by changing user interests. However, common datasets from SNAP~\citep{leskovec2016snapgeneralpurposenetwork}, \textsc{TUDatasets}~\citep{Mor+2020}, and benchmarks such as \textsc{Facebook}, \textsc{GitHub}, \textsc{Twitch}, \textsc{LastFM}, and \textsc{Reddit}~\citep{rozemberczki2021multiscaleattributednodeembedding,hamilton2018inductiverepresentationlearninglarge} often collapse interactions into static, undirected graphs with categorical targets and limited or outdated node features. \textsc{GraphLand}~\citep{Baz+2025} improves with temporal splits, but its social datasets largely rely on static, undirected\footnote{Only \textsc{pokec-regions} features \emph{directed} friendship-based edges.} friendship relations and limited profile-based features.

In \gb{}, we use \textsc{Bluesky}\footnote{\url{https://bsky.app/}}, a general, rapidly growing, and openly accessible social platform that exposes public posts and interactions. The task is to predict the number of engagements a user will receive on future posts, capturing near-term influence for applications such as user ranking, trend detection, influencer marketing, and proactive moderation. Compared to prior social graph benchmarks, our dataset provides realistic temporal splits, multi-relational time-evolving directed graphs, and content-grounded user representations from modern language model embeddings.

We next describe the learning task and datasets, with further details on data collection and processing.

\newcommand{\usmashed}{\smash{(}u\smash{)}}

\paragraph{Description of the learning task}~ We cast the problem as node-wise regression to predict an aggregated statistic on the engagements a user will receive on their \emph{future} posts. For a time interval $\tau_{A,B} \coloneqq (t_A,t_B] \coloneqq \{t_A{+}1,\ldots,t_B\}$, we define: \textbf{(i)} an \textit{Interaction graph} $G^{\iota}_{\tau_{A,B}} \coloneqq (V, E^{\iota}_{\tau_{A,B}})$ where $V$, the node set, consists of social network users, and $E^{\iota}_{\tau_{A,B}}$ is the set of directed observed interactions in $\tau_{A,B}$; \textbf{(ii)} \textit{Content} $T^{\usmashed}_{\tau_{A,B}} \coloneqq \{\mathbf{c}_{u,p_t} \mid t \in \tau_{A,B} \}$, the set of the textual contents for posts $p$ authored by user $u$ in $\tau_{A,B}$; \textbf{(iii)} \textit{Engagements} $E^{\kappa,\usmashed}_{\tau_{A,B}} \coloneqq \{e^{\kappa}_{u,p_t} \mid t \in \tau_{A,B} \}$, collecting the number of engagements of kind $\kappa$ received by the posts $p$ authored by user $u$ in $\tau_{A,B}$. We define an aggregate statistic $y^{\kappa, \usmashed}_{\tau_{A,B}} = \mu(E^{\kappa, \usmashed}_{\tau_{A,B}})$.
Given timestamps $t_{0} < t_{1} < t_{2}$, user $u$ and target engagement $\kappa$, the task is to predict value $y^{\kappa, \usmashed}_{\tau_{1,2}}$ given the overall interaction graph $G^\iota_{\tau_{0,1}}$ and post contents $\{T^{\usmashed}_{\tau_{0,1}} \mid u \in V \}$. We assess performance using the mean absolute error (MAE), the coefficient of determination $R^2$, and the Spearman correlation $\sigma$, which measure how well predictions preserve the relative ordering of users by future engagements, a key criterion for ranking influence in heavy-tailed social networks.

\paragraph{Details on the dataset} We build upon the publicly available \textsc{Bluesky} data curated by \citet{failla2024bluesky}, referring the reader to their paper for full details on data collection and curation. Our dataset contains three distinct graph structures $G^{\iota}$'s conveying different interaction kinds: a \emph{directed} edge from node $u$ to node $v$ is drawn when $u$ either \emph{quoted}, \emph{replied to}, or \emph{reposted} $v$'s posts. These interactions are timestamped, and we only construct edges from interactions observed in the interval $(t_0, t_1]$.
For each user $u$, node features are derived from the content of their posts $T^{(u)}$ in the same interval, while distinct prediction targets are obtained as the \emph{median} engagements counts $E^{\kappa,(u)}$ for its posts in time interval $(t_1, t_2]$, i.e., posterior to observed interactions and posts. Original post data span a time interval from $t_{\text{start}}$ to $t_{\text{end}}$, and we fix three key time points $t_{\text{A}}$, $t_{\text{B}}$, $t_{\text{C}}$ obtained so the proportion of posts in $\tau_{\text{start}, A}, \tau_{A, B}, \tau_{B, C}, \tau_{C, \text{end}}$ amounts to, resp., $55\%, 15\%, 15\%, 15\%$. The dataset is accordingly split into training, validation, and test sets. For each split we set $t_0 = t_{\text{start}}$ and choose $t_1, t_2$ as follows: \textit{Training}: $t_{\text{A}}$, $t_2 = t_{\text{B}}$; \textit{Val.}: $t_1 = t_{\text{B}}$, $t_2 = t_{\text{C}}$; \textit{Test}: $t_1 = t_{\text{C}}$, $t_2 = t_{\text{end}}$. See \Cref{sec:social_networks_details} for details on $t_{\text{start}}, t_{\text{end}}, t_0, \ldots, t_2$.

\paragraph{Challenges}~
The temporal split changes the user set, content-based representations, and interaction graph across splits. Engagement counts may also depend on factors outside the training data, such as seasonality. By design, edges are accumulated from $t_{\text{start}}$ to each split cutoff, making validation and test graphs denser than the training graph. This preserves interaction history, especially for users with sparse recent activity, but introduces structural shifts (see \Cref{tab:dataset_statistics}); sliding-window variants are left for future work.

\begin{table*}[t!]
	\tabledefaults
	\centering
	\caption{\textbf{Social Networks.} Performance metrics for each model across datasets. MAE, $R^2$, and Spearman correlation are reported with their standard deviation. For replies and reposts datasets GatedGCN+ ran out of memory in all tested configurations.}
	\label{table:bluesky}
	\vspace{-0.5ex}
	\resizebox{0.7\textwidth}{!}{
		\begin{tabular}{llccccccc}
			\toprule
			\textbf{Dataset} & \textbf{Model} & MAE $\downarrow$       & $R^2_\text{likes} \uparrow$ & $R^2_\text{replies} \uparrow$ & $R^2_\text{reposts} \uparrow$ & $\sigma_\text{likes} \uparrow$ & $\sigma_\text{replies} \uparrow$ & $\sigma_\text{reposts} \uparrow$ \\
			\midrule
			\multirow{6}{*}{quotes}
			                 & DeepSets       & \meanstd{0.810}{0.005} & \meanstd{0.140}{0.002}      & \meanstd{0.102}{0.002}        & \meanstd{0.138}{0.002}        & \meanstd{0.307}{0.004}         & \meanstd{0.178}{0.002}           & \meanstd{0.334}{0.003}           \\
			                 & MLP            & \meanstd{0.784}{0.001} & \meanstd{0.145}{0.001}      & \meanstd{0.107}{0.001}        & \meanstd{0.141}{0.001}        & \meanstd{0.308}{0.003}         & \meanstd{0.176}{0.002}           & \meanstd{0.335}{0.002}           \\
			                 & GNN            & \meanstd{0.768}{0.002} & \meanstd{0.165}{0.002}      & \meanstd{0.134}{0.002}        & \meanstd{0.175}{0.002}        & \meanstd{0.330}{0.003}         & \meanstd{0.192}{0.002}           & \meanstd{0.337}{0.022}           \\
			                 & GIN+           & \meanstd{0.737}{0.014} & \meanstd{0.165}{0.017}      & \meanstd{0.126}{0.030}        & \meanstd{0.166}{0.021}        & \meanstd{0.382}{0.005}         & \meanstd{0.358}{0.010}           & \meanstd{0.405}{0.005}           \\ 
			                 & GatedGCN+      & \meanstd{0.734}{0.010} & \meanstd{0.168}{0.016}      & \meanstd{0.132}{0.019}        & \meanstd{0.173}{0.018}        & \meanstd{0.384}{0.008}         & \meanstd{0.356}{0.008}           & \meanstd{0.411}{0.007}           \\ 
			                 & GCN+           & \meanstd{0.781}{0.021} & \meanstd{0.146}{0.008}      & \meanstd{0.108}{0.010}        & \meanstd{0.145}{0.009}        & \meanstd{0.351}{0.009}         & \meanstd{0.296}{0.010}           & \meanstd{0.351}{0.009}           \\ 
			\midrule
			\multirow{5}{*}{replies}
			                 & DeepSets       & \meanstd{0.789}{0.033} & \meanstd{0.086}{0.033}      & \meanstd{0.061}{0.024}        & \meanstd{0.104}{0.014}        & \meanstd{0.253}{0.003}         & \meanstd{0.130}{0.001}           & \meanstd{0.240}{0.006}           \\
			                 & MLP            & \meanstd{0.725}{0.004} & \meanstd{0.131}{0.003}      & \meanstd{0.087}{0.003}        & \meanstd{0.122}{0.003}        & \meanstd{0.249}{0.004}         & \meanstd{0.127}{0.005}           & \meanstd{0.243}{0.003}           \\
			                 & GNN            & \meanstd{0.694}{0.002} & \meanstd{0.158}{0.003}      & \meanstd{0.119}{0.002}        & \meanstd{0.159}{0.002}        & \meanstd{0.258}{0.009}         & \meanstd{0.132}{0.006}           & \meanstd{0.247}{0.011}           \\
			                 & GIN+           & \meanstd{0.669}{0.017} & \meanstd{0.161}{0.020}      & \meanstd{0.112}{0.019}        & \meanstd{0.153}{0.018}        & \meanstd{0.364}{0.004}         & \meanstd{0.327}{0.009}           & \meanstd{0.365}{0.004}           \\ 
			                 & GCN+           & \meanstd{0.686}{0.010} & \meanstd{0.139}{0.007}      & \meanstd{0.081}{0.010}        & \meanstd{0.128}{0.010}        & \meanstd{0.337}{0.004}         & \meanstd{0.287}{0.005}           & \meanstd{0.344}{0.004}           \\ 
			\midrule
			\multirow{5}{*}{reposts}
			                 & DeepSets       & \meanstd{0.918}{0.013} & \meanstd{0.049}{0.005}      & \meanstd{0.031}{0.012}        & \meanstd{0.050}{0.009}        & \meanstd{0.229}{0.004}         & \meanstd{0.129}{0.007}           & \meanstd{0.205}{0.009}           \\
			                 & MLP            & \meanstd{0.874}{0.003} & \meanstd{0.087}{0.002}      & \meanstd{0.051}{0.003}        & \meanstd{0.066}{0.002}        & \meanstd{0.234}{0.008}         & \meanstd{0.123}{0.006}           & \meanstd{0.206}{0.010}           \\
			                 & GNN            & \meanstd{0.832}{0.009} & \meanstd{0.131}{0.017}      & \meanstd{0.111}{0.022}        & \meanstd{0.128}{0.008}        & \meanstd{0.284}{0.044}         & \meanstd{0.143}{0.032}           & \meanstd{0.260}{0.051}           \\
			                 & GIN+           & \meanstd{0.810}{0.015} & \meanstd{0.150}{0.005}      & \meanstd{0.108}{0.005}        & \meanstd{0.141}{0.010}        & \meanstd{0.339}{0.008}         & \meanstd{0.314}{0.009}           & \meanstd{0.361}{ 0.006}          \\ 
			                 & GCN+           & \meanstd{0.831}{0.013} & \meanstd{0.101}{0.013}      & \meanstd{0.068}{ 0.016}       & \meanstd{0.092}{0.016}        & \meanstd{0.279}{0.005}         & \meanstd{0.262}{0.008}           & \meanstd{0.316}{0.005}           \\ 
			\bottomrule
		\end{tabular}}
	\vspace{-15pt}
\end{table*}

\paragraph{Results}~ \autoref{table:bluesky} shows a consistent ranking from DeepSets to MLP to GNNs across all interaction types (quotes, replies, reposts), with GNNs achieving lower MAE and higher $R^2$ and Spearman~($\sigma$). Since DeepSets and MLPs ignore graph structure, this gap suggests that local, directed relational information helps predict short-term engagement. DeepSets underperforms MLP despite using global feature aggregation, indicating that global aggregation may wash out user-specific signals preserved by a structure-agnostic MLP. However, overall $R^2$ and $\sigma$ remain modest, leaving room for stronger task-specific GNNs. GNNPlus architectures further improve performance across tasks. Due to the graph size, we omit positional encodings. Extended results and implementation details are provided in \Cref{appendix:datasetsarchchoices}.

\subsection{Hardware design}

The hardware design domain represents electronic circuits as graphs, with nodes denoting components such as transistors, resistors, or logic gates, and edges capturing electrical connections. These graphs support tasks such as circuit optimization and performance prediction, which are essential for efficient and reliable hardware development.

\subsubsection{Electronic circuits: Predicting voltage conversion ratio}

\begin{figure}
	\vspace{-1.5ex}
	\centering
	\includegraphics[width=0.45\columnwidth]{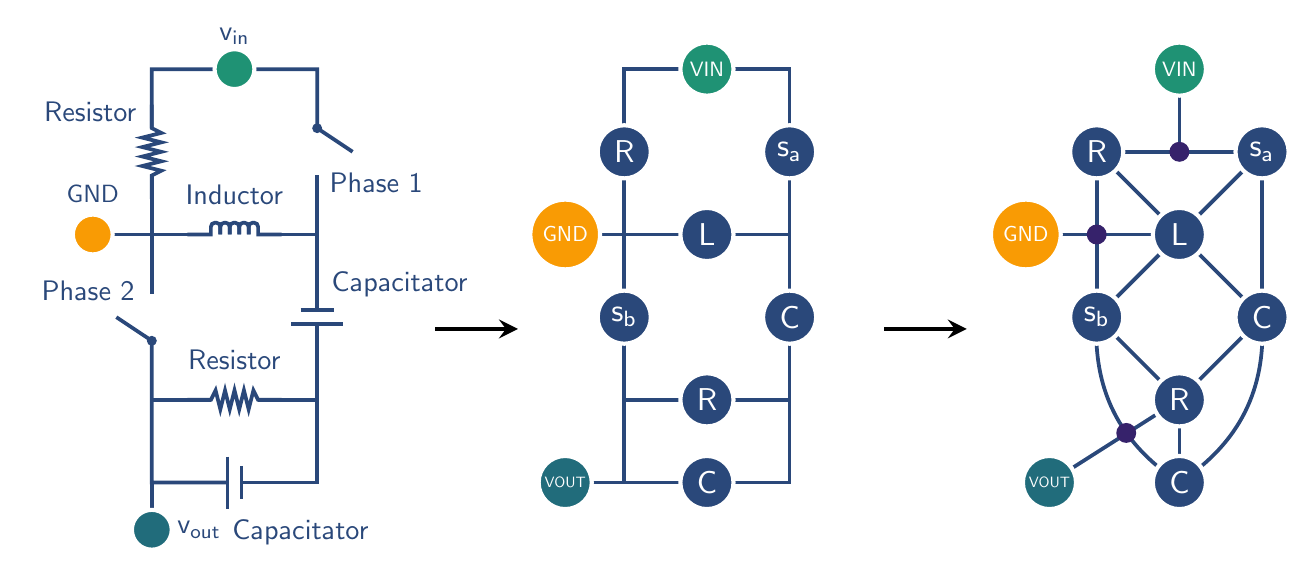}
	\caption{Transformation of an electronic circuit into a graph.
		\label{fig:electronic_circuits}}
	\vspace{-15pt}
\end{figure}

Analog circuit design is critical but remains labor-intensive in electronic design automation~\citep{Zhang2019CircuitGNNGN,Zhang2020GRANNITEGN,zhao2020automated}. Unlike mature digital design flows, it still depends heavily on expert intuition due to device variation and complex constraints. This is especially challenging for custom power converters, where topology generation and parameter tuning must balance competing objectives across vast design spaces. Learning-based surrogate models can accelerate this search by predicting circuit performance~\citep{Fan2021FromST,fan2024graph}. Since power converters are naturally graph-structured, they provide a strong benchmark for testing robustness to structural sensitivity and extreme class imbalance. \autoref{fig:electronic_circuits} illustrates how an electronic circuit is transformed into a graph.

\paragraph{Description of the learning task}~
Each circuit is represented as a graph $G=(V,E)$. The node set $V$ contains device components: capacitors ($C$), inductors ($L$), and two phase-specific switches ($S_a, S_b$), and three external terminal ports: input ($V_{\text{in}}$), output ($V_{\text{out}}$) and ground (\textsc{GND}). Each component node $v\in V$ has a type $\tau(v)$ and (optional) parameter vector $d_v$ and connects to other devices or ports via exactly two edges; each terminal node has a single outgoing edge. Edges $E$ denote electrical interconnections between ports. Given a graph $G$, encoding a power converter, the prediction task is to estimate two continuous performance metrics:
(1) the \emph{voltage conversion ratio} (output-to-input voltage) and
(2) the \emph{power conversion efficiency} (fraction of input power delivered to the load).
Performance is evaluated via the \emph{relative squared error} (RSE), reported separately for each target,
$\mathrm{RSE}(y,\hat y) \coloneqq
	\frac{\sum_{i=1}^{N} (y_i - \hat y_i)^2}{\sum_{i=1}^{N} (y_i - \bar y)^2},$
where $y_i$ is the ground-truth value from high-fidelity simulation, $\hat y_i$ the model prediction, $\bar y$ the mean of $\{y_i\}_{i=1}^N$, and $N$ the number of evaluation samples.

\paragraph{Details on the datasets}~
Following the setting proposed by \citet{fan2024graph}, we used datasets at three complexity levels, generated from random valid topologies with 5, 7, and 10 components. Topologies are sampled uniformly at random under connectivity constraints; isomorphic copies are removed to ensure uniqueness. Each instance is simulated with \textsc{NGSPICE}~\citep{ngspice} to obtain ground-truth voltage conversion ratio and efficiency; instances flagged as invalid during simulation are discarded. Unless otherwise specified, component and operating parameters are fixed as follows:
capacitors $10\,\mu\mathrm{F}$, inductors $100\,\mu\mathrm{H}$, input voltage $100\,\mathrm{V}$, switching frequency $1\,\mathrm{MHz}$, an input resistor of $0.1\,\Omega$ at $V_{\text{in}}$, and a $100\,\Omega$ load with a $10\,\mu\mathrm{F}$ output capacitor at $V_{\text{out}}$.
Generated graphs consist of internal nodes (depending on component size) and three terminal nodes each.

\paragraph{Challenges}~ Aside from domain-specific challenges in circuit design outlined in previous sections, major difficulties in the datasets arise with an increasing number of electronic components in a graph. As the number of components increases, the number of possible circuit arrangements and paths in each graph grows, making it more difficult to correctly predict the target.

\paragraph{Results}~
The results in \autoref{table:ec_res} confirm that the GT consistently achieves higher prediction accuracy than MPNN baselines for seven and ten components across tasks. However, GNNPlus baselines show improved performance on small circuits, indicating a potential advantage of local message passing.
The increase in RSE for larger circuits is mainly due to the reduced availability of training data from costly high-fidelity simulations, combined with the combinatorial growth of circuit space, which makes learning increasingly difficult for all models. However, performance differences among models are evident across all circuit sizes.

\begin{table*}[htbp]
	\centering
	\small
	\caption{Performance comparison of electronic circuits baseline models (RSE, lower is better).}
	\label{table:ec_res}
	\resizebox{0.5\textwidth}{!}{\begin{tabular}{l ccc c ccc}
			\toprule
			                & \multicolumn{3}{c}{\textbf{Efficiency}} &                      & \multicolumn{3}{c}{\textbf{Voltage}}                                                                         \\
			\cmidrule{2-4} \cmidrule{6-8}
			\textbf{Method} & \textbf{5 comp}                         & \textbf{7 comp}      & \textbf{10 comp}                     &  & \textbf{5 comp}      & \textbf{7 comp}      & \textbf{10 comp}     \\
			\midrule
			GT              & \meanstd{0.07}{0.03}                    & \meanstd{0.16}{0.01} & \meanstd{0.37}{0.02}                 &  & \meanstd{0.06}{0.01} & \meanstd{0.25}{0.02} & \meanstd{0.40}{0.03} \\
			GIN             & \meanstd{0.09}{0.06}                    & \meanstd{0.21}{0.04} & \meanstd{0.44}{0.13}                 &  & \meanstd{0.14}{0.05} & \meanstd{0.35}{0.03} & \meanstd{0.54}{0.12} \\
			GIN+            & \meanstd{0.04}{0.01}                    & \meanstd{0.22}{0.02} & \meanstd{0.49}{0.07}                 &  & \meanstd{0.05}{0.01} & \meanstd{0.33}{0.02} & \meanstd{0.64}{0.05} \\
			GCN+            & \meanstd{0.04}{0.01}                    & \meanstd{0.24}{0.05} & \meanstd{0.53}{0.05}                 &  & \meanstd{0.05}{0.01} & \meanstd{0.34}{0.03} & \meanstd{0.60}{0.03} \\
			GatedGCN+       & \meanstd{0.04}{0.01}                    & \meanstd{0.27}{0.07} & \meanstd{0.55}{0.01}                 &  & \meanstd{0.05}{0.01} & \meanstd{0.30}{0.02} & \meanstd{0.59}{0.0}  \\
			\bottomrule
		\end{tabular}}
\end{table*}

\subsubsection{Chip design: Learning to generate small circuits}
\label{sec:cd}

A central challenge in Boolean circuit synthesis is ensuring optimized circuits remain functionally equivalent to the original: regardless of transformations aimed at reducing gate count, delay, power, or area, outputs must match for all inputs. This makes the task delicate---small changes can alter behavior. Finding a minimal circuit is \textsf{NP}-hard, making synthesis a core computational challenge. Classical tools such as \textsc{ABC}~\citep{brayton2010abc} use heuristics to efficiently generate near-optimal circuits. Recently, learning-based methods have emerged as alternatives that use data-driven models to capture structural patterns and generalize across designs \cite{shi2023deepgate2, li2025circuit}. The natural representation of Boolean circuits as directed acyclic graphs (DAGs) makes graph-based generative learning particularly promising. However, generating DAGs is challenging due to the need for exact graph-truth-table matching, and because current methods are not designed to use DAGs as inputs.
We provide a dataset that tasks the network with constructing structurally efficient logic circuits that are equivalent to the given logic functions.
Details on the dataset, arising challenges, and experimental evaluation can be found in \Cref{sec:cd_details}.

\subsection{Reasoning and optimization}

This \gb{} domain encompasses graph learning tasks that require complex reasoning and hard optimization, ranging from theorem proving and hardware verification to logistics and network design. Models must capture fine-grained structural dependencies and generalize across instances. We consider three problem families: (i) \emph{SAT solver selection and performance prediction}, predicting solver behavior or choosing the best solver per instance; (ii) \emph{combinatorial optimization}, including \textsf{NP}-hard problems such as maximum independent set, max-cut, and graph coloring; and (iii) \emph{algorithmic reasoning}, where models approximate polynomial-time graph algorithms. Benchmarks assess accuracy, robustness, scalability, and out-of-distribution generalization.

\subsubsection{Combinatorial optimization: Learning to predict optimal solution objectives}
\label{sec:co}

\new{Combinatorial optimization} (CO) lies at the intersection of optimization, operations research, discrete mathematics, and computer science~\citep{Kor+2012}.
The goal is to find a solution from a discrete set that optimizes a given objective under constraints.
Formally, a \emph{combinatorial optimization problem instance} $I$ is described by a tuple $(\Omega, F, w)$, where
$\Omega(I)$ is a \emph{finite set},
$F(I) \subseteq 2^{\Omega(I)}$ is the set of \emph{feasible solutions}, and
$w \colon \Omega(I) \to \Rb$ assigns \new{weights} to elements of $\Omega(I)$.
The \new{cost} or \new{objective value} of a solution $S \in F(I)$ is given by $c(S) = \sum_{\omega \in S} w(\omega)$. The goal is to find an optimal solution $S_I^* \in F(I)$ that minimizes $c$ over the feasible set (maximization problems can be converted to minimization problems by switching the sign).

CO is central to applications such as vehicle routing, scheduling, and resource allocation~\citep{paschos2014applications}. Graphs are a natural representation for CO because many problems are inherently discrete, network-based, or easily reformulated as graphs. Their computational hardness~\citep{karp1972reducibility} makes CO well-suited for learning-based methods, where MPNNs and GNNs can serve as efficient surrogates for large instances while requiring expressive architectures to solve them effectively~\citep{chen2022milp,chen2024sb}.

\paragraph{Description of the learning tasks}~
We include tasks for both supervised and unsupervised learning settings.
For the supervised setting, the objective is to train a graph learning model to predict the optimal objective value of given CO instances. Given a set of CO problem instances $\mathcal{I}$ with corresponding optimal solutions $S^*_I$, and a model $m \colon \mathcal{I} \to \Rb$, the learning task is to minimize the MAE between the predicted and true objective values,
$\nicefrac{1}{|\mathcal{I}|} \sum_{I \in \mathcal{I}} \left| m(I) - c(S^*_I) \right|$.
While this setup focuses on predicting the objective value rather than producing explicit solutions, it offers a practical advantage: it avoids the challenge of defining a meaningful loss function when multiple distinct optimal solutions exist. This simplification enables consistent, well-defined evaluation and learning. The supervised setting relies on solver-generated solutions, which can be computationally infeasible for large instances.
This motivates the unsupervised setting, which is valuable when ground-truth solutions are unavailable or expensive to obtain, and it directly targets the CO problem.
We provide a differentiable surrogate loss function $\mathcal{L} \colon \Rb^{|\Omega(I)|} \to \Rb$ as well as decoders $d \colon \Rb^{|\Omega(I)|} \to F(I)$ for each CO problem.
The model $m \colon \mathcal{I} \to \Rb^{|\Omega(I)|}$ is trained in an unsupervised fashion to predict a score for each variable that indicates whether it belongs to the solution set, minimizing
$\nicefrac{1}{|\mathcal{I}|} \sum_{I \in \mathcal{I}} \mathcal{L}(m(I))$.
At test time, the model's output scores are converted into a feasible solution to the CO problem using the decoder.
The model's performance is measured based on the objective value of the decoded solution, $c(d(m(I)))$.
This setup is widely used in graph learning for CO literature \citep{karalias2021erdos,min2022can,wenkel2024towards}.
We adapt our surrogate loss functions and decoders from this existing literature.

\paragraph{Details on the datasets}~
We consider three representative hard CO problems on graphs, the \new{maximum independent set} (MIS), the \new{max-cut}, and the \new{graph coloring problem}.
They span a broad spectrum of structural properties and computational challenges.
We formally define these problems in \Cref{sec:co_details}.
In this section, we use MIS as an example to find the largest subset of non-adjacent nodes.
The generality of the $(\Omega, F, w)$ framework makes it applicable to a wide range of CO problems, beyond those proposed in our benchmark.
We use three well-established random graph models to generate problem graphs: RB graphs~\citep{rb_graphs}, Erd\H{o}s-R\'enyi (ER) graphs~\citep{erd6s1960evolution}, and Barab\'asi-Albert (BA) graphs~\citep{albert2002statistical}.
These models are widely used to train CO models.
We generate 50\,000 instances per dataset and provide both small- and large-scale variants. \Cref{sec:co_details} provides a summary of the parameters used for graph generation, which are taken from existing literature. The generated graphs include no node or edge features.

\paragraph{Challenges}~
Since most CO problems are challenging to solve with existing solvers such as KaMIS or Gurobi, a major challenge lies in providing an optimal solution via a graph based method. Furthermore, solving the respective CO problems is shown to require more than 1-WL expressivity \cite{NEURIPS2019_approximationratiosCO}, indicating the necessity for expressive architectures. In addition, the proposed unsupervised target provides additional challenges regarding training of MPNN methods, since most of them are usually trained without a surrogate loss.

\paragraph{Results}~
We show supervised learning results on MIS in \autoref{tab:co_results_su} and unsupervised results, including all three CO problems, in \Cref{sec:co_details}.
Extended unsupervised results, including the other two problems, can be found in \autoref{sec:co_details}.
Among the baselines, GIN achieves the best performance across most datasets. We attribute this to MPNNs' strong inductive bias, which aligns well with the structure of graph-based CO problems.
DeepSet generally outperforms the MLP baseline on the supervised task, except for the \dataset{ER} graphs.
This suggests that global information aggregation is beneficial for this task.
Conversely, the GT performs poorly on most datasets.
We hypothesize that this is due to the training difficulties associated with this task.

\begin{table*}[htbp]
	\tabledefaults
	\centering
	\small 
	\caption{MIS results for each CO dataset with supervised learning (MAE, lower is better).}
	\label{tab:co_results_su}
	\resizebox{0.6\textwidth}{!}{\begin{tabular}{l cc cc cc}
			\toprule
			                & \multicolumn{2}{c}{\textbf{\dataset{RB} graph}} & \multicolumn{2}{c}{\textbf{\dataset{ER} graph}} & \multicolumn{2}{c}{\textbf{\dataset{BA} graph}}                                                                                \\
			\cmidrule(lr){2-3} \cmidrule(lr){4-5} \cmidrule(lr){6-7}
			\textbf{Method} & \textbf{Small}                                  & \textbf{Large}                                  & \textbf{Small}                                  & \textbf{Large}          & \textbf{Small}           & \textbf{Large}          \\
			\midrule
			GIN             & \meanstd{0.491}{0.099}                          & \meanstd{2.125}{0.484}                          & \meanstd{0.234}{0.191}                          & \meanstd{0.352}{0.265}  & \meanstd{0.292}{0.041}   & \meanstd{0.111}{0.016}  \\
			GT              & \meanstd{4.112}{2.353}                          & \meanstd{0.915}{0.235}                          & \meanstd{6.486}{8.101}                          & \meanstd{9.641}{15.335} & \meanstd{3.481}{3.446}   & \meanstd{1.829}{1.383}  \\
			MLP             & \meanstd{1.583}{0.052}                          & \meanstd{1.437}{0.520}                          & \meanstd{0.751}{0.767}                          & \meanstd{0.914}{0.307}  & \meanstd{2.825}{0.949}   & \meanstd{3.383}{0.504}  \\
			DeepSets        & \meanstd{0.918}{0.186}                          & \meanstd{1.427}{0.224}                          & \meanstd{0.756}{0.711}                          & \meanstd{2.244}{0.364}  & \meanstd{2.304}{0.262}   & \meanstd{3.362}{1.491}  \\
			GCN+            & \meanstd{0.024}{0.004}                          & \meanstd{0.015}{0.001}                          & \meanstd{0.011}{0.001}                          & \meanstd{0.009}{0.0001} & \meanstd{0.0009}{0.0002} & \meanstd{0.004}{0.001}  \\
			GIN+            & \meanstd{0.021}{0.0002}                         & \meanstd{0.016}{0.001}                          & \meanstd{0.012}{0.001}                          & \meanstd{0.008}{0.001}  & \meanstd{0.0006}{0.0002} & \meanstd{0.001}{0.0003} \\
			\bottomrule
		\end{tabular}}
\end{table*}

\subsubsection{SAT solving: Algorithm selection and performance prediction}

The Boolean satisfiability problem (SAT) is 
a longstanding \textsf{NP}-complete~\citep{Cook71} problem that remains theoretically significant. Beyond its role in complexity theory, it has many applications, including hardware/software verification~\citep{BiereEtAl09}, automated planning~\citep{KautzSelman96}, and operations research~\citep{GomesEtAl08}. SAT instances can be expressed as graphs, reflecting the permutation-invariance induced by the commutativity and associativity of conjunction and disjunction.
We provide three datasets that span a wide range of instance sizes and, to the best of our knowledge, constitute the largest dataset for algorithm selection and performance prediction for SAT solvers. 
Each is provided with two learning tasks:
\new{performance prediction}, a regression problem whose goal is to predict the computation time of SAT solvers on unseen instances, and \new{algorithm selection}, a multi-class classification problem that aims to select the best performing algorithm for a given SAT instance.
Dataset-specific challenges include noisy target data due to inherently stochastic solver runtimes and large-scale graphs requiring capturing both local and global formula structure.
See \cref{sec:sat_details} for details on the data, baselines, and experimental results.

\subsubsection{Algorithmic reasoning: Learning to simulate algorithms}
\label{Section:AlgoReaso}

In addition to the SAT, many real-world tasks depend on efficient graph algorithms. Given the active exploration of the intersection of algorithms and neural networks across domains~\citep{Esterman2024PUZZLES,Fan2024NPHardEval,Kaiser2015NeuralGPUs,Zaremba2014Learning}, our goal is to provide dedicated datasets for graph algorithmic reasoning. Prior work~\citep{Velickovic2020NeuralExecution,Xu2020nnreasonabout,Bounsi2024TransfmeetNeuralAlgoReas} and benchmarks such as CLRS~\citep{Velickovic2022CLRSbenchmark} have demonstrated that GNN architectures can achieve strong performance on graph problems when provided with hints and graph invariants---tasks that require substantial theoretical expressiveness from graph learning models~\citep{Arvind2020WLsubgraph}.
To broaden this research area, we contribute 21 large-scale datasets that cover seven classic graph algorithms, each at three difficulty levels, thereby expanding the landscape of neural algorithmic reasoning benchmarks. To further investigate the algorithm's learning capabilities, all datasets include a shift in the data distribution of generated graphs, as well as larger graphs in the test sets. Throughout experiments, we observe strong baseline performance of GNN and GT models, while extrapolation regarding size and data distribution is not perfectly achieved.
For details, as well as baselines and experimental results, see \Cref{sec:algoreaso_details}.

\subsection{Earth systems}

The earth systems domain in \gb{} covers graph learning tasks from geospatial, environmental, and climate data. Nodes represent locations, sensors, or regions, while edges encode spatial or physical relationships. These tasks support applications such as weather forecasting, climate impact assessment, and resource management, testing whether graph-based models can integrate heterogeneous features, capture long-range dependencies, and generalize across time and space.

\subsubsection{Weather forecasting: Medium-range atmospheric state prediction}

Weather forecasting is critical for agriculture, energy, and public safety~\citep{diehl2013visual,ramar2014semantic}. Traditional \new{numerical weather prediction} (NWP) is accurate but computationally expensive due to physics-based simulations~\citep{bauer2015quiet}. Recent machine learning methods provide faster, scalable alternatives that can outperform NWP for medium-range forecasts and extreme events, while improving uncertainty quantification~\citep{price2023gencast}. Because earth system data are sparse, irregular, and interconnected~\citep{reichstein2019deep}, with interactions spanning local to planetary scales~\citep{bauer2015quiet}, graph-based models offer a natural way to capture multi-scale, non-Euclidean atmospheric dependencies.

\paragraph{Description of the learning task}~
The objective of the task is to model medium-range weather evolution by predicting the residual change in the atmospheric state over a fixed 12-hour time horizon. Specifically, given an initial snapshot of the current atmospheric state, the model forecasts the twelve-hour future change in meteorological variables. The training objective is a spatially and variable-weighted \emph{mean-squared error} (MSE) for the twelve-hour-ahead prediction, inspired by the training objective used in GraphCast \citep{lam2023learning}. For each verification time $d \in D$, the model uses $x_{d-2},x_{d-3} $ to predict a residual change $\Delta x$, yielding $\hat{x}^{d} = x_{d-2} + \Delta x$. The loss compares $\hat{x}^{d}$ with $x^{d}$. The loss function thus corresponds to
\begin{align*}
	\mathcal{L}(x^{d}, \hat{x}^{d})
	\coloneqq \frac{1}{|D|\,|G|\sum_{j\in J} |L_j|}
	\sum_{d\in D}\sum_{i\in G}\sum_{j\in J}\sum_{\ell\in L_j}
	a_i\, w_j\, s_{j,\ell}\, \bigl(\hat{x}^{d}_{i,j,\ell}-x^{d}_{i,j,\ell}\bigr)^2,
\end{align*}
with level weights $s_{j,\ell}$
where $D$ is the set of forecast date-times, $G$ the grid cells, $J$ the variables, $L_j$ the pressure levels for variable $j$, $a_i=\dfrac{\cos(\mathrm{lat}_i)}{\frac{1}{|G|}\sum_{k\in G}\cos(\mathrm{lat}_k)}$ are mean-normalized latitude weights, $w_j$ are variable weights, and $P_\ell$ are the pressure levels. During evaluation, an unweighted MSE is reported for each variable. Predicting a residual change $\Delta x$ rather than an absolute value improves stability and generalization~\citep{he2020resnet}.

\paragraph{Details on the dataset}~
We utilize reanalysis data from the ERA5 dataset, which has been preprocessed via the WeatherBench2 pipeline~\citep{rasp2024weatherbench}. Multiple resolutions of this dataset are available. We use a downsampled version, containing a $64 \times 32$ equiangular grid, employing a conservative area-preserving interpolation. ERA5 data has a temporal resolution of 6 hours, with time steps at 0h, 6h, 12h, and 18h. Each weather state contains 62 physical and derived variables: 15 are defined across the 13 pressure levels, and 47 are defined at the surface. Of those, we use five variables for the surface and six for the atmospheric levels each, as outlined by~\citet{price2023gencast}. The pressure-level variables include temperature, humidity, and wind speed. The surface-level variables also include information about the location, such as a land-sea mask, sea level pressure, and precipitation. The dataset includes the original grid data as a graph, the icosahedron as a mesh graph, and the edge mapping between the two. Following \cite{price2023gencast}, the mapping is learned during training. Each grid node contains the current weather state $x_{d-2}$ and the previous state $x_{d-3}$, as well as static location information. For variables used in evaluation, see \Cref{table:WeatherforecastingResults}, and for an extended description of the weather forecasting datasets, see \Cref{app:data_preproc}.  

\begin{table}[htbp!]
	\centering
	\caption{MSE for each weather variable for selected pressure levels and surface variables. Lower is better. The selected pressure levels align with the evaluated levels of WeatherBench2 \citep{rasp2024weatherbench}. Persistence is a basic weather forecasting model that provides a forecast by assuming that variable values remain constant, as in the input values. GraphCast is referred to in \citet{lam2023learning}. Number formatting: $k$ indicates a multiplicative factor of $10^3$ while n indicates $10^{-9}$.}

	\resizebox{0.6\columnwidth}{!}{%
		\label{table:WeatherforecastingResults}
		\tiny
		\begin{tabular}{llrrr}
			\toprule
			\vbox{\hbox{\strut \textbf{Pressure}}\hbox{\strut \textbf{Level}}} & \textbf{Variable}             & GT                       & Persistence & GraphCast \\
			\midrule
			\multirow{5}{*}{Surface}
			                                                                   & 2-m temperature (2T)          & 0.841$\tiny\pm$0.013     & 7.123       & 0.068     \\
			                                                                   & 10-m u wind component (10U)   & 1.162$\tiny\pm$0.016     & 2.166       & 0.012     \\
			                                                                   & 10-m v wind component (10V)   & 1.422$\tiny\pm$0.023     & 3.266       & 0.013     \\
			                                                                   & Mean sea level pressure (MSL) & 13.035k$\tiny\pm$0.294k  & 60.056k     & 240.832   \\
			                                                                   & Total precipitation (TP)      & 422.697n$\tiny\pm$9.143n & 714.517n    & 52.377n   \\
			\midrule
			\multirow{6}{*}{500}
			                                                                   & Temperature (T)               & 0.529$\tiny\pm$0.004     & 1.120       & 0.007     \\
			                                                                   & U component of wind (U)       & 4.398$\tiny\pm$0.082     & 6.658       & 0.048     \\
			                                                                   & V component of wind (V)       & 5.843$\tiny\pm$0.081     & 12.988      & 0.053     \\
			                                                                   & Geopotential (Z)              & 958.756$\tiny\pm$17.791  & 48.637k     & 155.057   \\
			                                                                   & Specific humidity (Q)         & 55.49n$\tiny\pm$1.362n   & 66.211n     & 1.090n    \\
			                                                                   & Vertical wind speed (W)       & 0.0051$\tiny\pm$0.00009  & 6.242       & 0.050     \\
			\midrule
			\multirow{6}{*}{700}
			                                                                   & Temperature (T)               & 0.539$\tiny\pm$0.007     & 1.051       & 0.006     \\
			                                                                   & U component of wind (U)       & 2.608$\tiny\pm$0.072     & 3.951       & 0.025     \\
			                                                                   & V component of wind (V)       & 3.312$\tiny\pm$0.082     & 6.754       & 0.027     \\
			                                                                   & Geopotential (Z)              & 634.775$\tiny\pm$15.533  & 32.773k     & 141.296   \\
			                                                                   & Specific humidity (Q)         & 191.388n$\tiny\pm$5.037n & 250.556n    & 3.304n    \\
			                                                                   & Vertical wind speed (W)       & 0.0046$\tiny\pm$0.00013  & 3.352       & 0.027     \\
			\midrule
			\multirow{6}{*}{850}
			                                                                   & Temperature (T)               & 0.735$\tiny\pm$0.018     & 1.351       & 0.009     \\
			                                                                   & U component of wind (U)       & 2.263$\tiny\pm$0.071     & 4.015       & 0.022     \\
			                                                                   & V component of wind (V)       & 2.839$\tiny\pm$0.065     & 6.565       & 0.023     \\
			                                                                   & Geopotential (Z)              & 630.858$\tiny\pm$15.533  & 32.453k     & 139.716   \\
			                                                                   & Specific humidity (Q)         & 276.236n$\tiny\pm$4.399n & 354.058n    & 4.881n    \\
			                                                                   & Vertical wind speed (W)       & 0.0031$\tiny\pm$0.00008  & 3.517       & 0.024     \\
			\bottomrule
		\end{tabular}}
	\vspace{-1.05ex}
\end{table}

\paragraph{Challenges}~
Weather data exhibit a strong temporal distribution shift: climate trends and global weather phenomena change atmospheric statistics over time, requiring models to be robust to more than interpolation. Although naturally graph-structured, the spatial grid is highly regular, with most nodes sharing similar local neighborhoods. This homogeneity makes connectivity alone weakly informative, increasing the importance of positional encodings and global context. The task also requires modeling long-range interactions, which challenge local message passing and can exacerbate oversmoothing. Finally, node features are high-dimensional and heterogeneous, spanning variables with different scales, units, and pressure levels, requiring models to balance competing signals during training.

\paragraph{Results}\label{Weather:Results}~
\Cref{table:WeatherforecastingResults} reports results for each weather variable at three pressure levels, following WeatherBench2~\citep{rasp2024weatherbench}. Alongside GraphCast~\cite{lam2023learning} and the Persistence baseline, we evaluate an additional model based on the GDT implementation of \citet{stoll2025generalizable}. We use a 12-hour forecast horizon across all variables. Our baseline provides a transparent and reproducible lower bound for medium-range weather forecasting. Although it underperforms GraphCast, this is expected given its simpler architecture, shorter training time, and lower-resolution training data. Nevertheless, it improves over Persistence for every variable, indicating effective learning. Implementation details are provided in \Cref{sec:weather_forecasting_details}.

\section{Format, licensing, and long-term access}
\label{sec:format}

Data are released under open licenses: Apache 2.0 for most datasets, including WeatherBench2; MIT for SAT instances; and GPL for AClib. For social networks, the original BlueSky data are hosted at \url{https://zenodo.org/records/11082879} under the Creative Commons Attribution 4.0 International License. These records may be updated to respect users' ``Right to Erasure'', and we are implementing a system to keep our dataset consistent with such updates. All \gb{} datasets are released under the Creative Commons Attribution-NonCommercial 4.0 license to ensure compliance with source licenses, except weather forecasting data, which follows WeatherBench2 and is released under Apache 2.0. \gb{} is continuously updated with new datasets and hosted on institutional servers and public repositories for persistent access.

\section{Conclusion and outlook}

We introduce \gb, a next-generation benchmark suite that addresses the fragmentation of graph learning evaluation. By unifying social sciences, hardware design, reasoning, and optimization, and earth systems within a standardized framework, \gb{} enables fair and reproducible comparison across diverse graph tasks. The suite covers node-, edge-, and graph-level prediction as well as generative tasks, using realistic splits, task-relevant metrics, and explicit out-of-distribution evaluation. Strong baselines with modern MPNNs and graph transformers provide reference points for future work. Our results highlight persistent challenges, including temporal distribution shift, scalability to large graphs, and modeling domain-specific structure. Through principled datasets, consistent evaluation, and robust baselines, \gb{} aims to advance generalizable and practically relevant graph learning.

\paragraph{Vision for \gb}~
\emph{We envision \gb{} as a living benchmark that continually expands to new domains, tasks, and evaluation paradigms, supporting progress in both fundamental graph machine learning research and real-world applications.} We further aim to extend \gb{} to support training next-generation multimodal graph foundation models.

\newpage

\bibliography{bibliography}
\bibliographystyle{icml2026}


\newpage
\appendix
\onecolumn
\section{The \gb{} framework}
\label{app:extending_graphbench}
In this section, we provide information on the \gb{} framework and its proposed expansion. Furthermore, we provide examples for adding new datasets, evaluation procedures, or optimization procedures, as well as guidance on using \gb{} in a custom model pipeline.
\subsection{Future of \gb{}}
We consider \gb{} a continuously expanded and updated benchmarking framework, with new domains and benchmarks added after the initial release. Since graph learning data spans many more domains than our initial dataset collection, we provide this underlying framework to enable the addition of additional datasets and domains. In addition, each domain provides extension capabilities, as larger datasets can be generated from the underlying generation procedures used for the provided datasets.

\begin{figure}[h]
	\centering

	\includegraphics[width=\textwidth]{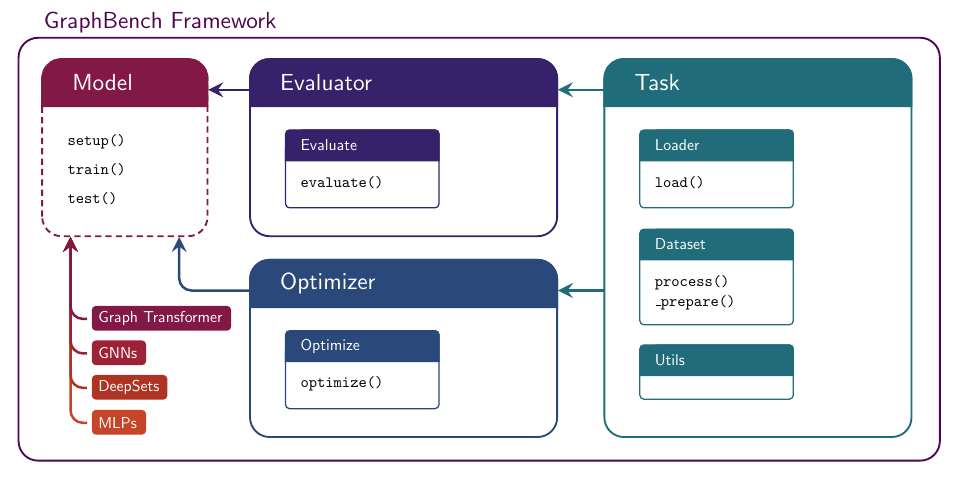}
	\caption{Modular design of GraphBench. Each top-level header includes one foundational part of \gb{} with methods and functions outlined below. \emph{Model}, shown with a dashed line, constitutes a planned addition to the framework.}\label{fig:graphbenchmodular}
\end{figure}

As shown in \autoref{fig:graphbenchmodular}, the \gb{} framework is built around dataset loading, evaluation, and optimization. Each part can be modified independently, allowing the use of only parts of \gb{} in custom models or the replacement of a proposed evaluation pipeline. Moreover, the \gb{} framework aims to be lightweight, relying only on essential packages to provide functionality.

Furthermore, \gb{} is modular by design, allowing collaborators and community members to update parts of \gb{} and expand it with additional datasets or evaluation tasks. Since each evaluator and dataset handler is self-contained and can be used with its corresponding dataset without requiring user implementation, the framework's expansion is significantly streamlined.

Beyond datasets and evaluation methods, we plan to provide additional scripts for GNN and graph transformer baselines, enabling replication and further research on the provided datasets. We present the framework structure along with a proposed expansion in \autoref{fig:graphbenchmodular}.

\subsection{Example usage of \gb{}}
\label{app:example_usage}

We provide a unified and easy-to-use interface for \gb{} datasets. Inspired by benchmarks such as Open Graph Benchmark (\textsc{OGB})~\citep{hu2020ogb}, we use wrapper methods to handle dataset loading. Unlike competing benchmarks, however, \gb{} offers a single loading class (\texttt{graphbench.Loader}) for all datasets, independent of task type. Each wrapper includes the necessary precomputations and comes with predefined splits, enabling direct use in downstream tasks. Nonetheless, we offer customization options for each dataset integrated in the loader method. For seamless integration, all datasets are fully compatible with \textsc{PyTorch Geometric} through its \texttt{InMemoryDataset} interface.
We provide \gb{} as a Python package distributed via PyPi, enabling continuous updates to code and datasets.

Beyond loading datasets, \gb{} provides utilities for experimentation. Given a model, hyperparameter optimization can be run directly via \texttt{graphbench.Optimizer}, which implements the optimization algorithms described in \autoref{sec:hpo}. For evaluation, \texttt{graphbench.Evaluator} loads the corresponding metrics for each selected dataset. Since specific datasets require specialized architectures (e.g., for graph generation), we also supply dataset information and usage guidelines in our code repository. \autoref{listing:graphbench_load} presents pseudocode for training a predefined model using \gb{}.

\begin{lstlisting}[
    caption={\textsc{PyTorch}-like pseudocode for \gb{} showcasing the usage of the dataset loader and application to downstream tasks.},
    label={listing:graphbench_load},
    language=Python,
    basicstyle=\ttfamily\footnotesize,numbers=left
]
import graphbench

model = #your torch model
dataset_name = #name of the task or list of tasks
root = #directory where datasets will be stored
pre_filter = #PyTorch Geometric filter matrix
pre_transform = #PyTorch Geometric-like transform during loading 
transform = #PyTorch Geometric-like transform at computation time

#Setting up the components of graphbench 
evaluator = graphbench.Evaluator(dataset_name)
optimizer = graphbench.Optimizer(optimization_args, training_method)
loader = graphbench.Loader(root, dataset_name, pre_filter, pre_transform, transform)

#load a GraphBench dataset and get splits 
dataset = loader.load()

#optimize your model 
opt_model = optimizer.optimize()

#use graphbench evaluator with targets y_true and predictions y_pred
results = evaluator.evaluate(y_true, y_pred)

\end{lstlisting}

\begin{lstlisting}[
    caption={\textsc{PyTorch}-like pseudocode for \gb{} on how to add additional datasets or evaluation handlers in the framework.},
    label={listing:graphbench_expand},
    language=Python,
    basicstyle=\ttfamily\footnotesize,numbers=left
]
#adding your own evaluator 
#graphbench/evaluator.py
class Evaluator():
...
#define logic for new evaluator
    def get_new_evaluator():
        ...
        #y_true as target values, y_pred as prediction
        #calculation logic in self.new_evaluator()
        return lambda y_pred, y_true: self.new_evaluator(y_pred, y_true)

#add evaluator to list:
    def _get_metric_from_name(self, metric_name):
        metric_dict = {...
        "new_eval": self.get_new_evaluator()
        }
        ...

#add evaluator to master.csv file to enable loading from Evaluator()
#task_type denotes the type of task for logging purposes
#graphbench/master.csv
new_evaluator,task_type,"new_eval"

#adding your own dataset to the loader
#graphbench/loader.py
class Loader():
...
    def _loader(self, dataset_name):
        ...
        elif "new_dataset_name" in dataset_name:
            from graphbench.datasets.new_dataset import newDataset
            train_dataset = newDataset(...)
            valid_dataset = newDataset(...)
            test_dataset = newDataset(...)
        ...

#add a new line to the csv containing the dataset name information
#graphbench/datasets.csv
...
new_dataset_name, new_dataset

#adding logic for new datasets
#graphbench/datasets/new_dataset.py
class newDataset(InMemoryDataset):
    #define torch geometric InMemoryDataset init method (also handles downloading)
    def __init__():
        ...
    #prepare dataset and save it to disk
    def _prepare():
        ...
    #optional processing logic, depending on the prepare method
    def process():
        ...

\end{lstlisting}
Moreover, GraphBench also provides access to the data download and processing handlers. As showcased in \autoref{listing:graphbench_expand}, a new dataset can be easily added by providing download and processing methods, as well as dataset names in the datasets.csv file. Similar instructions hold for the expansion of the evaluation and optimization module. Since optimization is done via SMAC3 \citep{}, optimization modules can be expanded using the established framework.

\begin{table}[h]
	\centering
	\caption{Summary of currently available \gb{} datasets for each subdomain.}
	\resizebox{\columnwidth}{!}{%
		\begin{tabular}{llcccccccccc}
			\toprule
			\textbf{Subdomain} & \textbf{Name}       & \textbf{Node}  & \textbf{Edge}  & \textbf{Directed}            & \textbf{Hetero}              & \textbf{\#Tasks} & \textbf{Split}  & \textbf{Split} & \textbf{Task}               & \textbf{Metric}              \\
			                   &                     & \textbf{Feat.} & \textbf{Feat.} &                              &                              &                  & \textbf{Scheme} & \textbf{Ratio} & \textbf{Type}               &                              \\
			\midrule
			\multirow{3}{*}{Social networks}
			                   & BlueSky -- quotes   & \cmark         & --             & \cmark                       & --                           & 3                & Temporal        & (55/15)/15/15  & Node regression             & MAE / R$^2$ / Spearman corr. \\
			                   & BlueSky -- replies  & \cmark         & --             & \cmark                       & --                           & 3                & Temporal        & (55/15)/15/15  & Node regression             & MAE / R$^2$ / Spearman corr. \\
			                   & BlueSky -- reposts  & \cmark         & --             & \cmark                       & --                           & 3                & Temporal        & (55/15)/15/15  & Node regression             & MAE / R$^2$ / Spearman corr. \\
			\midrule
			\multirow{1}{*}{Chip design}
			                   & AIG                 & \cmark         & \cmark         & \cmark                       & --                           & 1                & Fixed           & 80/10/10       & Generation                  & Ad-hoc Score                 \\
			\midrule
			\multirow{3}{*}{Electronic circuits}
			                   & 5 Components        & \cmark         & --             & --                           & --                           & 1                & Random          & 70/10/20       & Regression                  & RSE                          \\
			                   & 7 Components        & \cmark         & --             & --                           & --                           & 1                & Random          & 70/10/20       & Regression                  & RSE                          \\
			                   & 10 Components       & \cmark         & --             & --                           & --                           & 1                & Random          & 70/10/20       & Regression                  & RSE                          \\
			\midrule
			\multirow{3}{*}{SAT Solving}
			                   & \dataset{Small}     & --             & --             & --                           & \cmark \ \textbackslash \ -- & 5                & Fixed           & 80/10/10       & Classification / Regression & Closed Gap / RMSE            \\
			                   & \dataset{Medium}    & --             & --             & --                           & \cmark \ \textbackslash \ -- & 5                & Fixed           & 80/10/10       & Classification / Regression & Closed Gap / RMSE            \\
			                   & \dataset{Large}     & --             & --             & --                           & \cmark \ \textbackslash \ -- & 5                & Fixed           & 80/10/10       & Classification / Regression & Closed Gap / RMSE            \\
			\midrule
			\multirow{6}{*}{Combinatorial optimization}
			                   & RB -- Small         & --             & --             & --                           & --                           & 6                & Fixed           & 70/15/15       & Regression / Unsupervised   & MAE / CO                     \\
			                   & RB -- Large         & --             & --             & --                           & --                           & 6                & Fixed           & 70/15/15       & Regression / Unsupervised   & MAE / CO                     \\
			                   & ER -- Small         & --             & --             & --                           & --                           & 6                & Fixed           & 70/15/15       & Regression / Unsupervised   & MAE / CO                     \\
			                   & ER -- Large         & --             & --             & --                           & --                           & 6                & Fixed           & 70/15/15       & Regression / Unsupervised   & MAE / CO                     \\
			                   & BA -- Small         & --             & --             & --                           & --                           & 6                & Fixed           & 70/15/15       & Regression / Unsupervised   & MAE / CO                     \\
			                   & BA -- Large         & --             & --             & --                           & --                           & 6                & Fixed           & 70/15/15       & Regression / Unsupervised   & MAE / CO                     \\
			\midrule
			\multirow{7}{*}{Algorithmic reasoning}
			                   & Topological Sorting & \cmark         & \cmark         & \cmark                       & --                           & 3                & Fixed           & 98/1/1         & Node regression             & MAE                          \\
			                   & MST                 & --             & \cmark         & --                           & --                           & 3                & Fixed           & 98/1/1         & Edge classification         & Accuracy / F1                \\
			                   & Bridges             & --             & --             & --                           & --                           & 3                & Fixed           & 98/1/1         & Edge classification         & Accuracy / F1                \\
			                   & Steiner Trees       & \cmark         & \cmark         & --                           & --                           & 3                & Fixed           & 98/1/1         & Edge classification         & Accuracy / F1                \\
			                   & Max Clique          & --             & --             & --                           & --                           & 3                & Fixed           & 98/1/1         & Node classification         & Accuracy / F1                \\
			                   & Flow                & \cmark         & \cmark         & \cmark                       & --                           & 3                & Fixed           & 98/1/1         & Regression                  & MAE                          \\
			                   & Max Matching        & --             & \cmark         & --                           & --                           & 3                & Fixed           & 98/1/1         & Edge classification         & Accuracy / F1                \\
			\midrule
			\multirow{1}{*}{Weather forecasting}
			                   & ERA5 64x32          & \cmark         & \cmark         & \cmark \ \textbackslash \ -- & --                           & 1                & Temporal        & 86/5/9         & Node regression             & MSE                          \\
			\bottomrule
		\end{tabular}}
	\label{tab:graphbench_features}
\end{table}

\begin{table}[htbp]
	\centering
	\caption{Statistics of currently-available \gb{} datasets.}
	\label{tab:dataset_statistics}
	\resizebox{\columnwidth}{!}{%
		\begin{tabular}{llrrrr}
			\toprule
			\textbf{Subdomain} & \textbf{Name}       & \textbf{\#Graphs} & \textbf{Avg. \#Nodes}          & \textbf{Avg. \#Edges}                      & \textbf{Avg. Deg.}       \\
			\midrule
			\multirow{3}{*}{Social networks}
			                   & BlueSky -- quotes   & 1                 & 289\,136 / 286\,425 / 311\,272 & 3\,075\,967 / 3\,815\,996 / 4\,525\,872    & 10.64 / 13.32 / 14.54    \\
			                   & BlueSky -- replies  & 1                 & 470\,816 / 464\,867 / 569\,424 & 9\,287\,565 / 10\,769\,386 / 12\,318\,196  & 19.73 / 23.17 / 21.63    \\
			                   & BlueSky -- reposts  & 1                 & 498\,012 / 508\,818 / 580\,112 & 12\,049\,951 / 14\,658\,114 / 17\,261\,865 & 24.20 / 28.81 / 29.76    \\
			\midrule
			\multirow{1}{*}{Chip design}
			                   & AIG                 & 1\,200\,000       & 125.9                          & 236.3                                      & 3.7                      \\
			\midrule
			\multirow{3}{*}{Electronic circuits}
			                   & 5 Components        & 73\,000           & 13.0                           & 30.0                                       & 7.0                      \\
			                   & 7 Components        & 14\,000           & 17.0                           & 42.0                                       & 9.3                      \\
			                   & 10 Components       & 6\,000            & 24.0                           & 56.0                                       & 12.3                     \\
			\midrule
			\multirow{3}{*}{SAT Solving}
			                   & Small -- VG         & 69\,596           & 1\,323.96                      & 38\,791.52                                 & 33.9                     \\
			                   & Small -- VCG        & 69\,596           & 8\,304.22                      & 45\,402.51                                 & 5.4                      \\
			                   & Small -- LCG        & 69\,596           & 9\,628.12                      & 48\,050.44                                 & 4.75                     \\
			\midrule
			\multirow{6}{*}{Combinatorial optimization}
			                   & RB -- Small         & 50\,000           & 241.0                          & 9\,016.1                                   & 36.9                     \\
			                   & RB -- Large         & 50\,000           & 1\,027.8                       & 102\,644.8                                 & 99.4                     \\
			                   & ER -- Small         & 50\,000           & 249.9                          & 9\,463.5                                   & 37.3                     \\
			                   & ER -- Large         & 50\,000           & 750.2                          & 84\,450.1                                  & 112.4                    \\
			                   & BA -- Small         & 50\,000           & 249.9                          & 991.6                                      & 3.9                      \\
			                   & BA -- Large         & 50\,000           & 750.1                          & 2\,992.4                                   & 3.9                      \\
			\midrule
			\multirow{7}{*}{Algorithmic reasoning-\dataset{easy}}
			                   & Topological Sorting & 1\,010\,000       & 16.0 / 16.0 /128.0             & 41.006 / 40.997 / 1121.008                 & 2.563 / 2.562 / 8.758    \\
			                   & MST                 & 1\,010\,000       & 16.0 / 16.0 /128.0             & 24.024 / 24.004 / 1147.972                 & 1.5015/ 1.5003 / 8.9685  \\
			                   & Bridges             & 1\,010\,000       & 16.0 / 16.0 /128.0             & 20.754/ 20.794/ 277.08                     & 1.2971 / 1.2996 / 2.1647 \\
			                   & Steiner Trees       & 1\,010\,000       & 16.0 / 16.0 /128.0             & 28.739 / 28.733 / 755.284                  & 1.796 / 1.796 / 5.901    \\
			                   & Max Clique          & 1\,010\,000       & 16.0 / 16.0 /128.0             & 77.99 / 77.95 / 2698.5                     & 4.875 / 4.872 / 21.081   \\
			                   & Flow                & 1\,010\,000       & 16.0 / 16.0 /64.0              & 75.76 / 75.84 / 530.96                     & 4.734 / 4.740 / 8.296    \\
			                   & Max Matching        & 1\,010\,000       & 16.0 /16.0 / 128.0             & 38.890 / 38.897 / 746.453                  & 2,431 / 2,431 / 5,834    \\
			\cmidrule{2-6}
			\multirow{7}{*}{Algorithmic reasoning-\dataset{medium}}
			                   & Topological Sorting & 1\,000\,000       & 16.0 / 16.0 /128.0             & 36.167 / 36.137 / 1097.05                  & 2.260 / 2.259 / 8.571    \\
			                   & MST                 & 1\,010\,000       & 16.0 / 16.0 /128.0             & 23.976 / 24.003  / 1140.312                & 1.498/ 1.50 / 8.909      \\
			                   & Bridges             & 1\,010\,000       & 16.0 / 16.0 /128.0             & 17.539/ 17.54/ 277.550                     & 1.0962 / 1.0963 / 2.168  \\
			                   & Steiner Trees       & 1\,010\,000       & 16.0 / 16.0 /128.0             & 19,549 / 19.562 /742.301                   & 1.222 / 1,223 / 5.799    \\
			                   & Max Clique          & 1\,010\,000       & 16.0 / 16.0 /128.0             & 108.0 / 108.0 / 2687.22                    & 6.751 / 6.751 / 20.994   \\
			                   & Flow                & 1\,010\,000       & 16.0 / 16.0 /64.0              & 42.36 / 42.36 / 645.20                     & 2.648 / 2.648 / 10.08    \\
			                   & Max Matching        & 1\,010\,000       & 16.0 / 16.0 /128.0             & 16.305 / 16.306 / 726,908                  & 1,0189/ 1,020 /5,679     \\
			\cmidrule{2-6}
			\multirow{7}{*}{Algorithmic reasoning-\dataset{hard}}
			                   & Topological Sorting & 1\,010\,000       & 16.0 / 16.0 /128.0             & 36.143 / 36.142 / 827.09                   & 2.259 / 2.259 / 6.462    \\
			                   & MST                 & 1\,010\,000       & 16.0 /16.0 /128.0              & 24.008/ 23.998/ 978.825                    & 1,50 / 1.499 / 7.647     \\
			                   & Bridges             & 1\,010\,000       & 16.0 /16.0 /128.0              & 17.539/ 17.54/ 277.08                      & 1.096 / 1.096 / 2.165    \\
			                   & Steiner Trees       & 1\,010\,000       & 16.0 /16.0 /128.0              & 19.583 / 19.565 / 663.547                  & 1.224 /1,223 / 5.184     \\
			                   & Max Clique          & 1\,010\,000       & 16.0 / 16.0 /128.0             & 108.01 / 107.99 / 1812.04                  & 6.751 / 6.750 / 14.157   \\
			                   & Flow                & 1\,010\,000       & 16.0 / 16.0 /64.0              & 42.36 / 42.36 / 645.20                     & 2.648 / 2.648 / 10.08    \\
			                   & Max Matching        & 1\,010\,000       & 16.0 /16.0 / 128.0             & 16,303 / 16,304 / 746,521                  & 1,018/ 1,019 / 5,832     \\
			\midrule
			\multirow{1}{*}{Weather forecasting}
			                   & ERA5 64x32          & 93\,544           & 4\,610                         & 59\,667                                    & 12.9                     \\
			\bottomrule
		\end{tabular}}
\end{table}

\section{Subdomain Details}
\label{sec:data_details}

Here, we provide additional details on the subdomains and their datasets.

\subsection{Social networks: Predicting engagements on BlueSky}
\label{sec:social_networks_details}

\paragraph{Related work}~
A wide range of social network benchmarks are available in repositories such as \textsc{SNAP}~\citep{leskovec2016snapgeneralpurposenetwork} and \textsc{TUdatasets}~\citep{Mor+2020}, yet these datasets diverge substantially from real-world social networks. Benchmarks such as \textsc{Facebook}, \textsc{GitHub}, \textsc{Twitch}, \textsc{LastFM}, \textsc{Deezer}, and \textsc{Reddit}~\citep{rozemberczki2021multiscaleattributednodeembedding,hamilton2018inductiverepresentationlearninglarge} illustrate the main issues: (1) interactions are collapsed into static graphs, eliminating temporal dynamics; (2) prediction targets are categorical (e.g., user type, demographic attribute) rather than continuous behavioral outcomes; (3) topologies are simplified to undirected friendships or follows, omitting richer relations such as replies, reposts, or pull requests; (4) node features are weak or outdated, often sparse indicators or embeddings from GloVe/Word2Vec; and (5) preprocessing alters network structure in disruptive and biasing manners, for example by removing large communities in \textsc{Reddit}.
The \textsc{TUdataset} collection shares similar limitations. Their \textsc{REDDIT} variants represent discussion threads as featureless graphs with reply edges and categorical thread-level labels, while \textsc{COLLAB} and \textsc{IMDB} construct ego-networks labeled by field or genre. These datasets are likewise static, categorical, topologically restricted, feature-poor, and further constrained by artificial graph boundaries that remove long-range dependencies.

The recent \textsc{GraphLand} benchmark~\citep{Baz+2025} represents a substantial step toward more realistic social-network evaluation with the comprised \textsc{pokec-regions}, \textsc{twitch-views}, and the \textsc{artnet} datasets. We note that these include temporal splits and the prediction of continuous node-wise targets, such as view counts for digital content; however, they consider static and, in most cases, undirected friendship- or follow-based edge connectivity.

With respect to these previously proposed datasets, our datasets are designed to support more realistic forecasting, with temporal splits, continuous, forward-looking, engagement-based targets, time-evolving, directed, interaction-based topologies, and node features derived from user-generated post content using modern language-model embeddings. Finally, our datasets do not undergo \emph{arbitrary} subsampling.

\paragraph{More on metrics}~
We assess model performance using three metrics: the Mean Absolute Error (MAE), the coefficient of determination ($R^2$), and the Spearman correlation ($\sigma$). As for $R^2$, given a set of evaluation nodes $U \subset V$, reference engagement kind $\kappa$ and prediction interval $\tau_{1,2}$, it is defined as
\begin{equation*}
	R^2_{\kappa, \tau_{1,2}} \coloneqq 1 - \frac{\sum_{u \in U} (y^{\kappa, (u)}_{\tau_{1,2}} - \hat{y}^{\kappa, (u)})^2}{\sum_{u \in U} (y^{\kappa, (u)}_{\tau_{1,2}} - \bar{y}^{\kappa}_{\tau_{1,2}})^2},
\end{equation*}
where $\hat{y}^{\kappa, (u)}$ is the model's prediction for engagement $\kappa$ and user $u$ and  $\bar{y}^{\kappa}_{\tau_{1,2}}$ is the mean target value over $U$. This standard regression metric indicates the overall variance captured by the model, enabling assessment of its predictive accuracy relative to that of a trivial predictor. The Spearman correlation metric $\sigma$ is defined, instead, as:
\begin{equation*}
	\sigma_{\kappa, \tau_{1,2}} \coloneqq 1 - \frac{6 \cdot \sum_{u \in U}\big(\text{R}[y^{\kappa, (u)}_{\tau_{1,2}}] - \text{R}[\hat{y}^{\kappa, (u)}]\big)^2}{n_U(n_U^2 - 1)},
\end{equation*}
\noindent where $n_U = |U|$ and expression $\big( \text{R}[y^{\kappa, (u)}_{\tau_{1,2}}] - \text{R}[\hat{y}^{\kappa, (u)}] \big)$ evaluates the difference in the ranking of user $u \in U$ within either targets ($y^{\kappa}_{\tau_{1,2}}$) or model's predictions ($\hat{y}^{\kappa}$). This measures how well predictions preserve the relative ordering of users by future engagement, and is not necessarily captured by accuracy metrics such as $R^2$.

\paragraph{More details on the dataset}~

The \textsc{Bluesky} dataset curated by \citet{failla2024bluesky}, is available from \url{https://zenodo.org/records/14669616}. More details on our benchmark construction follow:
\begin{itemize}
	\item \emph{Graph structures.} The dataset contains three graph structures that capture network-wide \emph{interactions} based on quotes, replies, and reposts: a \emph{directed} edge from node $u$ to node $v$ indicates that $u$ quoted, replied to, or reposted $v$'s posts. These interactions are originally timestamped, and we only construct edges from interactions observed in $(t_0, t_1]$. We only consider one single edge when multiple interactions are present in $\tau_{0,1}$ for the same user pair. In future iterations of this dataset, we will consider endowing edges with additional features conveying the amount and time distribution of such interactions.
	\item \emph{Node features.} Each node corresponds to a user, which we describe with an aggregated representation of the content of their posts in the interval $(t_0, t_1]$; these are obtained by a pretrained language model.
	      For each user, we employ the \textsc{sentence-transformers/all-MiniLM-L6-v2} language model to embed the text obtained by concatenating the content of each of their posts at the monthly granularity.\footnote{\url{https://huggingface.co/sentence-transformers/all-MiniLM-L6-v2}} We subsequently aggregate these (by averaging) over the time interval $(t_0, t_1]$.
	\item \emph{Node targets.} These are calculated as the \emph{median} number of engagements obtained by the user's posts ($y^{\kappa, (u)}_{\tau_{A,B}} = \text{median}\!\big(E^{\kappa, (u)}_{\tau_{A,B}}\big)$). These are considered as separate prediction targets and are measured in the time interval $(t_1, t_2]$, i.e., posterior to observed interactions and posts. We observed that the median is a more robust statistic than, e.g., the mean, in this context. We apply a logarithmic transformation to reduce skew in these prediction targets.
	\item \emph{Splitting procedure.}
	      The splitting strategy ensures that models are trained exclusively on past information and evaluated on later interactions. The posts considered in deriving node features and targets span the time interval from February 17th, 2023 ($t_{\text{start}}$) to March 18th, 2024 ($t_{\text{end}}$). We consider three key time points: $t_{\text{A}}$, $t_{\text{B}}$, $t_{\text{C}}$. These are obtained in a way that the proportion of posts in $\tau_{\text{start}, A}, \tau_{A, B}, \tau_{B, C}, \tau_{C, D}$ amounts to, resp., $55\%, 15\%, 15\%, 15\%$.
	      The dataset is then accordingly split into training, validation, and test splits as follows:
	      \begin{itemize}
		      \item Training: $t_0 = t_{\text{start}}$, $t_1 = t_{\text{A}}$, $t_2 = t_{\text{B}}$;
		      \item Validation: $t_0 = t_{\text{start}}$, $t_1 = t_{\text{B}}$, $t_2 = t_{\text{C}}$;
		      \item Test: $t_0 = t_{\text{start}}$, $t_1 = t_{\text{C}}$, $t_2 = t_{\text{end}}$.
	      \end{itemize}
	      In particular:
	      \begin{itemize}
		      \item $t_A$: December 11th, 2023;
		      \item $t_B$: January 22nd, 2024;
		      \item $t_C$: February 20th, 2024.
	      \end{itemize}
	      While ensuring the model is not trained on post contents and network interactions from the future, by setting $t_0 = t_{\text{start}}$ in all splits this partitioning strategy also reflects a realistic scenario where a social network ``grows in time,'' in the sense that user representations evolve as they generate new content and their connections expand as they interact with more users.
\end{itemize}

\subsection{Hardware design}

\subsubsection{Chip design: Learning to generate small circuits}
\label{sec:cd_details}

\begin{figure}[t]
	\centering
	\includegraphics[width=0.6\textwidth]{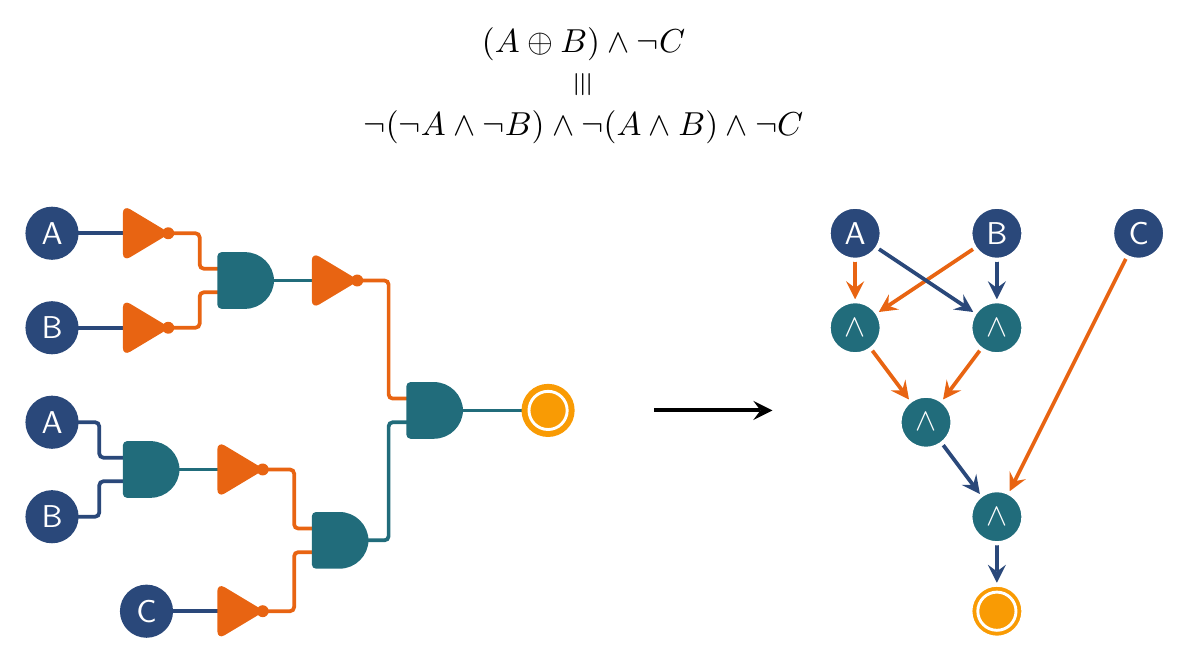}
	\caption{Transformation of a Boolean formula into an And–Inverter Graph (AIG). The left side shows the logic network with explicit gates and inversions; the right side shows the corresponding AIG representation using only AND nodes (teal) and negated edges (orange).
		\label{fig:chip_design}}
\end{figure}

Chip design is among the most complex engineering challenges today, with significant implications for the global economy and supply chains~\citep{Ou2024}. Fabricating a competitive system-on-chip requires arranging billions of transistors under strict performance, power, and area constraints. This process is orchestrated by \new{electronic design automation} (EDA) tools, which automate key stages from high-level synthesis to physical layout. Yet, as design scale and complexity continue to grow, bottlenecks in design space exploration and functional verification are becoming increasingly severe.

\new{Logic synthesis} is a fundamental step within the EDA flow, translating a behavioral specification into a structural implementation of interconnected logic gates~\citep{rai2021logic}. The resulting Boolean circuit provides the core representation of a system’s logic. Optimizing this circuit, typically by minimizing its size (area), is a well-known \textsf{NP}-hard problem~\citep{10.4230/LIPIcs.CCC.2020.22}. For decades, progress has relied on carefully engineered heuristics and combinatorial optimization techniques embedded in EDA tools. However, these classical approaches are reaching diminishing returns, and their computational costs scale poorly with modern circuit complexity~\citep{tsaras2025elf}.

Recently, machine learning has begun to influence key stages of chip design, including macro placement~\citep{alphachip} and floor planning~\citep {mallappa2024floorset}. In this work, we propose focusing these efforts on logic synthesis, framing it as a conditional graph generation problem: generating a circuit that correctly implements a given Boolean function while minimizing the number of gates. The goal is thus to produce circuits that satisfy a prescribed truth table exactly while optimizing structural efficiency.

A central challenge in Boolean circuit synthesis is ensuring optimized circuits remain functionally equivalent to the original. Regardless of transformations that reduce gate count, delay, power, or area, the outputs must match for all inputs. This makes the task delicate---small changes can alter behavior. Finding a minimal circuit is \textsf{NP}-hard, making synthesis a core computational challenge. Classical tools such as \textsc{ABC}~\citep{brayton2010abc} use heuristics to efficiently generate near-optimal circuits. Recently, learning-based methods have emerged as alternatives~\citep{alphachip,li2025circuit} that use data-driven models to capture structural patterns and generalize across designs. The natural representation of Boolean circuits as directed acyclic graphs makes graph-based generative learning particularly promising.

Formally, let $f \colon \{0,1\}^n \to \{0,1\}^m$ be a Boolean function with $n$ inputs and $m$ outputs. Its behavior is fully specified by a truth table $\vec{T}^f \in \{0,1\}^{m \times 2^n}$, where each column $\vec{T}^f_{\vec{x}}$ corresponds to an input $\vec{x} \in \{0,1\}^n$ and satisfies $f(\vec{x}) = \vec{T}^f_{\vec{x}}$. A logic circuit $C \coloneqq (V,E)$ is represented as a \new{directed acyclic graph} (DAG) of in-degree at most two. The node set is partitioned into input nodes $V^{\text{in}} \coloneqq \{v^{\text{in}}_1,\ldots,v^{\text{in}}_n\}$, gate nodes $V^{\text{gate}} \coloneqq \{v^{\text{gate}}_1,\ldots,v^{\text{gate}}_k \}$, each computing a Boolean function of arity at most two (e.g., \textsc{And}), and output nodes $V^{\text{out}} \coloneqq \{v^{\text{out}}_1,\ldots,v^{\text{out}}_m\}$, each of in-degree one. The edge set $E \subseteq V \times V \times \{0,1\}$ specifies connections, where a label of $1$ denotes signal inversion (\textsc{Not}) and $0$ denotes direct transmission. For an input $\vec{x} \in \{0,1\}^n$, values are assigned to input nodes, propagated through gate nodes (applying inversion where specified), and collected at output nodes to yield the circuit output $C(\vec{x}) \in \{0,1\}^m$. Because the combination of \textsc{And} and \textsc{Not} is functionally complete, any Boolean function can be represented using only these two operations. We therefore adopt the standard representation of \new{and-inverter graphs} (AIGs), where circuits are DAGs of \textsc{And} gates connected by edges that may invert their signals~\citep{mishchenko2005fraigs}. AIGs are widely used in industrial logic synthesis and provide a compact, canonical representation of circuits.
\cref{fig:chip_design} provides an overview of how the AIG is constructed from a Boolean formula and how it compares to traditional logic gates.

\paragraph{Related work}~
AlphaChip~\citep{alphachip} kickstarted efforts to bring the latest machine learning advancements into a relatively conservative area of chip design, applying reinforcement learning to the macro placement task and outperforming human engineers on specific metrics, as well as processing designs for newer generations of TPUs~\citep{goldie2024chip}. In logic synthesis, considerable work has focused on discriminative objectives~\citep{zheng2025deepgate} rather than generative ones. One of the main benchmarks in the field is the contest organized by the \emph{International Workshop on Logic \& Synthesis} (IWLS)~\citep{iwls2023}, in which participants are asked to optimize $100$ preselected circuits. Historically, combinatorial optimization methods prevailed due to the lack of training splits and the limited amount of available data. However, recent machine learning methods have shown promising results by training on large amounts of synthetic data simulated with standard EDA tools.

\paragraph{Description of the learning task}~
Given a set of $N$ Boolean functions, $\{f_i\}_{i=1}^N$, $N \in \mathbb{N}$, $f_i \colon  \{0, 1\}^{n_i} \to \{0, 1\}^{m_i}$, the goal of Boolean circuit synthesis is to construct, for each function, a logic circuit $C_i = (V_i, E_i)$ with $n_i$ input and $m_i$ output nodes that satisfies the following conditions.
\begin{enumerate}[leftmargin=.75cm]
	\item \textbf{Logical correctness (constraint)}~ For all $\vec{x} \in \{0, 1\}^{n_i}$, the output of the circuit must match the output of $f_i$, i.e., $C_i(\vec{x}) = f_i(\vec{x})$, where $C_i(\vec{x})$ denotes the vector of output values produced by the circuit on input $\vec{x}$. In that case, $C_i$ is said to be \new{logically equivalent} to $f_i$, which we denote $C_i \equiv f_i$.

	\item \textbf{Structural efficiency (objective)}~ For each $f_i$, among all circuits  $C = (V, E)$ satisfying logical correctness, select a circuit that optimizes a predefined circuit score function $\mathrm{s}(C)$, where $\mathrm{s} \colon \mathcal{C}_{f_i} \to \mathbb{R}_{\geq 0}$, which measures structural properties of the circuit, such as the number of gate nodes,
	      \begin{equation*}
		      \arg\max_{C \in \mathcal{C}_{f_i}} \mathrm{s}(C),
	      \end{equation*}
	      where $\mathcal{C}_{f_i} = \{ C \mid  C \equiv f_i \}$ is the set of all circuits implementing $f_i$. The score function should favor AIGs with fewer internal nodes while penalizing incorrect solutions. Therefore, we define $\mathrm{score}$ over the set of functions $\{f_i\}_{i=1}^N$ as
	      \begin{equation}\label{eq:cd_score}
		      \mathrm{score}(\Tilde{C}) \coloneqq \frac{100}{N} \sum_{i=1}^N \frac{\#C_i^{\mathrm{ref}}}{\#\Tilde{C}_i} \cdot \mathds{1}_{\Tilde{C}_i \equiv f_i},
	      \end{equation}
	      where $\Tilde{C}$ is the set of generated AIGs, $\#\Tilde{C}_i$ and $\#C_i^{\mathrm{ref}}$ are the number of internal nodes of the $i$-th generated circuit and of the corresponding baseline AIG generated by ABC, $\mathds{1}_{\Tilde{C}_i \equiv f_i} = 1$ if $\Tilde{C}_i$ is logically equivalent to $f_i$, and 0 otherwise.
\end{enumerate}
With this benchmark, we introduce a new generative task paired with a challenging optimization objective, i.e., models must generate valid tree-like DAGs that not only satisfy a target truth table but also minimize circuit size. This setting provides a rigorous testbed for evaluating the capabilities of generative graph methods under realistic and computationally challenging constraints.

\paragraph{Details on the dataset}~
For this benchmark, we require paired datasets of truth tables and their corresponding AIGs to train generative models. To construct such a dataset, we first generated 1{\,}200{\,}000 random truth tables in the PLA format~\citep{WGT+:2008}, each with 6-8 inputs and 1 or 2 outputs. These truth tables were then compiled into optimized AIGs in the \textsc{AIGER} format~\citep{biere2011aiger} using \textsc{ABC}. During this step, we applied a sequence of standard optimization commands (\texttt{strash}, \texttt{resyn2}, \texttt{resyn2}, \texttt{dc2}, \texttt{resyn2rs}, \texttt{resyn2}).  We subsequently converted the AIGs into DOT format~\citep{gansner2006drawing} via the \textsc{AIGER} library, from which they can be conveniently transformed into \textsc{PyTorch Geometric} graphs. Alongside these graph representations, we store the truth tables explicitly as matrices of size $m \times 2^n$, where $n$ is the number of inputs and $m$ the number of outputs. We split the dataset into training, validation, and test sets at 80/10/10. Details concerning the dataset are presented in \Cref{tab:aig}.

\begin{table}[h]
	\centering
	\caption{Statistics of the synthetic AIG dataset.}
	\resizebox{0.8\textwidth}{!}{
		\begin{tabular}{c c c c c c}
			\toprule
			\textbf{\#Graphs} & \textbf{\#Inputs} & \textbf{\#Outputs} & \textbf{Mean \#Nodes} & \textbf{Median \#Nodes} & \textbf{Max. \#Nodes} \\
			\midrule
			1.2M              & 6-8               & 1-2                & 125.9                 & 104.0                   & 335.0                 \\
			\bottomrule
		\end{tabular}
	}
	\label{tab:aig}
\end{table}

\paragraph{Experimental evaluation and baselines}~
We report results using four standard ABC variants.
\begin{itemize}
	\item \solver{Strash} converts the circuit into an AIG representation using structural hashing. This serves as the basis for subsequent optimizations.
	\item \solver{Resyn} is a lightweight synthesis script that alternates between balancing and rewriting.
	\item \solver{Compress2} applies balancing, rewriting, and refactoring, together with zero cost; neither method produces a structure that enables further simplifications.
	\item \solver{Resyn2rs} is a heavyweight optimization script that performs multiple rounds of balancing, rewriting, refactoring, and resubstitution, progressively increasing the cut sizes ($K=6$ to $12$).
\end{itemize}
As you may notice, the methods are ordered by complexity, with \solver{Strash} being the simplest and lightest, and \solver{Resyn2rs} the most complex and expensive. Notably, we do not include any learning-based baselines for this dataset. While numerous graph-generative approaches exist (e.g., \cite{vignac2022digress}), most are designed for undirected graphs and thus unsuitable for DAGs. Moreover, our experiments with two DAG-specific generative methods, LayerDAG~\citep{lilayerdag} and Directo~\citep{carballo2025directo}, indicate that neither method produces circuits functionally equivalent to the target truth table. This suggests a new and unexplored direction for the graph generation community, i.e., developing conditional DAG generative models that enforce strict constraints, such as functional equivalence with a given truth table.

\paragraph{Dataset challenges}~
Aside from general graph-generation difficulties, this dataset presents specific challenges related to DAG generation and comparison with truth tables. As providing an exact match to the target truth table is very hard to achieve in general, even with sophisticated solvers as seen below, learning methods struggle to achieve good performance. Furthermore, small changes in generation can lead to substantially worse outcomes, requiring exact methods which most graph generation models are not. In addition to valid DAG generation, the target chip design score also takes circuit size into account, pushing models to learn smaller circuits.

\paragraph{Results}~
Our results using four different ABC variants are reported in \Cref{tab:cd_res}. We report the score defined in \Cref{eq:cd_score}. The methods rank consistently with their complexity across all settings, with the simple \solver{Strash} yielding the worst results and the advanced \solver{Resyn2rs} achieving the best performance.
\begin{table}[htbp]
	\tabledefaults
	\centering
	\caption{\textbf{Chip Design.} Results of the different ABC methods on the \dataset{AIG} dataset. Higher is better.}
	\label{tab:cd_res}
	\begin{tabular}{@{} L{2.4cm} *{7}{C{1.2cm}} @{}}
		\toprule
		                   & \multicolumn{6}{c}{\textbf{Number of (inputs, outputs)}} &                                                 \\
		\cmidrule(l){2-8}
		\textbf{Method}    & (6,1)                                                    & (6,2) & (7,1) & (7,2) & (8,1) & (8,2) & Overall \\
		\midrule
		\solver{Strash}    & 73.90                                                    & 74.23 & 76.55 & 75.71 & 77.63 & 76.52 & 75.76   \\
		\solver{Resyn}     & 86.79                                                    & 88.16 & 89.28 & 89.05 & 90.45 & 89.78 & 88.92   \\
		\solver{Compress2} & 92.53                                                    & 93.57 & 93.18 & 92.79 & 93.28 & 92.49 & 92.97   \\
		\solver{Resyn2rs}  & 93.30                                                    & 95.24 & 95.15 & 95.64 & 96.24 & 96.11 & 95.28   \\
		\bottomrule
	\end{tabular}
\end{table}

\subsubsection{Electronic circuits: Predicting voltage conversion ratio}
\label{sec:ec_details}

\paragraph{Related work}~
Learning for circuit design has been explored for physical implementation~\citep{Lu2020TPGNNAG} and parameter optimization~\citep{Wang2020GCNRLCD}, typically under the assumption of fixed topologies. Topology exploration, in contrast, has relied on heuristic search, including evolutionary and genetic algorithms~\citep{TrustworthyGP2011}, as well as tree-based strategies~\citep{Fan2021FromST,zhao2020automated}. Our benchmark differs in that it treats performance prediction as a supervised graph learning task, using standardized splits and metrics to enable systematic comparison across architectures.

\subsection{Reasoning and Optimization}

\subsubsection{SAT solving: Algorithm selection and performance prediction}
\label{sec:sat_details}

\begin{figure}[t]
	\centering
	\includegraphics[width=0.75\textwidth]{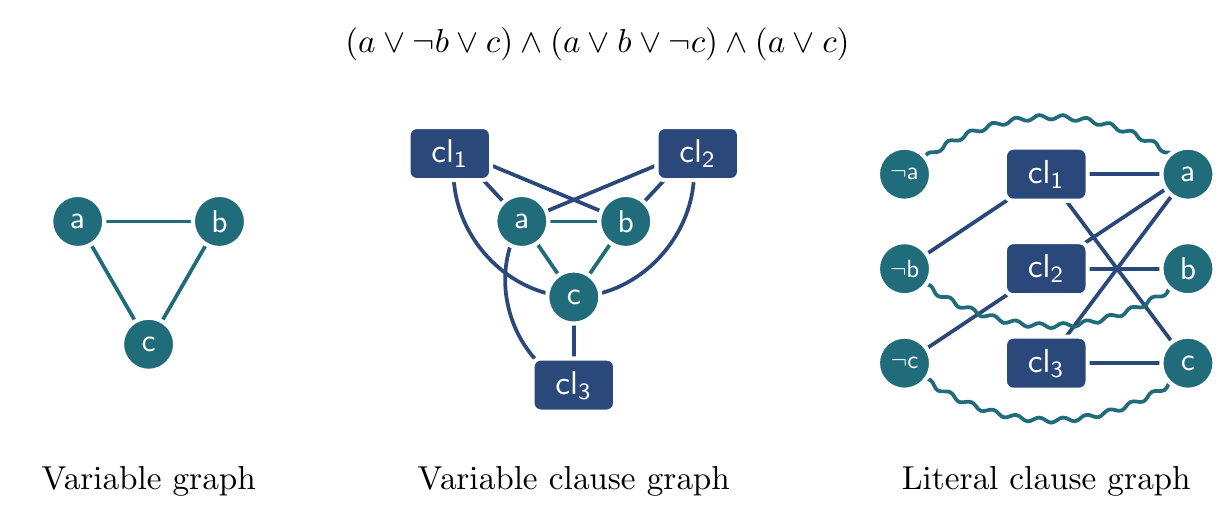}
	\caption{Overview of the three graph types generated from a given SAT formula.
		\label{fig:sat}}
\end{figure}

The \new{Boolean satisfiability problem} (SAT) is a central and longstanding \textsf{NP}-complete problem~\citep{karp1972reducibility,biere_handbook_2021}. It asks whether there exists a truth assignment that satisfies a Boolean formula $\varphi$ over variables $v_1,\ldots,v_r$, typically expressed in \new{conjunctive normal form} (CNF):
\begin{equation*}
	\varphi = \bigwedge_{i=1}^n c_i, \quad \text{where each } c_i = \bigvee_{j=1}^{m_i} l_{i, j}.
\end{equation*}
Here, each $l_{i,j}$ is a literal, i.e., either a variable $v_p$ or its negation $\lnot v_p$. The task is to decide whether there exists an assignment $a$ such that $\varphi(a) = \text{true}$. SAT is of significant theoretical importance as one of the first problems proven \textsf{NP}-complete~\citep{Cook71}, and it underpins numerous practical applications in hardware and software verification~\citep{BiereEtAl09}, automated planning~\citep{KautzSelman96}, and operations research~\citep{GomesEtAl08}. Its significance has led to the development of a wide range of solvers, systematically benchmarked in annual SAT competitions (e.g.,~\citep{FroEtAl21,BalEtAl17,HeuEtAl19}). A key property of modern SAT solvers is \textit{performance complementarity}---no single solver dominates across all instances. This motivates the study of \emph{algorithm selection}~\citep{Rice76,XuEtAl08}, where the aim is to choose the most effective solver for a given instance. \new{Empirical performance prediction} is a closely related problem, which seeks to forecast the computation time (or another performance metric) of a solver on a given instance. Such predictions can support algorithm selection, configuration, scheduling, and explainability~\citep{HutterEtAl14}. State-of-the-art methods for algorithm selection and performance prediction rely on \textit{hand-crafted features} extracted from SAT instances. The most widely adopted feature set is that of the \textsc{SATzilla} family of algorithm selectors~\citep{NudEtAl04,HutterEtAl14,ShaHoo24}. These features include basic structural properties (e.g., number of clauses and variables), graph-based descriptors derived from instance structure, and probing features obtained by briefly running a solver to collect computation-time statistics. Machine learning models---most often tabular models---are then trained on these features for prediction or selection tasks.

SAT instances can be naturally represented as graphs, reflecting their permutation-invariant structure that arises from the commutativity and associativity of logical conjunction and disjunction. Indeed, many of the hand-crafted \textsc{SATzilla} features are derived from such graph representations, including the
\begin{itemize}
	\item \new{Variable-clause graph}: a bipartite, undirected graph with a node for each variable $v$ and each clause $c$, where an edge connects a variable to a clause if and only if the variable appears in that clause.

	\item \new{Variable graph}: an undirected graph with one node per variable, where two variables $v_i$ and $v_j$ are connected if they co-occur in at least one clause.

	\item \new{Clause graph}: an undirected graph with one node per clause, where two clauses $c_i$ and $c_j$ are connected if they share at least one negated literal.
\end{itemize}

\textsc{SATzilla} extracts statistical properties of these graphs---such as mean degree, coefficient of variation, clustering coefficient, and graph diameter---and uses them as features for downstream learning tasks.

Several studies have explored the use of GNNs or MPNNs to predict the satisfiability of a formula, e.g.,~\citet{SelEtAl19}. Some of this work employs an additional representation, the \new{literal-clause graph}, a bipartite graph with one node per literal (a variable or its negation) and one node per clause, with edges linking literals to the clauses they appear in. However, many such approaches are \emph{unsound}---they can produce incorrect results---or cannot guarantee valid satisfying assignments or proofs of unsatisfiability, limiting their practical usefulness~\citep{SelEtAl19,LiEtAl24}.  Other research has integrated MPNNs into existing SAT solvers to replace or guide solver heuristics~\citep{WanEtAl24,TonGro25}. These approaches retain the soundness and completeness guarantees of classical solvers while potentially accelerating the solving process. Nonetheless, benchmarking novel MPNN architectures in this setting is challenging because it requires extensive pretraining and large-scale evaluation.  A complementary line of work applies MPNNs to algorithm selection in SAT~\citep{ZhaEtAl24,Shavit23}, predicting the most effective solver for a given instance directly from graph-based representations.

We add node features to provide the GNN with additional information about the formula's structure. Our node embeddings are inspired by the hand-crafted features used in \textsc{SATzilla}~\citep{NudEtAl04} and the node-level encodings proposed in \textsc{GraSS}~\citep{ZhaEtAl24}. These features capture local syntactic properties (e.g., polarity counts, clause membership statistics), normalized and relational quantities (e.g., positive-to-negative ratio, normalized frequency), and sinusoidal positional encodings for clause nodes. \Cref{tab:vcg_features,tab:lcg_features,tab:vg_features} summarize the node features for each graph type. All features are computed during graph construction and stored as floating-point node attributes.

\begin{table}[htbp]
	\centering
	\caption{Node features for the variable-clause graph (VCG). Variable nodes carry 12 features, and clause nodes carry 22 features. Edges between variables and clauses are attributed with the literal polarity ($+1$/$-1$); reverse edges are included.}
	\label{tab:vcg_features}
	\renewcommand{\arraystretch}{1.2}
	\resizebox{\columnwidth}{!}{%
		\begin{tabular}{@{} p{0.22\textwidth} c p{0.62\textwidth} @{}}
			\toprule
			\textbf{Node type} & \textbf{Dim.} & \textbf{Feature description}                                              \\
			\midrule
			\multirow{11}{=}{\textbf{Variable}\newline(12 features)}
			                   & 1--2          & Constant placeholders (reserved, set to zero)                             \\
			                   & 3             & Number of Horn clauses in which this variable appears                     \\
			                   & 4             & Number of occurrences as a positive literal                               \\
			                   & 5             & Number of occurrences as a negative literal                               \\
			                   & 6             & Total number of clause appearances (degree)                               \\
			                   & 7             & Ratio of positive to negative occurrences                                 \\
			                   & 8             & Number of appearances in binary clauses (length two)                      \\
			                   & 9             & Number of appearances in unit clauses (length one)                        \\
			                   & 10            & Number of appearances in ternary clauses (length three)                   \\
			                   & 11            & Normalised frequency (degree divided by the total number of clauses)      \\
			                   & 12            & Mean length of clauses containing this variable                           \\
			\midrule
			\multirow{13}{=}{\textbf{Clause}\newline(22 features)}
			                   & 1--10         & Sinusoidal positional encoding (10 dimensions)                            \\
			                   & 11            & Number of positive literals                                               \\
			                   & 12            & Number of negative literals                                               \\
			                   & 13            & Ratio of positive to negative literals                                    \\
			                   & 14            & Clause length (number of literals)                                        \\
			                   & 15            & Indicator: unit clause (length one)                                       \\
			                   & 16            & Indicator: binary clause (length two)                                     \\
			                   & 17            & Indicator: ternary clause (length three)                                  \\
			                   & 18            & Indicator: Horn clause (at most one positive literal)                     \\
			                   & 19            & Normalized clause length (divided by the maximum clause length)           \\
			                   & 20            & Normalized clause position (index divided by the total number of clauses) \\
			                   & 21            & Indicator: definite Horn clause (exactly one positive literal)            \\
			                   & 22            & Indicator: negative clause (all literals are negative)                    \\
			\bottomrule
		\end{tabular}}
\end{table}

\begin{table}[htbp]
	\centering
	\caption{Node features for the literal-clause graph (LCG). Literal nodes carry 12 features, and clause nodes carry 22 features. Literal-to-clause edges carry a polarity attribute ($+1$/$-1$); reverse and complement edges (linking each literal to its negation) are also included.}
	\label{tab:lcg_features}
	\renewcommand{\arraystretch}{1.2}
	\resizebox{\columnwidth}{!}{%
		\begin{tabular}{@{} p{0.22\textwidth} c p{0.62\textwidth} @{}}
			\toprule
			\textbf{Node type} & \textbf{Dim.} & \textbf{Feature description}                                              \\
			\midrule
			\multirow{12}{=}{\textbf{Literal}\newline(12 features)}
			                   & 1             & Indicator: positive literal                                               \\
			                   & 2             & Indicator: negative literal                                               \\
			                   & 3             & Number of Horn clauses in which this literal appears                      \\
			                   & 4             & Total number of clause occurrences                                        \\
			                   & 5             & Number of clause occurrences of the complementary literal                 \\
			                   & 6             & Degree (equal to the occurrence count)                                    \\
			                   & 7             & Ratio of this literal's occurrences to those of its complement            \\
			                   & 8             & Number of appearances in binary clauses (length two)                      \\
			                   & 9             & Number of appearances in unit clauses (length one)                        \\
			                   & 10            & Number of appearances in ternary clauses (length three)                   \\
			                   & 11            & Normalized frequency (occurrences divided by the total number of clauses) \\
			                   & 12            & Mean length of clauses containing this literal                            \\
			\midrule
			\multirow{13}{=}{\textbf{Clause}\newline(22 features)}
			                   & 1--10         & Sinusoidal positional encoding (10 dimensions)                            \\
			                   & 11            & Number of positive literals                                               \\
			                   & 12            & Number of negative literals                                               \\
			                   & 13            & Ratio of positive to negative literals                                    \\
			                   & 14            & Clause length (number of literals)                                        \\
			                   & 15            & Indicator: unit clause (length one)                                       \\
			                   & 16            & Indicator: binary clause (length two)                                     \\
			                   & 17            & Indicator: ternary clause (length three)                                  \\
			                   & 18            & Indicator: Horn clause (at most one positive literal)                     \\
			                   & 19            & Normalized clause length (divided by the maximum clause length)           \\
			                   & 20            & Normalized clause position (index divided by the total number of clauses) \\
			                   & 21            & Indicator: definite Horn clause (exactly one positive literal)            \\
			                   & 22            & Indicator: negative clause (all literals are negative)                    \\
			\bottomrule
		\end{tabular}}
\end{table}

\begin{table}[htbp]
	\centering
	\caption{Node features for the variable graph (VG). Each variable node carries 12 features. Edges connect co-occurring variables and carry no attributes.}
	\label{tab:vg_features}
	\renewcommand{\arraystretch}{1.2}
	\resizebox{\columnwidth}{!}{%
		\begin{tabular}{@{} p{0.22\textwidth} c p{0.62\textwidth} @{}}
			\toprule
			\textbf{Node type} & \textbf{Dim.} & \textbf{Feature description}                                                                \\
			\midrule
			\multirow{12}{=}{\textbf{Variable}\newline(12 features)}
			                   & 1             & Number of Horn clauses in which this variable appears                                       \\
			                   & 2             & Number of occurrences as a positive literal                                                 \\
			                   & 3             & Number of occurrences as a negative literal                                                 \\
			                   & 4             & Total co-occurrence count (degree in the variable graph)                                    \\
			                   & 5             & Ratio of positive to negative occurrences                                                   \\
			                   & 6             & Number of appearances in binary clauses (length two)                                        \\
			                   & 7             & Number of appearances in unit clauses (length one)                                          \\
			                   & 8             & Number of appearances in ternary clauses (length three)                                     \\
			                   & 9             & Normalized frequency (clause appearances divided by the total number of clauses)            \\
			                   & 10            & Mean length of clauses containing this variable                                             \\
			                   & 11            & Total number of distinct clauses containing this variable                                   \\
			                   & 12            & Mean co-occurrence degree (co-occurrence count divided by the number of containing clauses) \\
			\bottomrule
		\end{tabular}}
\end{table}

In \gb, we introduce, to the best of our knowledge, the largest algorithm-selection benchmark for SAT solvers, covering eleven solvers and more than 100\,000 instances. Our benchmark supports multiple graph representations of SAT formulae and also provides the \textsc{SATzilla} 2024 feature set~\citep{ShaHoo24}, enabling their combination with MPNNs. Unlike existing GNN benchmarks for SAT~\citep{LiEtAl24}, which primarily focus on satisfiability prediction or assignment generation, our benchmark is designed for practical downstream tasks in SAT solving---specifically, algorithm selection and performance prediction.
\cref{fig:sat} provides an overview of how the three graph types are constructed from a given Boolean formula.

\paragraph{Related work}~
As the use of MPNNs for algorithm selection is a relatively new research direction, no benchmarks exist for this task. Previously, \citet{Shavit23} used a dataset of 3\,000 synthetically generated SAT instances, while \citet{ZhaEtAl24} used a proprietary dataset as well as the 2022 Anniversary track of the SAT Competition, while removing large instances. Outside of the graph domain, the \textsc{ASlib} benchmark suite is used to benchmark algorithm selection methods~\citep{BisEtAL16}. \textsc{ASlib} contains multiple algorithm-selection datasets across various domains, including SAT and TSP. Each dataset contains features extracted from instances (such as \textsc{SATzilla}) as well as the computation times of several algorithms on those instances. We note that \textsc{ASlib} provides SAT scenarios based on a subset of SAT competitions, including instances and solvers from specific years. In contrast, our dataset comprises all available SAT competition instances, as well as numerous instances from other sources.

\paragraph{Description of the learning task}~
We introduce two learning tasks. \new{Performance prediction} is a regression problem that aims to predict the computation time of SAT solvers on unseen instances, arising in applications such as algorithm selection \citep{Rice76} and algorithm configuration \citep[e.g.][]{LinEtAl22}. \new{Algorithm selection} is a multi-class classification problem that aims to select the best algorithm for a given instance.
While algorithm selection can be reduced to performance prediction by choosing the solver with the lowest predicted computation time, we treat them as distinct problems. For algorithm selection, we adopt the loss function proposed by~\citep{ZhaEtAl24}, which penalizes the predicted probabilities in proportion to the solver's computation time. Intuitively, a solver that only slightly underperforms on an instance incurs a small penalty, while poor selections are penalized more heavily, i.e.,
\begin{equation*}
	\frac{1}{N} \sum_{i=1}^N \left( \sum_{k=1}^K p_i^k \cdot t_i^k - t_i^* \right)^2,
\end{equation*}
where $p_i^k$ is the predicted probability of selecting solver $k$ for instance $i$, $t_i^k$ is the computation time of solver $k$ on instance $i$, and $t_i^* = \min_k t_i^k$ is the computation time of the \new{virtual best solver} (VBS). Each instance is provided with
three graph-based representations VG, VCG, and LCG, described above.
In all cases, \textsc{SATzilla} features can be integrated as node attributes to enrich the graph representation.

The performance of a solver $s$ on an instance set $\Pi=\{\pi_1, \pi_2, \dots, \pi_n\}$ is measured using the widely adopted \new{penalized average computation time} ($\text{PAR}_k$) metric, where $k$ denotes the penalty factor for unsolved instances under a cutoff time $c$. The $\text{PAR}_k(s, \Pi)$ score is defined as the mean computation time of $s$ across $\Pi$, where each unsolved instance contributes a computation time of $k \cdot c$. Lower $\text{PAR}_k$ values indicate better performance. For algorithm selectors, both feature extraction and model inference time are included in the total solving time, implying that MPNN inference must be efficient to be practically helpful.

For the performance prediction task, the model output is the base-10 logarithm of the $\text{PAR}_{10}$ score of solver $s$,
\begin{equation*}
	\log_{10}(\text{PAR}_{10}(s, \Pi)),
\end{equation*}
a transformation shown to improve predictive accuracy in prior work~\citep{HutterEtAl14}. Model performance is evaluated using the \new{root mean-squared error} (RMSE) between predicted and actual log-scaled computation times.

For algorithm selection, we use the \new{closed gap} (CG) metric, which quantifies how close a selector comes to the performance of the VBS relative to the single best solver (SBS). Formally,
\begin{equation*}
	\text{CG}(\Pi) = \frac{\text{PAR}_{10}(\text{SBS}, \Pi) - \text{PAR}_{10}(s, \Pi)}{\text{PAR}_{10}(\text{SBS}, \Pi) - \text{PAR}_{10}(\text{VBS}, \Pi)}.
\end{equation*}
A higher CG score indicates stronger selection performance, with CG $=1$ corresponding to VBS performance and CG $=0$ to SBS performance.

\paragraph{Details on the datasets}~
We created the largest algorithm-selection scenario for SAT solvers to date. We collected all SAT Competition instances via GBD~\citep{IseJab24} and added all instances from the AClib benchmark for algorithm configuration~\citep{HutEtAl14b}. Together, these sources include SAT formulae from diverse real-world applications and synthetically generated instances. To further enrich the dataset, we generated additional formulae using three well-established SAT instance generators, i.e.,

\begin{itemize}
	\item \new{FuzzSAT}~\citep{BruEtAl10} produces CNF fuzzing instances originally designed to identify issues in digital circuits. It generates random Boolean circuits as DAGs and converts them into CNF via the Tseitin transformation~\citep{tseitin1983complexity}. To construct the DAG, we used $v \in [50,150]$ input variables; operands were randomly paired and connected by randomly chosen operator nodes until each input was referenced at least $t \in [1,5]$ times. Finally, random clauses of length $s \in [3,8]$ were added to increase complexity.

	\item \new{Cryptography}~\citep{SaeEtAl17,AlaEtAl24} generates SAT instances encoding preimage attacks on MD4, SHA-1, and SHA-256. Each encoding applies multiple \emph{rounds} of operations; higher numbers of rounds yield harder instances. We set the number of rounds to $d \in [16,30]$, which yields formulae that are challenging for modern solvers but not unsolvable. Additionally, we fixed $i \in [0,384]$ of the 512 input bits; fixing more bits results in easier instances.

	\item \new{Community attachment}~\citep{GirLev16} produces pseudo-random SAT formulae with community structure, a property often observed in real-world SAT encodings (e.g., hardware and software verification tasks~\citep{AnsEtAl12}). The generator can create relatively small but difficult instances. We used $n \in [1\,000, 3\,000]$ variables, $c \in [5,100]$ communities, a clause-to-variable ratio of $[3{.}7, 4{.}5]$, and modularity factor $q \in [0{.}3,0{.}9]$. Higher modularity values enforce stronger community structure.
\end{itemize}

For each generator, parameter ranges were chosen to yield challenging but still solvable instances.  To ensure this, we excluded ranges of values yielding no instances solvable by \solver{Kissat}, the winner of the 2024 SAT Competition, between 1 and 5\,000 seconds. After generation, we applied the \textsc{Satellite} preprocessor to remove redundant clauses and variables. A summary of all generated instances is provided in \cref{tab:sat_generatred_instances}.

We pre-selected solvers for the benchmark by constructing a portfolio of $m$ complementary state-of-the-art SAT solvers, based on the results of the 2024 SAT Competition. Portfolios were built using beam search to identify the set of $m$ solvers achieving the VBS performance. To determine $m$, we evaluated portfolios of all possible sizes and plotted portfolio size against achieved VBS. We then identified the inflection point (the ``knee'') of this curve using the Kneedle algorithm~\citep{SatEtlA11} and adopted the corresponding portfolio. The ASF library~\url{https://github.com/hadarshavit/asf} was used to perform the selection procedure. The same procedure was repeated on the SAT Competition 2023.

In total, our dataset includes the following solvers. From the 2024 SAT Competition: Kissat~\citep{BieEtAl24}, BreadIdKissat~\citep{BogEtAl24}, AMSAT~\citep{LiEtAl24b}, and KissatMABDC~\citep{LiuEtAl24}. From the 2023 SAT Competition: SBVA CadiCal~\citep{HabEtAl23}, Reencode Kissat~\citep{ReeEtAl23}, Minisat XOR, KissatMAB Prop PR~\citep{GaoEtAl23}, hKissatInc~\citep{KonEtAl23}, CadiCal vivinst~\citep{BieEtAl23}, and BreakID Kissat~\citep{BogEtAl23}. Solver computation times were measured in CPU time using \texttt{runsolver}~\citep{Roussel11} and the generic wrapper tools~\citep{EggEtAl19}, ensuring accurate and consistent performance measurements.

\begin{table}
	\centering
	\caption{Overview of the instances for the SAT solving benchmark.}
	\resizebox{\columnwidth}{!}{
		\begin{tabular}{ll@{\hskip 0.5in}llll@{\hskip 0.5in}llll}
			\toprule
			\textbf{Source}      & \textbf{\# of Instances} & \multicolumn{4}{c}{\textbf{\# of Variables}} & \multicolumn{4}{c}{\textbf{\# of Clauses}}                                                                                   \\
			                     &                          & min                                          & max                                        & avg        & median     & min    & max           & avg            & median      \\
			\midrule
			SAT Competition      & 31\,024                  & 3                                            & 25\,870\,369                               & 69\,190.77 & 13\,168.50 & 7      & 871\,935\,536 & 1\,022\,025.50 & 118\,526.00 \\
			Community Attachment & 29\,994                  & 922                                          & 2\,956                                     & 1\,931.05  & 1\,928.00  & 3\,566 & 13\,413       & 8\,088.71      & 8\,041.00   \\
			AClib                & 33\,955                  & 6                                            & 248\,738                                   & 3\,319.65  & 1\,270.00  & 24     & 103\,670\,100 & 136\,471.03    & 10\,000.00  \\
			FuzzSAT              & 8\,020                   & 2                                            & 944\,685                                   & 10\,217.75 & 1\,820.00  & 1      & 4\,094\,516   & 45\,638.30     & 8\,074.50   \\
			Cryptography         & 4\,873                   & 6                                            & 12\,415                                    & 4\,264.89  & 3\,225.00  & 24     & 185\,006      & 55\,309.63     & 45\,316.00  \\\midrule
			\textbf{Total}       & 107\,866                 & 2                                            & 25\,870\,369                               & 22\,434.71 & 1\,879.00  & 1      & 871\,935\,536 & 345\,051.72    & 10\,177.50  \\
			\bottomrule
		\end{tabular}
	}
	\label{tab:sat_generatred_instances}
\end{table}
We provide three datasets, \dataset{small}, which includes only small formulae with up to 3\,000 variables and 15\,000 clauses, \dataset{medium}, which consists of the small formulae with additional medium size formulae with size of up to 20\,000 variables and 80\,000 clauses and \dataset{large}, which includes all formulae. This way, we provide datasets with varying levels of hardness, as formula size can pose hurdles for training time and GPU memory. While the \dataset{large} and \dataset{medium} datasets are unlikely to be accessible on current hardware, we provide them to keep the benchmark relevant in the long term.

\paragraph{Dataset challenges:} Algorithm selection and performance prediction are particularly challenging in this setting for several reasons. First, solver performance is inherently stochastic: repeated runs of the same solver on the same instance can result in substantially different running times. This induces noisy target values that the learning algorithm must account for. Second, the targets are right-censored due to the use of solver cutoffs: when a solver does not terminate before the cutoff, its true running time is only known to exceed the cutoff. Third, the SAT instances can be very large, with corresponding graphs containing thousands to millions of variable nodes, making scalable graph processing a major challenge. Finally, accurate prediction requires the GNN to capture both local structural patterns and the formula's global properties.

\paragraph{Experimental evaluation and baselines}~
All SAT solving tasks in the benchmark are graph-level. For the graph transformer baselines, we employ an encoder-processor-decoder pipeline, as described in~\Cref{Section:ExpSetup}. We used the following architectures: GIN, GCN, GIN+, GCN+, Gated GCN+. While we considered GT for the tasks, we encountered high computational costs in calculating the positional embeddings (more than a CPU day) and therefore excluded them from the evaluation. To ensure a fair comparison with hand-crafted features and to evaluate the capacity of GNNs to extract relevant information purely from the formula's topology, we did not provide the models with any initial node-level features, forcing them to learn solely from the graph structure. Furthermore, when training on medium- and large-sized graphs, we encountered out-of-memory errors even on an H100. Consequently, we only consider small formulae for MPNNs. Similarly, we observed out-of-memory issues with the clause graphs and therefore excluded them from the evaluation. We optimized the hyperparameters of these models using SMAC3~\citep{LinEtAl22} with a multi-fidelity facade, running 60 trials. Full details of the hyperparameter configuration space and the tuning procedure can be found in~\Cref{sec:hpo}.

We note that traditionally, algorithm selection datasets are evaluated using cross-validation, primarily due to limited data availability. In contrast, our new dataset contains tens of thousands of formulas. Since it is also intended for deep learning approaches, which are computationally expensive and thus impractical to evaluate using 10-fold cross-validation, we instead provide a fixed train–validation–test split and report results across 3 random seeds.

For the performance prediction task, we compare the MPNNs against traditional baselines, which use the \solver{SATzilla} features and a (tabular) machine learning model. We use random forest~\citep{Breiman01} and XGBoost~\citep{CheEtAl16}, which are commonly used as empirical performance models (EPM) (see, e.g., \citet{BanEtAl22,EggEtAl15,HutEtAl14}).
We use the \solver{SATzilla} 2024 features as input and predict the log10-scaled computation time. These baselines are implemented using the ASF library~\url{https://github.com/hadarshavit/asf}.
For algorithm selection, we provide several well-known baselines. First, we use pairwise regression~\citep{KerEtAl18} (PR; predicts performance differences between each pair of solvers, then picks the solver with the best sum), pairwise classification~\citep{XuEtAl12} (PC; predicts the better solver in each pair, then uses majority vote, 
winner of the ICON challenge on algorithm selection~\citep{LinEtAl19}
) as well as a simple multi-class classification~\citep{KotEtAl13} (MC; directly predicts the best solver). Similar to the EPM baselines, these baselines use the \solver{SATzilla} 2024 features with a random forest. We tune these baselines using SMAC3~\citep{LinEtAl22} for 50 trials.

\begin{table}[htbp]
	\tabledefaults
	\centering
	\caption{\textbf{SAT Solving.} RMSE (log\textsubscript{10} time) for performance prediction on \dataset{small} SAT instances. We report results on three different graph representations (\dataset{VG}, \dataset{VCG}, \dataset{LCG}) and using the hand-crafted SATzilla features. Lower is better.}
	\label{table:sat_epm_res_small}
	\resizebox{0.6\textwidth}{!}{%
		\begin{tabular}{llcccc}
			\toprule
			\textbf{Solver} & \textbf{Method} & \dataset{VG}         & \dataset{VCG}        & \dataset{LCG}        & \dataset{SATzilla}   \\
			\midrule
			\multirow{8}{*}{\solver{Kissat}}
			                & GIN             & \meanstd{0.66}{0.02} & \meanstd{0.64}{0.07} & \meanstd{0.70}{0.01}                        \\
			                & GCN             & \meanstd{0.98}{0.00} & \meanstd{0.70}{0.01} & \meanstd{0.82}{0.07}                        \\
			                & GCN+            & \meanstd{0.69}{0.03} & \meanstd{0.51}{0.02} & \meanstd{0.67}{0.09}                        \\
			                & GIN+            & \meanstd{0.78}{0.02} & \meanstd{0.87}{0.05} & \meanstd{0.83}{0.07}                        \\
			                & Gated GCN+      & --                   & \meanstd{1.05}{0.04} & \meanstd{1.04}{0.04}                        \\
			                & RF              &                      &                      &                      & \meanstd{0.65}{0.00} \\
			                & XGB             &                      &                      &                      & \meanstd{0.64}{0.00} \\
			\midrule
			\multirow{8}{*}{\solver{BreakIDKissat}}
			                & GIN             & \meanstd{0.65}{0.01} & \meanstd{0.63}{0.04} & \meanstd{0.68}{0.01}                        \\
			                & GCN             & \meanstd{0.96}{0.00} & \meanstd{0.70}{0.01} & \meanstd{0.80}{0.09}                        \\
			                & GCN+            & \meanstd{0.69}{0.01} & \meanstd{0.58}{0.03} & \meanstd{0.62}{0.03}                        \\
			                & GIN+            & \meanstd{0.82}{0.10} & \meanstd{0.89}{0.03} & \meanstd{0.95}{0.08}                        \\
			                & Gated GCN+      & --                   & \meanstd{0.94}{0.02} & \meanstd{0.94}{0.03}                        \\
			                & RF              &                      &                      &                      & \meanstd{0.68}{0.00} \\
			                & XGB             &                      &                      &                      & \meanstd{0.67}{0.00} \\
			\midrule
			\multirow{8}{*}{\solver{KissatMABDC}}
			                & GIN             & \meanstd{0.64}{0.03} & \meanstd{0.72}{0.01} & \meanstd{0.70}{0.02}                        \\
			                & GCN             & \meanstd{0.99}{0.00} & \meanstd{0.75}{0.02} & \meanstd{0.74}{0.01}                        \\
			                & GCN+            & \meanstd{0.74}{0.07} & \meanstd{0.86}{0.01} & \meanstd{0.78}{0.07}                        \\
			                & GIN+            & \meanstd{0.83}{0.09} & \meanstd{1.05}{0.06} & \meanstd{0.94}{0.01}                        \\
			                & Gated GCN+      & --                   & \meanstd{0.91}{0.15} & \meanstd{1.05}{0.18}                        \\
			                & RF              &                      &                      &                      & \meanstd{0.70}{0.00} \\
			                & XGB             &                      &                      &                      & \meanstd{0.69}{0.00} \\
			\midrule
			\multirow{8}{*}{\solver{AMSAT}}
			                & GIN             & \meanstd{0.71}{0.02} & \meanstd{0.70}{0.01} & \meanstd{0.71}{0.01}                        \\
			                & GCN             & \meanstd{1.01}{0.00} & \meanstd{0.82}{0.10} & \meanstd{0.75}{0.01}                        \\
			                & GCN+            & \meanstd{0.85}{0.09} & \meanstd{0.61}{0.02} & \meanstd{0.74}{0.06}                        \\
			                & GIN+            & \meanstd{0.92}{0.07} & \meanstd{1.05}{0.04} & \meanstd{1.04}{0.06}                        \\
			                & Gated GCN+      & --                   & \meanstd{1.02}{0.01} & \meanstd{1.03}{0.07}                        \\
			                & RF              &                      &                      &                      & \meanstd{0.74}{0.00} \\
			                & XGB             &                      &                      &                      & \meanstd{0.74}{0.00} \\
			\bottomrule
		\end{tabular}}
\end{table}

\begin{table}[htbp]
	\tabledefaults
	\centering
	\caption{\textbf{SAT Solving.} RMSE (log\textsubscript{10} time) for performance prediction on \dataset{medium} and \dataset{large} SAT instances. Lower is better.}
	\label{table:sat_epm_res_med_large}
	\begin{minipage}{0.48\textwidth}
		\centering
		\subcaption{Medium-size SAT instances}
		\resizebox{0.8\linewidth}{!}{%
			\begin{tabular}{llc}
				\toprule
				\textbf{Solver} & \textbf{Method} & \dataset{SATzilla}   \\
				\midrule
				\multirow{2}{*}{\solver{Kissat}}
				                & RF              & \meanstd{0.63}{0.00} \\
				                & XGB             & \meanstd{0.62}{0.00} \\
				\midrule
				\multirow{2}{*}{\solver{BreakIDKissat}}
				                & RF              & \meanstd{0.66}{0.00} \\
				                & XGB             & \meanstd{0.65}{0.00} \\
				\midrule
				\multirow{2}{*}{\solver{KissatMABDC}}
				                & RF              & \meanstd{0.68}{0.00} \\
				                & XGB             & \meanstd{0.67}{0.00} \\
				\midrule
				\multirow{2}{*}{\solver{AMSAT}}
				                & RF              & \meanstd{0.70}{0.00} \\
				                & XGB             & \meanstd{0.69}{0.00} \\
				\bottomrule
			\end{tabular}}
		\label{table:sat_epm_res_med}
	\end{minipage}
	\hfill
	\begin{minipage}{0.48\textwidth}
		\centering
		\subcaption{Large SAT instances}
		\resizebox{0.8\linewidth}{!}{%
			\begin{tabular}{llc}
				\toprule
				\textbf{Solver} & \textbf{Method} & \dataset{SATzilla}   \\
				\midrule
				\multirow{2}{*}{\solver{Kissat}}
				                & RF              & \meanstd{0.64}{0.00} \\
				                & XGB             & \meanstd{0.64}{0.00} \\
				\midrule
				\multirow{2}{*}{\solver{BreakIDKissat}}
				                & RF              & \meanstd{0.66}{0.00} \\
				                & XGB             & \meanstd{0.65}{0.00} \\
				\midrule
				\multirow{2}{*}{\solver{KissatMABDC}}
				                & RF              & \meanstd{0.67}{0.00} \\
				                & XGB             & \meanstd{0.66}{0.00} \\
				\midrule
				\multirow{2}{*}{\solver{AMSAT}}
				                & RF              & \meanstd{0.69}{0.00} \\
				                & XGB             & \meanstd{0.69}{0.00} \\
				\bottomrule
			\end{tabular}}
		\label{table:sat_epm_res_large}
	\end{minipage}
\end{table}

\paragraph{Results}~
We start by presenting the results for the performance prediction task. We present the RMSE of the log-scaled computation time for the solvers from SAT Competition 2024 (\solver{Kissat}, \solver{BreakIDKissat}, \solver{KissatMABDC}, \solver{AMSAT}) for the small formulae in table~\Cref{table:sat_epm_res_small}. Note that missing values (--) for Gated GCN Plus with the VG representation indicate that the model encountered out-of-memory (OOM) errors on the GPU.
The table shows that the best performance for \solver{Kissat}, \solver{BreakIDKissat}, and \solver{AMSAT} is achieved by GCN Plus with the VCG graph, as this model outperforms the SATzilla features. For \solver{KissatMABDC}, a simple GIN with a variable graph performs best. We also note that the SAT-specific architecture, GraSS, performs surprisingly poorly across the board compared to general-purpose architectures.
To the best of our knowledge, this is the first time the SATzilla features have been outperformed on the performance prediction task. We furthermore present the results of random forest and XGBoost on medium-size and large formulae in~\Cref{table:sat_epm_res_med,table:sat_epm_res_large}, such that they can be compared against graph machine learning solutions in the future. Overall, we observe similar performance between XGBoost and random forest, with a slight advantage to XGBoost.

For the algorithm selection task, we report the
closed gap metric of different selectors in~\Cref{table:sat_as_res}.
For the small instances, we observe that gated GCN plus performs best, followed by GCN plus and GIN, all of which are trained on literal-clause graphs. These graph-based approaches outperform the SATzilla feature-based baselines, with only pairwise regression achieving a positive closed gap. Comparing the different graph representations across tasks, we observe clear trade-offs: while the VCG representation yields the strongest performance for the prediction task, the LCG representation enables the highest gap closure for algorithm selection. This suggests that the explicit literal polarity and complementary edges in LCG might be more beneficial for relative performance ranking than for absolute runtime prediction.
Among the three baselines based on hand-crafted features, we observe that multi-class classification underperforms both pairwise classification and regression, leaving a negative gap on the small and medium-sized datasets. This highlights that algorithm selection is not a simple multi-class classification problem. In that formulation, the model loses valuable information about the performance of algorithms other than the best one.
Finally, we find more gaps closed for the large dataset across all three baselines, as it includes harder instances. As time increases, the benefit of algorithm selection grows because feature computation accounts for a smaller fraction of the total computation time. Moreover, for harder instances, multiple solvers frequently time out, so choosing a solver that successfully solves the instance becomes especially valuable, given the additional penalty imposed by the PAR score.

\begin{table}[htbp]
	\tabledefaults
	\centering
	\caption{\textbf{SAT Solving.} Algorithm selection results (gap closed; higher is better).}
	\label{table:sat_as_res}
	\resizebox{0.6\textwidth}{!}{%
		\begin{tabular}{llcccc}
			\toprule
			\textbf{Instances} & \textbf{Method} & \dataset{VG}         & \dataset{VCG}        & \dataset{LCG}        & \dataset{SATzilla}    \\
			\midrule
			\multirow{9}{*}{\dataset{Small}}
			                   & GIN             & \meanstd{0.45}{0.00} & \meanstd{0.45}{0.00} & \meanstd{0.45}{0.01}                         \\
			                   & GCN             & \meanstd{0.00}{0.00} & \meanstd{0.40}{0.00} & \meanstd{0.38}{0.03}                         \\
			                   & GCN+            & \meanstd{0.31}{0.00} & \meanstd{0.41}{0.01} & \meanstd{0.47}{0.02}                         \\
			                   & GIN+            & \meanstd{0.28}{0.05} & \meanstd{0.02}{0.01} & \meanstd{0.16}{0.08}                         \\
			                   & Gated GCN+      & --                   & \meanstd{0.40}{0.04} & \meanstd{0.48}{0.03}                         \\
			                   & PR              &                      &                      &                      & \meanstd{0.38}{0.02}  \\
			                   & MC              &                      &                      &                      & \meanstd{-0.44}{0.01} \\
			                   & PC              &                      &                      &                      & \meanstd{-0.26}{0.00} \\
			\midrule
			\multirow{3}{*}{\dataset{Medium}}
			                   & PR              &                      &                      &                      & \meanstd{0.35}{0.01}  \\
			                   & MC              &                      &                      &                      & \meanstd{-0.43}{0.04} \\
			                   & PC              &                      &                      &                      & \meanstd{0.01}{0.01}  \\
			\midrule
			\multirow{3}{*}{\dataset{Large}}
			                   & PR              &                      &                      &                      & \meanstd{0.51}{0.01}  \\
			                   & MC              &                      &                      &                      & \meanstd{-0.05}{0.01} \\
			                   & PC              &                      &                      &                      & \meanstd{0.21}{0.01}  \\
			\bottomrule
		\end{tabular}}
\end{table}

\subsubsection{Combinatorial optimization: Learning to predict optimal solution objectives}
\label{sec:co_details}

\begin{figure}[t]
	\centering
	\includegraphics[width=\columnwidth]{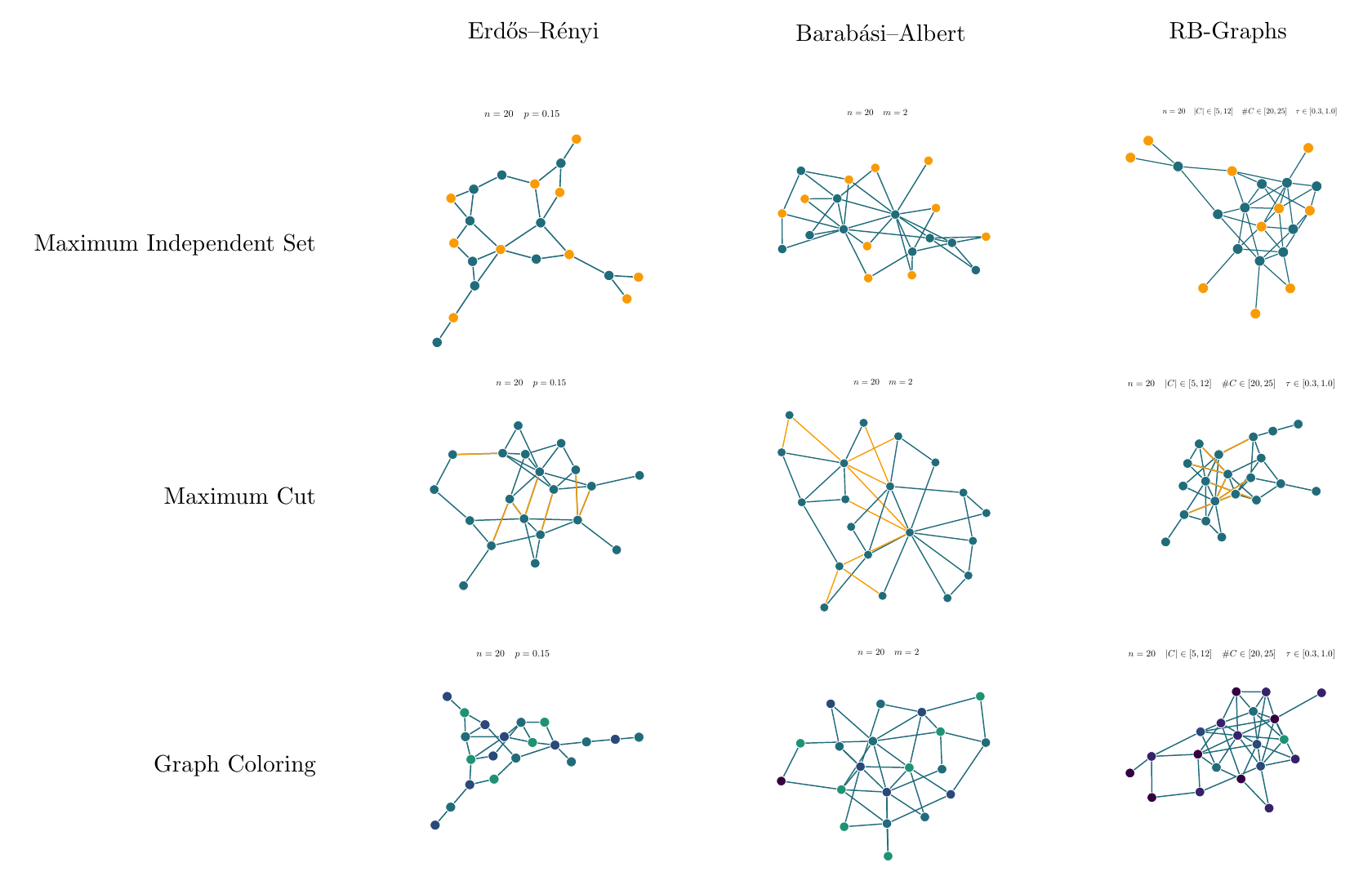}
	\caption{Down-scaled sample graphs and solutions across all problems and graph generators for the \emph{Combinatorial Optimization} datasets with our chosen parameters.
		\label{fig:combinatorial_optimzation}}
\end{figure}

\paragraph{Related work}~
Many recent learning-based optimization work rely on synthetic data \citep{toenshoff2021graph,tonshoff2022one,zhang2023let,karalias2021erdos}, either RB graphs~\citep{rb_graphs}, Erd\H{o}s-R\'enyi (ER) graphs~\citep{erd6s1960evolution}, or Barab\'asi-Albert (BA) graphs~\citep{albert2002statistical}. Some also utilize pre-existing datasets from earlier graph benchmarks, e.g., \citep{karalias2021erdos} uses TUDatasets \citep{Mor+2020}, and \citet{li2022rethinking} uses Planetoid datasets \citep{yang2016revisiting}. Earlier efforts have also produced benchmarks for specific graph CO problems. For graph coloring, examples include hard 3-colorability datasets featuring cliques \citep{mizuno2007toward,mizuno2008constructive}, queen graphs \citep{Chv04}, Latin square graphs \citep{gomes2002completing}, $K$-Insertions graphs \citep{CarOlm02a}, and $K$-FullIns graphs \citep{CarOlm02b}. The DIMACS benchmark collection \citep{dimacs} provides hundreds of real-world networks and problem instances across various CO domains. Still, their heterogeneity and the limited number of instances per category make them less well-suited to modern machine learning pipelines. More recently, \citet{nath2024maxcutbench} introduced a max-cut benchmark covering a variety of graph classes.
\citet{graip} also introduces datasets for CO, but frames it as an inverse problem.
Our motivation differs from these prior works.
We aim to design a unified set of graph datasets shared across multiple CO tasks, enabling direct comparison between methods in a consistent experimental setting. While some prior work generates random instances for individual problems, we standardize the process by fixing both the generation procedures and the training-validation splits, ensuring that all methods compare on the same instances.

\paragraph{Definitions of CO problems}~
Given a graph $G=(V(G),E(G))$, the MIS problem is to find the largest subset of non-adjacent nodes, formally,
\begin{itemize}
	\item $\Omega(I) \coloneqq V(G)$
	\item $F(I) \coloneqq \{ S \subseteq V(G) \mid \forall\, u,v \in S,\, (u,v) \notin E(G) \}$
	\item $w \colon V(G) \to \Rb$, in the unweighted case $w(v) \coloneqq -1$.
\end{itemize}
The objective is to find $S^*_I \in F(I)$ minimizing $c(S) \coloneqq \sum_{v \in S} w(v)$.

The max-cut problem seeks a 2-way partition of the node set such that the total weight of edges across the cut is maximized, i.e.,
\begin{itemize}
	\item $\Omega(I) \coloneqq E(G)$,
	\item $F(I) \coloneqq \{ S \subseteq E(G) \mid \exists\, V'(G) \subset V(G), \forall (u,v) \in S \colon u \in V(G), v \in V(G) \setminus V'(G) \}$,
	\item $w \colon E(G) \to \Rb$, in the unweighted case $w(e) \coloneqq -1.$
\end{itemize}
The objective is to find $S^*_I \in F(I)$ minimizing $c(S) \coloneqq \sum_{e \in S} w(e)$.

Graph coloring is the problem of assigning at most $k$ colors to the nodes such that adjacent nodes receive different colors. Inspired by the assignment-based model \citep{jabrayilov2018new}, it can be formulated in search of a coloring function $f \colon V(G) \to [k]$, such that
\begin{itemize}
	\item $\Omega(I) \coloneqq [k]$,
	\item $F(I) \coloneqq \{ S \subseteq [k] \mid \forall\, (u,v) \in E(G) \colon f(v) \neq f(u) \land \forall\, v \in V(G) \colon f(v) \in S \}$,
	\item $w \colon \Omega(I) \to \{1\}$.
\end{itemize}
The objective is to find the best $S^*_I \in F(I)$, such that the minimum number of colors is used, $c(S) \coloneqq \sum_{s \in S} w(s)$.
\autoref{fig:combinatorial_optimzation} shows examples of all three CO problems across the three graph generators we use.

\paragraph{Graph generation parameters}~
We consider three well-established random graph models: RB, ER, and BA graphs. These models are widely used in machine learning to solve CO problems. The diversity and difficulty of each dataset are controlled via graph generation parameters. Specifically,
\begin{itemize}
	\item for RB graphs, we specify integers $n$ (number of cliques), $k$ (nodes per clique), and a float $p$ (constraint tightness). The total number of nodes is $kn$, and we discard graphs whose size falls outside a predefined range.
	\item For BA graphs, we control the number of nodes and the parameter $m$, which determines how many edges are attached from each new node.
	\item For ER graphs, we use the number of nodes and the edge probability $p$, which controls the probability that an edge exists between any given pair of nodes.
\end{itemize}
We generate 50\,000 instances per dataset and provide both small- and large-scale variants.
\autoref{tab:co_stats} shows the parameters we used to generate the graphs in our datasets.
For RB graphs, parameters are taken from \cite{zhang2023let, sanokowski2024diffusion, wenkel2024towards, sun2022annealed}. Parameters for large ER graphs are from \cite{bother2022what, qiu2022dimes, sun2023difusco, yu2024disco}. Parameters for large BA graphs are from \cite{bother2022what}. The parameters for ER-small and BA-small are adjusted to match RB-small's graph size.
When a parameter is specified as a range, it is sampled uniformly at random for each instance.
The only exception to this is the number of nodes in RB graphs, where rejection sampling is used to ensure that the graph size falls inside a specified range, as described in \Cref{sec:co}.
We provide datasets in both \textsc{PyTorch Geometric} format \citep{fey2019fast,fey2025pyg} and \textsc{NetworkX} format \citep{hagberg2020networkx}.

\paragraph{Heuristic solutions for supervised learning}~
We use KaMIS \citep{lamm2016finding} to calculate heuristic MIS solutions.
The solution to max-cut is obtained by formulating it as an integer programming problem and solving it with \cite{gurobi}, with a timeout of 3600 seconds.
We use GCol \citep{lewis2025gcol} for a heuristic graph chromatic number, with a maximum of $10^8$ search iterations.

\begin{table*}
	\centering
	\caption{Parameters for CO problems generation.}
	\begin{tabular}{cccccccc}
		\toprule
		Generator           & Size  & Nodes       & Nodes/Clique & Cliques  & Tightness & Edge Prob. & Attached Edges \\
		\midrule
		\multirow{2}{*}{RB} & small & [200, 300]  & [5, 12]      & [20, 25] & [0.3, 1]  & -          & -              \\
		                    & large & [800, 1200] & [20, 25]     & [40, 55] & [0.3, 1]  & -          & -              \\
		\midrule
		\multirow{2}{*}{ER} & small & [200, 300]  & -            & -        & -         & 0.15       & -              \\
		                    & large & [700, 800]  & -            & -        & -         & 0.15       & -              \\
		\midrule
		\multirow{2}{*}{BA} & small & [200, 300]  & -            & -        & -         & -          & 2              \\
		                    & large & [700, 800]  & -            & -        & -         & -          & 2              \\
		\bottomrule
	\end{tabular}
	\label{tab:co_stats}
\end{table*}

\paragraph{Model architecture}~
Our proposed model uses the encoder-decoder architecture. The encoder, an MLP, processes node features derived from RWSE~\citep{Dwivedi2022RWSE} and node degrees. The processor module is the only part that differs between baselines and performs the bulk of the calculations. It is followed by a decoder that generates node representations. For the supervised setting, we apply summation aggregation to obtain a scalar graph-level prediction as the final output. For the unsupervised setting, we leave the node-level scores as the output.

\paragraph{Extended results}~
\autoref{tab:co_results_unsupervised} shows the baseline results for unsupervised CO across all three CO problems.
We report the average objective value of solutions obtained by running the decoder on the model's output scores for test-set problem instances.
Out of the four baselines, GIN performs best across all three CO problems, with some exceptions.
As expected, the graph-based models GIN and GT generally perform better than MLP and DeepSets.
One exception is graph coloring on the (very sparse) BA graphs, where DeepSets outperforms the other two models across all graph sizes.
Note that our baselines do not use diffusion models or other advanced architectures or techniques explicitly tailored for CO.
Their performance is therefore much weaker than that of methods that do use them, e.g.\@ \citet{sanokowski2024diffusion, zhang2023let}. Also note that most learning-based methods currently cannot compete with exact solvers and some hand-crafted heuristics in terms of solution quality, but are often much quicker \citep{sun2023difusco, sanokowski2024diffusion, zhang2023let}.

\begin{table}[htbp]
	\tabledefaults
	\centering
	\caption{\textbf{Combinatorial Optimization.} Results for each CO dataset with unsupervised learning.}
	\label{tab:co_results_unsupervised}
	\resizebox{0.6\textwidth}{!}{%
		\begin{tabular}{lllcc}
			\toprule
			                 &                               &                         & \multicolumn{2}{c}{\textbf{Size}}                           \\
			\cmidrule(lr){4-5}
			\textbf{Problem} & \textbf{Dataset}              & \textbf{Method}         & Small                             & Large                   \\
			\midrule
			\multirow{12}{*}{\shortstack{\vspace{1.5ex}                                                                                              \\ MIS \\ (MIS size $\uparrow$})}
			                 & \multirow{5}{*}{\dataset{RB}}
			                 & GIN                           & \meanstd{17.294}{0.328} & \meanstd{13.999}{0.321}                                     \\
			                 &                               & GT                      & \meanstd{16.542}{0.477}           & \meanstd{13.406}{0.140} \\
			                 &                               & MLP                     & \meanstd{16.105}{0.097}           & \meanstd{13.040}{0.214} \\
			                 &                               & DeepSets                & \meanstd{16.021}{0.032}           & \meanstd{13.183}{0.035} \\
			                 &                               & Solver                  & \meanstd{20.803}{1.817}           & \meanstd{42.547}{4.449} \\
			\cmidrule(l){2-5}
			                 & \multirow{5}{*}{\dataset{ER}}
			                 & GIN                           & \meanstd{25.418}{0.407} & \meanstd{26.276}{0.408}                                     \\
			                 &                               & GT                      & \meanstd{22.984}{0.473}           & \meanstd{24.980}{0.292} \\
			                 &                               & MLP                     & \meanstd{23.183}{0.016}           & \meanstd{24.259}{0.449} \\
			                 &                               & DeepSets                & \meanstd{23.050}{0.061}           & \meanstd{24.220}{0.056} \\
			                 &                               & Solver                  & \meanstd{33.604}{1.428}           & \meanstd{45.637}{0.631} \\
			\cmidrule(l){2-5}
			                 & \multirow{5}{*}{\dataset{BA}}
			                 & GIN                           & \meanstd{100.16}{3.674} & \meanstd{135.00}{0.720}                                     \\
			                 &                               & GT                      & \meanstd{99.579}{6.448}           & \meanstd{114.26}{0.601} \\
			                 &                               & MLP                     & \meanstd{95.108}{2.042}           & \meanstd{114.49}{0.758} \\
			                 &                               & DeepSets                & \meanstd{95.076}{0.173}           & \meanstd{114.89}{0.016} \\
			                 &                               & Solver                  & \meanstd{142.86}{16.54}           & \meanstd{433.77}{19.17} \\
			\midrule
			\multirow{12}{*}{\shortstack{\vspace{1.5ex}                                                                                              \\ Maximum Cut \\ (maximum \\ cut size $\uparrow$)}}
			                 & \multirow{5}{*}{\dataset{RB}}
			                 & GIN                           & \meanstd{2106.7}{14.62} & \meanstd{24748.}{87.76}                                     \\
			                 &                               & GT                      & \meanstd{1925.7}{32.75}           & \meanstd{21524.}{184.0} \\
			                 &                               & MLP                     & \meanstd{1727.7}{165.1}           & \meanstd{20357.}{249.6} \\
			                 &                               & DeepSets                & \meanstd{140.02}{155.5}           & \meanstd{3575.9}{730.0} \\
			                 &                               & Solver                  & \meanstd{2920.1}{97.23}           & \meanstd{33914.}{7861.} \\
			\cmidrule(l){2-5}
			                 & \multirow{5}{*}{\dataset{ER}}
			                 & GIN                           & \meanstd{2327.9}{24.78} & \meanstd{20878.}{107.9}                                     \\
			                 &                               & GT                      & \meanstd{2172.7}{91.75}           & \meanstd{16534.}{278.0} \\
			                 &                               & MLP                     & \meanstd{1866.7}{67.64}           & \meanstd{7335.4}{57.49} \\
			                 &                               & DeepSets                & \meanstd{33.634}{20.84}           & \meanstd{27.663}{6.763} \\
			                 &                               & Solver                  & \meanstd{2835.5}{607.6}           & \meanstd{23884.}{1809.} \\
			\cmidrule(l){2-5}
			                 & \multirow{5}{*}{\dataset{BA}}
			                 & GIN                           & \meanstd{397.00}{0.605} & \meanstd{1044.1}{0.649}                                     \\
			                 &                               & GT                      & \meanstd{363.76}{0.639}           & \meanstd{986.93}{3.128} \\
			                 &                               & MLP                     & \meanstd{308.73}{0.224}           & \meanstd{929.20}{4.060} \\
			                 &                               & DeepSets                & \meanstd{1.0669}{0.800}           & \meanstd{154.31}{151.5} \\
			                 &                               & Solver                  & \meanstd{460.91}{50.13}           & \meanstd{1260.4}{48.81} \\
			\midrule
			\multirow{12}{*}{\shortstack{\vspace{1.5ex}                                                                                              \\ Graph Coloring \\ (number of \\ colors used $\downarrow$)}}
			                 & \multirow{5}{*}{\dataset{RB}}
			                 & GIN                           & \meanstd{25.166}{0.288} & \meanstd{55.513}{0.526}                                     \\
			                 &                               & GT                      & \meanstd{25.146}{0.253}           & \meanstd{55.562}{0.648} \\
			                 &                               & MLP                     & \meanstd{24.733}{0.667}           & \meanstd{55.558}{0.557} \\
			                 &                               & DeepSets                & \meanstd{26.723}{0.189}           & \meanstd{71.051}{0.604} \\
			                 &                               & Solver                  & \meanstd{19.970}{3.465}           & \meanstd{41.480}{6.634} \\
			\cmidrule(l){2-5}
			                 & \multirow{5}{*}{\dataset{ER}}
			                 & GIN                           & \meanstd{16.182}{0.202} & \meanstd{34.587}{0.545}                                     \\
			                 &                               & GT                      & \meanstd{16.188}{0.201}           & \meanstd{34.385}{0.413} \\
			                 &                               & MLP                     & \meanstd{17.110}{0.144}           & \meanstd{34.658}{0.394} \\
			                 &                               & DeepSets                & \meanstd{18.266}{0.018}           & \meanstd{55.345}{0.551} \\
			                 &                               & Solver                  & \meanstd{10.235}{0.836}           & \meanstd{22.933}{0.772} \\
			\cmidrule(l){2-5}
			                 & \multirow{5}{*}{\dataset{BA}}
			                 & GIN                           & \meanstd{5.1318}{0.114} & \meanstd{6.2028}{0.283}                                     \\
			                 &                               & GT                      & \meanstd{5.0939}{0.070}           & \meanstd{6.1167}{0.086} \\
			                 &                               & MLP                     & \meanstd{5.9900}{0.127}           & \meanstd{9.4215}{0.186} \\
			                 &                               & DeepSets                & \meanstd{3.2780}{0.122}           & \meanstd{3.1981}{0.239} \\
			                 &                               & Solver                  & \meanstd{3.0000}{0.000}           & \meanstd{3.0000}{0.000} \\
			\bottomrule
		\end{tabular}}
\end{table}

\subsubsection{Algorithmic reasoning: Learning to simulate algorithms}
\label{sec:algoreaso_details}

\begin{figure}[t]
	\centering
	\includegraphics[width=1\textwidth]{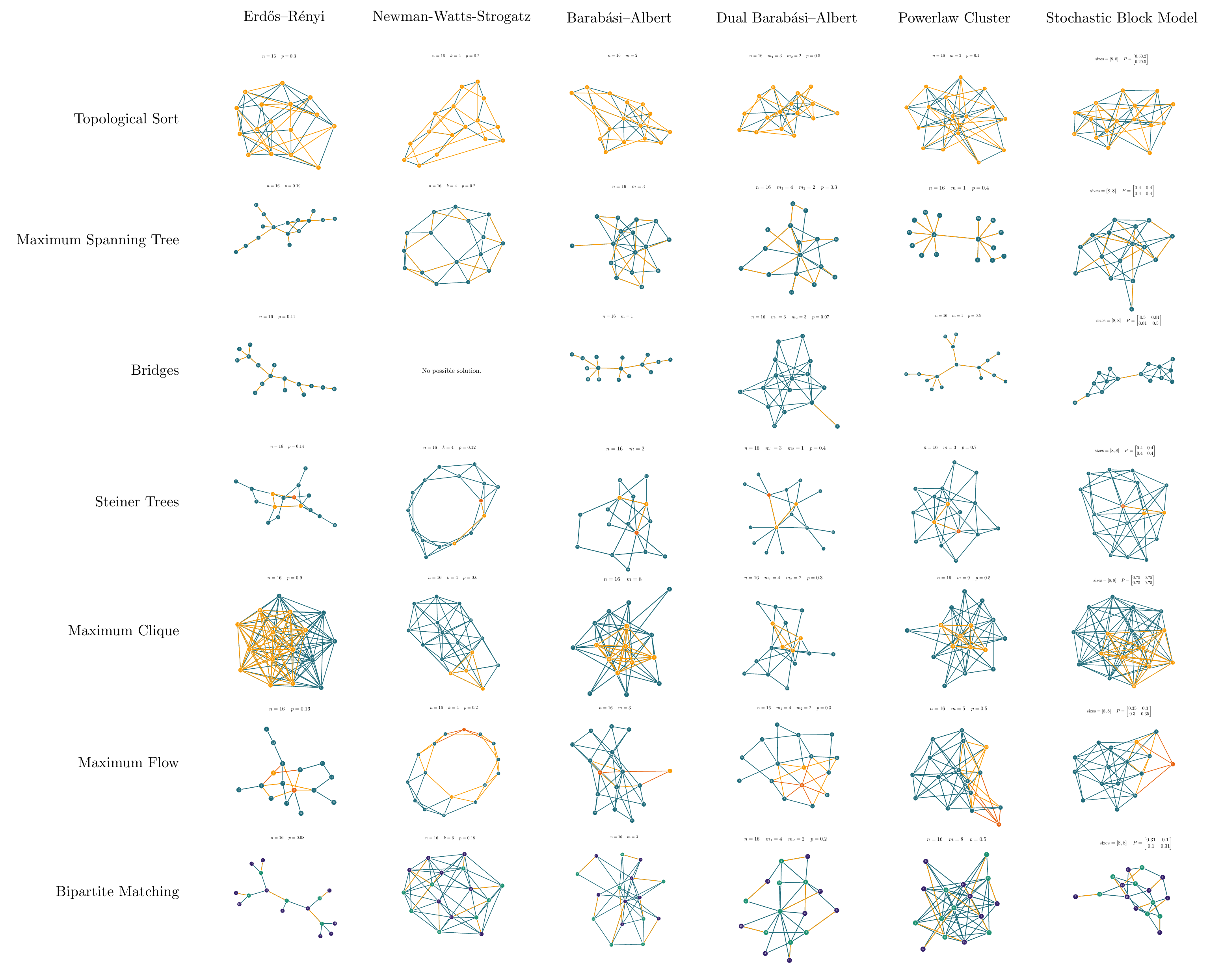}
	\caption{Example graphs and solutions across all graph generators and all performed tasks for the \emph{Algorithmic Reasoning} datasets with our chosen parameters.
		\label{fig:algorithmic_reasoning}}
\end{figure}

In addition to SAT, many real-world tasks depend on the efficient solution of graph algorithms. Given the active exploration of the intersection of algorithms and neural networks across domains~\citep{Esterman2024PUZZLES,Fan2024NPHardEval,Kaiser2015NeuralGPUs,Zaremba2014Learning}, our goal is to provide dedicated datasets for graph algorithmic reasoning. Prior work~\citep{Velickovic2020NeuralExecution,Xu2020nnreasonabout,Bounsi2024TransfmeetNeuralAlgoReas} and benchmarks such as CLRS~\citep{Velickovic2022CLRSbenchmark} and SALSA-CLRS ~\citep{minder2023salsaclrs} have demonstrated that GNN architectures can achieve strong performance on graph problems when provided with hints and graph invariants---tasks that require substantial theoretical expressiveness from graph learning models~\citep{Arvind2020WLsubgraph}. To broaden this research area, we contribute large-scale datasets for several common graph algorithm tasks, expanding the landscape of neural algorithmic reasoning benchmarks.
\cref{fig:algorithmic_reasoning} provides an overview of our graph generators, the tasks we perform, and the parameters we chose for generation.

\paragraph{Related work}~
Algorithmic reasoning has been adopted for various architectures ~\citep{Li2024OpenBook, Velickovic2020PointerGraphNetworks}. Apart from neural networks ~\citep{Diao2023RelationalAttn, Mahdavi2023TowardsBetterOOD, Bounsi2024TransfmeetNeuralAlgoReas, Rodionov2023AlgoReasoIntSup} and reinforcement learning approaches ~\citep{Esterman2024PUZZLES}, ~\cite{Velickovic2022CLRSbenchmark} propose the CLRS benchmark for a variety of algorithmic tasks, derived from the \textit{Introduction to Algorithms} textbook ~\citep{Cormen2009IntroAlgorithms}. This benchmark comprises 30 algorithmic problems, each accompanied by hints to guide the solution. However, this benchmark excludes regression tasks and limits learning tasks to specific algorithms. Furthermore, follow-up works investigated the importance of hints given in CLRS ~\citep{Rodionov2023AlgoReasoIntSup} and proposed a variety of architectures for solving algorithmic tasks~\citep{Bevilacqua2023NeuralAlgoReas, Ibarz2022GeneralistAlgoLearner, Li2024OpenBook}.

\paragraph{Description of learning task}~
We consider a collection of seven algorithmic reasoning tasks, inspired by the task collection of \citet{stoll2025generalizable} and selected from classical graph problems studied in algorithmics. These tasks include computing the topological sorting ~\citep{KahnTopoSorting}, identifying bridges in a graph, computing the minimum spanning tree~\citep{kruskal1956shortest}, determining the maximum flow~\citep{10.1145/12130.12144}, finding the maximum clique, computing Steiner trees~\citep{kou1981fast}, and solving bipartite matching~\citep{hopcroft1973n}. Our task set spans different output granularities: node-level tasks (topological sorting, bipartite matching, max clique), edge-level tasks (bridges, minimum spanning tree, Steiner tree), and a graph-level task (max flow). To introduce additional modeling challenges, we provide three levels of difficulty per task, representing variations in the underlying graph distributions. Each task is formulated as either a binary classification or a regression problem. Given a set of graphs $\mathcal{G}_A$, an algorithmic reasoning task $A$, its ground-truth solution function $S_A$, and a model $m$, the goal is to train
\begin{equation*}
	m \colon \mathcal{G}_A \rightarrow \mathbb{R} \quad \text{or} \quad m \colon \mathcal{G}_A \rightarrow [0, 1],
\end{equation*}
depending on the task. The objective is to minimize the MAE between the model’s predictions and the ground-truth solution,
\begin{equation*}
	\frac{1}{|\mathcal{G}_A|} \sum_{G \in \mathcal{G}_A} \left| m(G) - S_A(G) \right|.
\end{equation*}
For binary classification tasks, we instead report the F1 score computed from the objective to maximize the accuracy between predictions and ground truth,
\begin{equation*}
	\frac{1}{|\mathcal{G}_A|} \sum_{G \in \mathcal{G}_A} \mathds{1} (m(G) = S_A(G))
\end{equation*}
where $\mathds{1}$ denotes the indicator function.

\paragraph{Details on the datasets}~
For all algorithmic reasoning tasks, we work with sets of synthetically generated graphs. We begin by sampling graphs using a variety of graph generators, including Erdős–Rényi (ER) graphs ~\citep{erd6s1960evolution}, stochastic block model (SBM) graphs ~\citep{holland1983stochastic}, power-law cluster (PC) graphs~\citep{holme2002growing}, Newman–Watts–Strogatz (NWS) graphs~\citep{newman1999renormalization}, Barabási–Albert (BA) graphs ~\citep{barabasi1999emergence}, and dual Barabási–Albert (DBA)  graphs~\citep{moshiri2018dualbarabasi}. During sampling, we ensure that each resulting graph is unique; otherwise, it is resampled. For each task, we choose generator parameters shown in \Cref{table:algres_params} to yield meaningful distributions of task-specific properties. We further introduce a parameter for component connections, which determines the number of random edges between disconnected components in a graph. This allows us to sample graphs with low edge probabilities without disconnected components. Additionally, based on the difficulty level, we adjust the probability distribution over the selected graph generators to further control the complexity of generated instances, as seen in \Cref{table:AlgoReasDifficulties}. For minimum spanning tree computations, we also adjust the graph generator parameters, as indicated in \Cref{table:algres_params}, to require stronger generalization capabilities, as the same parameters proved too easy to generalize from. Ground-truth labels are computed using the \textsc{NetworkX} library~\citep{hagberg2020networkx}, which provides reference implementations of most of the required graph algorithms.
\begin{table}[htbp]
	\tabledefaults
	\centering
	\caption{Graph generation parameters for algorithmic reasoning tasks. The cc parameter denotes the number of connections between disconnected components in Erdős–Rényi graphs.}
	\label{table:algres_params}
	\resizebox{\textwidth}{!}{
		\begin{tabular}{llcccccccc}
			\toprule
			\textbf{Generator} & \textbf{Param.} & Top. Sort         & MST               & MST (shift)        & Bridges            & Steiner Trees     & Max. Clique        & Flow               & Max. Matching \\
			\midrule
			\multirow{2}{*}{ER}
			                   & p               & 0.3               & 0.19              & 0.17               & 0.11               & 0.14              & 0.9                & 0.16               & 0.08          \\
			                   & cc              & 1                 & 1                 & 1                  & 1                  & 1                 & 1                  & 1                  & 1             \\
			\midrule
			\multirow{2}{*}{NWS}
			                   & k               & 2                 & 4                 & 2                  & -                  & 4                 & 4                  & 4                  & 6             \\
			                   & p               & 0.2               & 0.2               & 0.15               & -                  & 0.12              & 0.6                & 0.2                & 0.18          \\
			\midrule
			\multirow{1}{*}{BA}
			                   & m               & 2                 & 3                 & 2                  & 1                  & 2                 & 8                  & 3                  & 3             \\
			\midrule
			\multirow{3}{*}{DBA}
			                   & $\text{m}_1$    & 3                 & 4                 & 2                  & 3                  & 3                 & 4                  & 4                  & 4             \\
			                   & $\text{m}_2$    & 2                 & 2                 & 1                  & 1                  & 1                 & 2                  & 2                  & 2             \\
			                   & p               & 0.5               & 0.3               & 0.05               & 0.07               & 0.4               & 0.3                & 0.3                & 0.2           \\
			\midrule
			\multirow{2}{*}{PC}
			                   & p               & 0.1               & 0.4               & 0.35               & 0.5                & 0.7               & 0.5                & 0.5                & 0.5           \\
			                   & m               & 3                 & 1                 & 5                  & 1                  & 3                 & 9                  & 5                  & 8             \\
			\midrule
			\multirow{2}{*}{SBM}
			                   & p. mat.         & \(\begin{bmatrix}
				                                         $0.5$ & $0.2$ \\
				                                         $0.2$ & $0.5$ \\
			                                         \end{bmatrix}\) & \(\begin{bmatrix}
				                                                             $0.4$ & $0.4$ \\
				                                                             $0.4$ & $0.4$ \\
			                                                             \end{bmatrix}\) & \(\begin{bmatrix}
				                                                                                 $0.31$ & $0.01$ \\
				                                                                                 $0.01$ & $0.31$ \\
			                                                                                 \end{bmatrix}\) & \(\begin{bmatrix}
				                                                                                                     $0.5$  & $0.01$ \\
				                                                                                                     $0.01$ & $0.5$  \\
			                                                                                                     \end{bmatrix}\) & \(\begin{bmatrix}
				                                                                                                                         $0.4$ & $0.4$ \\
				                                                                                                                         $0.4$ & $0.4$ \\
			                                                                                                                         \end{bmatrix}\) & \(\begin{bmatrix}
				                                                                                                                                             $0.75$ & $0.75$ \\
				                                                                                                                                             $0.75$ & $0.75$ \\
			                                                                                                                                             \end{bmatrix}\) & \(\begin{bmatrix}
				                                                                                                                                                                 $0.35$ & $0.3$  \\
				                                                                                                                                                                 $0.3$  & $0.35$ \\
			                                                                                                                                                                 \end{bmatrix}\) & \(\begin{bmatrix}
				                                                                                                                                                                                     $0.31$ & $0.1$  \\
				                                                                                                                                                                                     $0.1$  & $0.31$ \\
			                                                                                                                                                                                     \end{bmatrix}\) \\
			                   & sizes           & 1/2, 1/2          & 1/2, 1/2          & 1/2, 1/2           & 1/2, 1/2           & 1/2, 1/2          & 1/2, 1/2           & 1/2, 1/2           & 1/2, 1/2      \\
			\bottomrule
		\end{tabular}}
\end{table}

\begin{table}[h]
	\tabledefaults
	\centering
	\caption{Graph generation parameters for algorithmic reasoning tasks.}
	\label{table:AlgoReasDifficulties}
	\resizebox{0.6\textwidth}{!}{
		\begin{tabular}{llcccccc}
			\toprule
			\textbf{Split} & \textbf{Difficulty} & ER  & PC  & NWS & BA  & DBA & SBM \\
			\midrule
			\multirow{3}{*}{Train}
			               & \dataset{Easy}      & 1/3 & 0   & 1/3 & 0   & 1/3 & 0   \\
			               & \dataset{Medium}    & 1   & 0   & 0   & 0   & 0   & 0   \\
			               & \dataset{Hard}      & 1   & 0   & 0   & 0   & 0   & 0   \\
			\midrule
			\multirow{3}{*}{Valid/Test}
			               & \dataset{Easy}      & 1/6 & 1/6 & 1/6 & 1/6 & 1/6 & 1/6 \\
			               & \dataset{Medium}    & 1/6 & 1/6 & 1/6 & 1/6 & 1/6 & 1/6 \\
			               & \dataset{Hard}      & 0   & 1/5 & 1/5 & 1/5 & 1/5 & 1/5 \\
			\bottomrule
		\end{tabular}}
\end{table}

For edge-level tasks, we also generate an edge-level representation to support models that require tokenized input. Each task includes one million generated training graphs and 10\,000 validation and test graphs. To assess generalization beyond function approximation, we fix the number of nodes in the training graphs to $16$ and set the validation and test graphs to have $128$ nodes.

To further study generalization, we introduce size-generalization tasks, in which pre-trained models are evaluated on graphs with sizes ranging from 128 to 512 nodes. We exclude max-flow computation from size generalization, as the MAE scales with the number of nodes in the flow network and is therefore not indicative of size generalization. This setup enables an empirical analysis of how well models extrapolate to unseen graph sizes. Except for the topological sorting, Steiner tree, and bipartite matching tasks, no node features are given. For the minimum spanning tree, max flow, and Steiner tree tasks, corresponding edge weights are included as edge attributes.

\paragraph{Dataset challenges}~
Across all algorithmic reasoning datasets, we distinguish training and test sets along two axes of data distribution: graph size and graph-generation sampling distributions. Drawing inspiration from CLRS \cite{Velickovic2022CLRSbenchmark} and SALSA-CLRS \cite{minder2023salsaclrs}, as well as the concept of scaffolding splits from molecular chemistry, we provide these two axes splits in order to test for OOD concepts. Since prior work \cite{minder2023salsaclrs} indicates potential size generalization capabilities in graph neural networks trained to simulate graph algorithms, we specifically test for these capabilities by increasing the graph size by a factor of 8 between the training and test sets. Furthermore, distributions are significantly shifted between training and test sets, as evidenced by three difficulty levels ranked by the magnitude of changes in the generation distribution.

\paragraph{Experimental evaluation and baselines}~
We use the encoder-processor-decoder pipeline described in \Cref{Section:ExpSetup}. For node and graph-level tasks, the pipeline is applied directly with the task-specific encoder described below. Since the graph transformer baseline expects node-level tokenization, we provide an additional graph transform.

Since most of our tasks provide only node or edge features, we use learned vectors, as described in the encoder, for the remaining embeddings. Across all datasets, we employ one-layer linear embeddings for both integer and real-valued features. Depending on the node-, edge-, or graph-level tasks, the decoder applies linear layers that represent the classification or regression target. In the case of bridges, minimum spanning tree, Steiner trees, max clique and bipartite matching a binary classification target is used for node or edge level predictions. For flow and topological sorting, a one-dimensional regression target is used instead. For pretrained models used in size generalization experiments, we use the model directly as encoder, processor, and decoder. However, the input graph size is increased.
In contrast to complete training examples, we use a few-shot scenario with only 1\,000 graphs to enable fast inference. In addition, we do not consider the flow task here because it is a regression setting. Throughout all experiments, we use the same random seed for a pretrained model to enable comparison to the 128-node test set.

\paragraph{Results}\label{AlgoReaso:Results}
We report results on all datasets for each of the three selected difficulties. These are denoted using the terms \dataset{easy}, \dataset{medium}, and \dataset{hard} to reflect the additional generalization requirements introduced by out-of-distribution sampling. For classification tasks, we report F1 scores, and for regression tasks, we report MAE. All results are grouped by their respective estimated task difficulty in \Cref{table:algo_reas_results} and \Cref{table:algo_reas_results_2}  below.
Additionally, we report size generalization results using pre-trained baselines from the \dataset{medium} setting of our training procedure. For this, we evaluate each pre-trained model on graphs with 128 to 512 nodes. Results are provided in \Cref{table:algo_reas_sizegenresults}. For the minimum spanning tree, max clique, and max matching datasets, the graph transformer baseline shows improved performance. However, on the max flow, bridges, and Steiner tree datasets, the GIN baseline model performs better across the proposed difficulty levels. Furthermore, we observe a reduction in F1 scores as difficulty increases across datasets, highlighting variation in generator selection in the training data. However, the differences are not as pronounced as expected for the bridges dataset. We further observe that the GIN baseline is inconsistent across seeds for the minimum spanning tree task, whereas the graph transformer baseline does not exhibit similar issues.
Additionally the GNNPlus baselines provide improved performance over the GIN baseline model in minimum spanning tree, steiner trees, maximum matching and max clique. However, we note that this advantage does not appear across all selected tasks with singular tasks such as topological sorting providing worse results than GIN or GT baselines. We further observe that there is no clear overall best performing architecture over all tasks, with results indicating that the architecture selection is depending on the algorithmic reasoning task.
As an additional result, differences between difficulty levels are not as pronounced as expected from the data distribution shift. In most tasks increasing the difficulty level only reduces F1 scores marginally. Flow is the exception as a task here as MAE values increase significantly with an increasing distribution shift.

Overall, the performance of all baselines is lowest on Steiner trees and max clique tasks, indicating that these tasks are more complex for graph learning baselines. Similar to the results obtained in the CLRS benchmark \citep{Velickovic2022CLRSbenchmark} and by \citet{2024muellertowards} for an expressive graph transformer architecture, MST, bridges, and topological sorting provide more manageable tasks for graph learning baselines. We note that, despite previous benchmarks such as CLRS, there are currently no other baseline results on the synthetic algorithmic reasoning datasets we provide.

In the case of size generalization, we observe that both baselines are relatively robust to scaling of the test graph size. However, we observe that the generalization capability with respect to graph size is task-dependent, with the minimum spanning tree and Steiner tree tasks improving with larger graphs. Nonetheless, we note that this behavior may be observed due to shifts in average degree and node connectivity, using the same parameters as in test set generation, since each selected graph generator scales with the number of nodes.
Nonetheless, performance is decreasing for topological sorting, max matching, and max clique tasks, highlighting the inability of our selected baselines to be size-generalization invariant across all tasks.

\begin{table}[htbp]
	\tabledefaults
	\centering
	\caption{Resulting F1 scores for algorithmic reasoning datasets, including minimum spanning tree, bridge finding and Steiner tree computation.}
	\label{table:algo_reas_results}
	\begin{tabular}{lllc}
		\toprule
		\textbf{Task} & \textbf{Difficulty}\hspace{-0.1em} & \textbf{Model}      & F1                       \\
		\midrule
		\multirow{15}{*}{MST}
		              & \multirow{5}{*}{\dataset{easy}}    & \dataset{GIN}       & \meanstd{0.6906}{0.1655} \\
		              &                                    & \dataset{GT}        & \meanstd{0.8566}{0.0068} \\
		              &                                    & \dataset{GIN+}      & \meanstd{0.8225}{0.0020} \\
		              &                                    & \dataset{GCN+}      & \meanstd{0.8310}{0.0028} \\
		              &                                    & \dataset{GatedGCN+} & \meanstd{0.8485}{0.0220} \\
		\cmidrule(l){2-4}
		              & \multirow{5}{*}{\dataset{medium}}  & \dataset{GIN}       & \meanstd{0.7288}{0.0894} \\
		              &                                    & \dataset{GT}        & \meanstd{0.8504}{0.0148} \\
		              &                                    & \dataset{GIN+}      & \meanstd{0.7802}{0.0286} \\
		              &                                    & \dataset{GCN+}      & \meanstd{0.5317}{0.0318} \\
		              &                                    & \dataset{GatedGCN+} & \meanstd{0.8068}{0.0315} \\
		\cmidrule(l){2-4}
		              & \multirow{5}{*}{\dataset{hard}}    & \dataset{GIN}       & \meanstd{0.6107}{0.2015} \\
		              &                                    & \dataset{GT}        & \meanstd{0.8421}{0.0115} \\
		              &                                    & \dataset{GIN+}      & \meanstd{0.8449}{0.0067} \\
		              &                                    & \dataset{GCN+}      & \meanstd{0.8554}{0.0098} \\
		              &                                    & \dataset{GatedGCN+} & \meanstd{0.8421}{0.0069} \\
		\midrule
		\multirow{15}{*}{Bridges}
		              & \multirow{5}{*}{\dataset{easy}}    & \dataset{GIN}       & \meanstd{0.9831}{0.0184} \\
		              &                                    & \dataset{GT}        & \meanstd{0.9269}{0.0103} \\
		              &                                    & \dataset{GIN+}      & \meanstd{0.9141}{0.0363} \\
		              &                                    & \dataset{GCN+}      & \meanstd{0.9221}{0.0151} \\
		              &                                    & \dataset{GatedGCN+} & \meanstd{0.9339}{0.0056} \\
		\cmidrule(l){2-4}
		              & \multirow{5}{*}{\dataset{medium}}  & \dataset{GIN}       & \meanstd{0.9622}{0.0077} \\
		              &                                    & \dataset{GT}        & \meanstd{0.8762}{0.0258} \\
		              &                                    & \dataset{GIN+}      & \meanstd{0.8288}{0.0363} \\
		              &                                    & \dataset{GCN+}      & \meanstd{0.9199}{0.0465} \\
		              &                                    & \dataset{GatedGCN+} & \meanstd{0.7007}{0.0930} \\
		\cmidrule(l){2-4}
		              & \multirow{5}{*}{\dataset{hard}}    & \dataset{GIN}       & \meanstd{0.968}{0.0178}  \\
		              &                                    & \dataset{GT}        & \meanstd{0.8897}{0.0304} \\
		              &                                    & \dataset{GIN+}      & \meanstd{0.9090}{0.0562} \\
		              &                                    & \dataset{GCN+}      & \meanstd{0.8893}{0.0565} \\
		              &                                    & \dataset{GatedGCN+} & \meanstd{0.6921}{0.0153} \\
		\midrule
		\multirow{15}{*}{Steiner Trees}
		              & \multirow{5}{*}{\dataset{easy}}    & \dataset{GIN}       & \meanstd{0.6691}{0.0288} \\
		              &                                    & \dataset{GT}        & \meanstd{0.6691}{0.0112} \\
		              &                                    & \dataset{GIN+}      & \meanstd{0.6656}{0.0111} \\
		              &                                    & \dataset{GCN+}      & \meanstd{0.7073}{0.0021} \\
		              &                                    & \dataset{GatedGCN+} & \meanstd{0.6589}{0.0130} \\
		\cmidrule(l){2-4}
		              & \multirow{5}{*}{\dataset{medium}}  & \dataset{GIN}       & \meanstd{0.5845}{0.0404} \\
		              &                                    & \dataset{GT}        & \meanstd{0.5672}{0.0790} \\
		              &                                    & \dataset{GIN+}      & \meanstd{0.4830}{0.0170} \\
		              &                                    & \dataset{GCN+}      & \meanstd{0.6388}{0.0188} \\
		              &                                    & \dataset{GatedGCN+} & \meanstd{0.6459}{0.0081} \\
		\cmidrule(l){2-4}
		              & \multirow{5}{*}{\dataset{hard}}    & \dataset{GIN}       & \meanstd{0.6383}{0.0675} \\
		              &                                    & \dataset{GT}        & \meanstd{0.5212}{0.0219} \\
		              &                                    & \dataset{GIN+}      & \meanstd{0.6639}{0.0030} \\
		              &                                    & \dataset{GCN+}      & \meanstd{0.6933}{0.0046} \\
		              &                                    & \dataset{GatedGCN+} & \meanstd{0.6638}{0.0107} \\
		\bottomrule
	\end{tabular}
\end{table}

\begin{table}[htbp]
	\tabledefaults
	\centering
	\caption{Resulting F1 scores for algorithmic reasoning datasets max clique and max matching.}
	\label{table:algo_reas_results}
	\begin{tabular}{lllc}
		\toprule
		\textbf{Task} & \textbf{Difficulty}\hspace{-0.1em} & \textbf{Model}      & F1                       \\
		\midrule
		\multirow{15}{*}{Max Clique}
		              & \multirow{5}{*}{\dataset{easy}}    & \dataset{GIN}       & \meanstd{0.4584}{0.0101} \\
		              &                                    & \dataset{GT}        & \meanstd{0.5407}{0.0088} \\
		              &                                    & \dataset{GIN+}      & \meanstd{0.4635}{0.0096} \\
		              &                                    & \dataset{GCN+}      & \meanstd{0.4835}{0.0065} \\
		              &                                    & \dataset{GatedGCN+} & \meanstd{0.4687}{0.0194} \\
		\cmidrule(l){2-4}
		              & \multirow{5}{*}{\dataset{medium}}  & \dataset{GIN}       & \meanstd{0.3996}{0.0852} \\
		              &                                    & \dataset{GT}        & \meanstd{0.4859}{0.0017} \\
		              &                                    & \dataset{GIN+}      & \meanstd{0.4921}{0.0056} \\
		              &                                    & \dataset{GCN+}      & \meanstd{0.5186}{0.0439} \\
		              &                                    & \dataset{GatedGCN+} & \meanstd{0.4948}{0.0389} \\
		\cmidrule(l){2-4}
		              & \multirow{5}{*}{\dataset{hard}}    & \dataset{GIN}       & \meanstd{0.4102}{0.0820} \\
		              &                                    & \dataset{GT}        & \meanstd{0.4868}{0.0013} \\
		              &                                    & \dataset{GIN+}      & \meanstd{0.4807}{0.0027} \\
		              &                                    & \dataset{GCN+}      & \meanstd{0.5385}{0.0171} \\
		              &                                    & \dataset{GatedGCN+} & \meanstd{0.5362}{0.0245} \\
		\midrule
		\multirow{15}{*}{Max Matching}
		              & \multirow{5}{*}{\dataset{easy}}    & \dataset{GIN}       & \meanstd{0.7527}{0.0051} \\
		              &                                    & \dataset{GT}        & \meanstd{0.7402}{0.0172} \\
		              &                                    & \dataset{GIN+}      & \meanstd{0.7017}{0.0687} \\
		              &                                    & \dataset{GCN+}      & \meanstd{0.7988}{0.0047} \\
		              &                                    & \dataset{GatedGCN+} & \meanstd{0.7939}{0.0074} \\
		\cmidrule(l){2-4}
		              & \multirow{5}{*}{\dataset{medium}}  & \dataset{GIN}       & \meanstd{0.6399}{0.0231} \\
		              &                                    & \dataset{GT}        & \meanstd{0.6915}{0.009}  \\
		              &                                    & \dataset{GIN+}      & \meanstd{0.7158}{0.0188} \\
		              &                                    & \dataset{GCN+}      & \meanstd{0.6702}{0.0551} \\
		              &                                    & \dataset{GatedGCN+} & \meanstd{0.7210}{0.0224} \\
		\cmidrule(l){2-4}
		              & \multirow{5}{*}{\dataset{hard}}    & \dataset{GIN}       & \meanstd{0.6595}{0.0164} \\
		              &                                    & \dataset{GT}        & \meanstd{0.6743}{0.0038} \\
		              &                                    & \dataset{GIN+}      & \meanstd{0.6333}{0.1262} \\
		              &                                    & \dataset{GCN+}      & \meanstd{0.6854}{0.0398} \\
		              &                                    & \dataset{GatedGCN+} & \meanstd{0.7310}{0.0322} \\
		\bottomrule
	\end{tabular}
\end{table}

\begin{table}[htbp]
	\tabledefaults
	\centering
	\caption{Resulting MAE for topological sorting and flow.}
	\label{table:algo_reas_results_2}
	\begin{tabular}{lllc}
		\toprule
		\textbf{Task} & \textbf{Difficulty}\hspace{-0.1em} & \textbf{Model}      & MAE                       \\
		\midrule
		\multirow{15}{*}{Topological Sorting}
		              & \multirow{5}{*}{\dataset{easy}}    & \dataset{GIN}       & \meanstd{0.1001}{0.0089}  \\
		              &                                    & \dataset{GT}        & \meanstd{0.116}{0.0058}   \\
		              &                                    & \dataset{GIN+}      & \meanstd{0.1351}{0.0377}  \\
		              &                                    & \dataset{GCN+}      & \meanstd{0.3064}{0.0300}  \\
		              &                                    & \dataset{GatedGCN+} & \meanstd{0.1088}{0.0064}  \\
		\cmidrule(l{0.5em}){2-4}
		              & \multirow{5}{*}{\dataset{medium}}  & \dataset{GIN}       & \meanstd{0.1537}{0.0031}  \\
		              &                                    & \dataset{GT}        & \meanstd{0.1305}{0.0094}  \\
		              &                                    & \dataset{GIN+}      & \meanstd{0.1346}{0.0029}  \\
		              &                                    & \dataset{GCN+}      & \meanstd{0.3122}{0.0035}  \\
		              &                                    & \dataset{GatedGCN+} & \meanstd{0.1257}{0.0024}  \\
		\cmidrule(l{0.5em}){2-4}
		              & \multirow{5}{*}{\dataset{hard}}    & \dataset{GIN}       & \meanstd{0.1301}{0.0046}  \\
		              &                                    & \dataset{GT}        & \meanstd{0.1532}{0.0214}  \\
		              &                                    & \dataset{GIN+}      & \meanstd{0.1157}{0.0024}  \\
		              &                                    & \dataset{GCN+}      & \meanstd{0.3004}{0.0391}  \\
		              &                                    & \dataset{GatedGCN+} & \meanstd{0.1100}{0.0024}  \\
		\midrule
		\multirow{15}{*}{Flow}
		              & \multirow{5}{*}{\dataset{easy}}    & \dataset{GIN}       & \meanstd{3.4387}{0.0631}  \\
		              &                                    & \dataset{GT}        & \meanstd{4.2737}{0.0646}  \\
		              &                                    & \dataset{GIN+}      & \meanstd{5.0902}{0.2041}  \\
		              &                                    & \dataset{GCN+}      & \meanstd{4.4852}{0.0551}  \\
		              &                                    & \dataset{GatedGCN+} & \meanstd{3.0029}{0.0205}  \\
		\cmidrule(l){2-4}
		              & \multirow{5}{*}{\dataset{medium}}  & \dataset{GIN}       & \meanstd{9.5960}{0.1707}  \\
		              &                                    & \dataset{GT}        & \meanstd{6.3786}{0.4262}  \\
		              &                                    & \dataset{GIN+}      & \meanstd{9.4751}{0.4324}  \\
		              &                                    & \dataset{GCN+}      & \meanstd{8.5583}{0.9136}  \\
		              &                                    & \dataset{GatedGCN+} & \meanstd{9.8846}{0.3982}  \\
		\cmidrule(l){2-4}
		              & \multirow{5}{*}{\dataset{hard}}    & \dataset{GIN}       & \meanstd{9.5061}{0.1265}  \\
		              &                                    & \dataset{GT}        & \meanstd{6.4833}{0.0869}  \\
		              &                                    & \dataset{GIN+}      & \meanstd{10.3892}{0.3140} \\
		              &                                    & \dataset{GCN+}      & \meanstd{8.9073}{1.2009}  \\
		              &                                    & \dataset{GatedGCN+} & \meanstd{8.6309}{1.4895}  \\
		\bottomrule
	\end{tabular}
\end{table}

\begin{table}[htbp]
	\tabledefaults
	\centering
	\caption{Size generalization results on algorithmic reasoning datasets from \Cref{Section:AlgoReaso}. Each column represents an evaluation on 1000 graphs with the given number of nodes (128--512). For each experiment, the same generation parameters were used as in the results presented in \Cref{table:algo_reas_results} and \Cref{table:algo_reas_results_2}. OOT denotes the case where the underlying size generalization data was not computed within 24 hours of computation on a single cluster node.
		All generated graphs are based on the \emph{\dataset{medium}} difficulty setting. We use the same single seed for each pretrained model in each task to obtain inference results across different sizes.}
	\label{table:algo_reas_sizegenresults}
	\resizebox{0.8\textwidth}{!}{
		\begin{tabular}{llccccccc}
			\toprule
			\textbf{Dataset (Score)} & \textbf{Model} & 128    & 192    & 256    & 384    & 512    \\
			\midrule
			\multirow{2}{*}{Topological Sorting (MAE)}
			                         & GT             & 0.1346 & 0.1730 & 0.1827 & 0.2018 & 0.2071 \\
			                         & GIN            & 0.1492 & 0.181  & 0.1981 & 0.2015 & 0.2179 \\

			\midrule
			\multirow{2}{*}{MST (F1)}
			                         & GT             & 0.8720 & 0.8773 & 0.8891 & 0.8874 & 0.8817 \\
			                         & GIN            & 0.6103 & 0.8132 & 0.8605 & 0.8765 & 0.8823 \\

			\midrule
			\multirow{2}{*}{Bridges (F1)}
			                         & GT             & 0.8799 & 0.8842 & 0.8959 & 0.9049 & 0.9118 \\
			                         & GIN            & 0.9579 & 0.9213 & 0.9190 & 0.9203 & 0.9212 \\

			\midrule
			\multirow{2}{*}{Steiner Trees (F1)}
			                         & GT             & 0.5160 & 0.5322 & 0.5221 & 0.5762 & 0.5578 \\
			                         & GIN            & 0.5499 & 0.5739 & 0.6338 & 0.6651 & 0.6502 \\

			\midrule
			\multirow{2}{*}{Max Clique (F1)}
			                         & GT             & 0.4877 & 0.4849 & 0.4890 & 0.4915 & 0.4931 \\
			                         & GIN            & 0.3496 & 0.3112 & 0.3148 & 0.2926 & 0.2673 \\

			\midrule
			\multirow{2}{*}{Max Matching (F1)}
			                         & GT             & 0.7010 & 0.6552 & 0.6206 & 0.5770 & OOT    \\
			                         & GIN            & 0.6271 & 0.6326 & 0.6372 & 0.6382 & OOT    \\
			\bottomrule
		\end{tabular}}
\end{table}

\subsection{Earth systems}

\subsubsection{Weather forecasting: Medium-range atmospheric state prediction}
\label{sec:weather_forecasting_details}

\begin{figure}[t]
	\centering
	\includegraphics[width=0.3\textwidth]{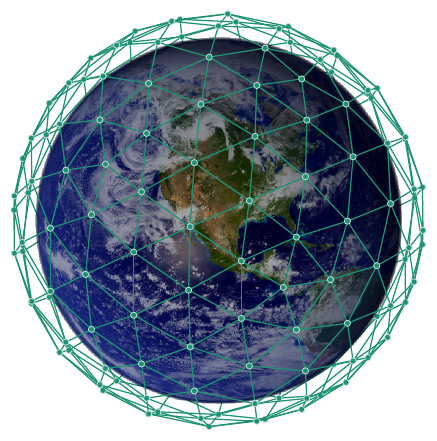}
	\caption{Down-scaled version of our node icosahedron spanning the globe. Each node contains weather variables of the corresponding location across 13 pressure levels.
		\label{fig:weather_forecasting}}
\end{figure}

\autoref{fig:weather_forecasting} provides an intuitive image of how the graph icosahedron spans the globe.

\paragraph{Related work}~
The idea of utilizing machine learning models for weather forecasting first emerged in 2020, when ~\cite{rasp2020weatherbench} and ~\cite{weyn2020improving} employed CNN architectures to predict global weather at resolutions of 5.625° and 1.9°, respectively. A key breakthrough occurred when ~\cite{keisler2022forecasting} first applied MPNNs to more naturally represent the globe, enabling 3D state predictions 6 hours ahead and facilitating multi-day forecasts. This approach has already achieved a skill comparable to that of 1° global NWP (GFS/ECMWF) on specific metrics, outperforming prior data-driven models. Building on this idea, \cite{lam2023learning} introduced \new{GraphCast}, a multi-scale MPNN-based system that delivers today's forecasts at 0.25° resolution. GraphCast achieved unprecedented accuracy, outperforming even ECMWF’s \new{high-resolution deterministic model} (HRES) out to ten days. The success of these deterministic forecasts raised the question of probabilistic prediction. ~\cite{price2023gencast} introduced \new{GenCast}, a generative weather model that produces an ensemble of forecasts rather than a single deterministic run. For this, they used a diffusion-based approach built on an MPNN backbone, outperforming ECMWF’s \new{ensemble system} (ENS), the world’s top operational probabilistic forecast on over 97\% of evaluated metrics. Several other advances can be noted. \cite{oskarsson2024probabilistic} have proposed Graph-EFM, a model featuring a hierarchical GNN that seamlessly handles global and regional forecasts. With \new{OneForecast},~\cite{gao2025oneforecast} introduced an MPNN framework that uses nested multi-scale graphs and adaptive message passing to improve local extreme-event predictions within a global context. These works indicate a trend toward hybrid solutions that densify graphs in target regions for high-resolution detail while maintaining global consistency. Current state-of-the-art models, such as Pangu-Weather~\Citep{bi2022pangu}, Aurora~\Citep{bodnar2024aurora}, and FGN~\Citep{alet2025skillful}, all present excellent results that, in many areas, far exceed those of conventional NWP methods. It is important to note, however, that while machine learning-based approaches exceed NWP skill on tested time scales, they do not explicitly enforce conservation laws. Over long integration horizons, this can lead to subtle physical drifts that may limit specific climate-related uses. This becomes particularly relevant under regime shifts, for example, caused by climate change, that lie outside the distributions the models were trained on. Furthermore, compared with NWP models, ML-based approaches offer very limited interpretability and function mainly as black boxes.

\subsection{Limitations}
In the following, we discuss the limitations of each dataset and the proposed baseline methods. We highlight additional dataset-specific limitations and results in the respective sections above. For Bluesky, the biggest limitation is the graph sizes across tasks in the dataset, as they prevent running graph transformer baselines and larger GNNs on them.

CO and algorithmic reasoning datasets are purely synthetic, as collecting real-world instances for each task is a major challenge. However, both the datasets and their evaluation framework can be expanded to include real-world data for training or testing as well.

For weather forecasting and SAT solving, graph sizes prevent many computationally intensive methods from being run on them. In addition, weather forecasting data is known to require substantial preprocessing as outlined by \citet{lam2023learning}. Aside from small instances, SAT instances are currently too large to run available solvers on them to generate node features and targets.

Lastly, the chip design task is currently very challenging for learning-based models, with no baselines available. However, we include non-learning baselines to showcase the potential performance a learning-based method should achieve.

Given the number of tasks evaluated in the benchmark, we restricted ourselves to common GNN baselines from \citet{luo2025GNNPlus}, as they achieve strong performance across currently available benchmarks. This also limits the application of HPO (see \autoref{sec:hpo}) and the size of hyperparameter tuning budgets. Furthermore, we provide a simple graph transformer baseline. Since there are, to our knowledge, no widely available graph foundation models that work across tasks and evaluation targets, including node-, edge-, and graph-level tasks, we have decided not to include them so far.
\section{Hardware}

\subsection{SAT solving}
We ran the SAT solvers on a cluster equipped with two AMD EPYC 7\,543 32-core processors, each with 512MB of L3 cache and 1 TB of RAM, running Rocky Linux 9. We allowed each solver to use up to 64GB of memory, so up to 8 solvers could run on the same node in parallel. Before conducting the experiments, we performed a preliminary experiment, running Kissat on 100 random SAT instances 10 times, ensuring consistent results (in terms of computation time).

For training the GNNs, we used a cluster with 4 GPU nodes, each containing 4 H100 NVL GPUs with 94GB of vRAM. Each task for GIN and GRASS ran for two to three hours, and ten hours for the GNNPlus models. Furthermore, the HPO sweep outlined in \autoref{sec:hpo} was conducted over 60 configurations.

\subsection{Algorithmic reasoning}
Each algorithmic reasoning task was evaluated on a cluster node equipped with a single NVIDIA L40 GPU, with 48GB of vRAM and 512GB of RAM. Throughout all experiments, the same cluster configuration was used to ensure comparable wall-clock times. For each single task a single  NVIDIA L40 GPU with 120GB of RAM was used. Due to computational time constraints, positional and structural encodings (such as RWSE and LPE) and edge transformations for the graph transformer baseline were provided beforehand.
Runtimes across tasks were very similar, with GNNPlus baselines running for 20 minutes each after dataset preprocessing. The GIN baseline ran for 29 minutes on each task.

The edge level task preprocessing as outlined in \autoref{app:data_preproc} took between 1 hour and 40 minutes for the maximum matching task and 4 hours for the minimum spanning tree task. For preprocessing, the complete cluster node was used.
\subsection{Weather forecasting}
For weather forecasting, a cluster node with four NVIDIA L40 GPUs was used during evaluation. Additionally, a cluster node with four NVIDIA A100 GPUs was utilized during training to accommodate larger batch sizes. Both cluster nodes use 512GB of RAM. Training took 1 day, 18 hours, and 35 minutes for a single seed.
\subsection{Combinatorial Optimization}
Experiments for unsupervised CO were performed on an NVIDIA H100 GPU with 80GB of vRAM, and for supervised learning were on an NVIDIA L40S GPU with 48 GB vRAM. Each run on the GNNPlus models took an average of 14 hours for the small datasets and 55 hours for the large datasets.
\subsection{Electronic Circuits}
We conducted all experiments on a single NVIDIA L40 GPU with 120GB of RAM. Furthermore, RWSE was computed during data loading. The GT and GIN baseline evaluation took 15 minutes for 5 components, 13 minutes for 7 components, and 12 minutes for 10 components, respectively. GNNPlus baselines took 13 minutes per run for 5 and 7 components, and 11 minutes for 10 components.
\subsection{Social networks}
We conducted all experiments on a single  NVIDIA A100 GPU with 120GB of RAM on the same cluster node. Training and testing of GNNPlus methods took on average 1 hour, with models on the Bluesky quotes task running longer at 2 hours.

\section{Experimental setup}\label{Section:ExpSetup}
Similar to~\cite{Spe+2025, stoll2025generalizable}, we provide an encoder-processor-decoder architecture for all tasks, which we detail in the following. While the task-specific components of our architecture vary, we provide a general architecture to measure the performance of selected baselines on our tasks. For the chip design and weather forecast dataset, we adapt this architecture to provide the necessary task-dependent encoder and decoder functionality.

\paragraph{Encoder}~
For each task, we provide a task-specific encoder that derives encodings of node- and edge-level features with a common dimension $d \in \mathbb{R}^+$. These features are then passed to the processor as node and/or edge features. For models that require tokenized input, we provide node-level tokens. For missing node or edge features, a learnable vector of dimension $d$ is used in their place. In case of the graph transformer baseline, we add a \texttt{[cls]} token for graph-level representations, following ~\cite{Spe+2025, stoll2025generalizable}.

\paragraph{Processor}~
As our processor, we select one of the baseline models we evaluate. These consist of two dummy baselines: an MLP and a DeepSet architecture; a GNN baseline using a GIN architecture; and a Graph Transformer baseline. Independent of the selected baseline, we compute updated node and graph representations at each layer as follows, given a graph $G$ with node representations $\vec{X}$,
\begin{equation}\label{eq:processor_eq}
	\vec{X} = \phi(\textsf{LayerNorm}(\vec{X}), G)
\end{equation}
\begin{equation*}
	\vec{X} = \vec{X} + \textsf{MLP}(\textsf{LayerNorm}(\vec{X}))
\end{equation*}
with the following MLP layer:
\begin{equation*}
	\textsf{MLP}(\vec{x}) \coloneqq \sigma(\textsf{LayerNorm}(\vec{x})\vec{W}_1)\vec{W}_2,
\end{equation*}
where $\sigma$ denotes the GeLU nonlinearity and $\vec{W}_1, \vec{W}_2$ are weight matrices.
Further, $\phi$ is the selected baseline, and \textsf{MLP} denotes a two-layer multilayer perceptron with GELU non-linearity~\citep{Hendrycks2016Gelu}. For the graph transformer baseline, we employed biased attention as described by~\cite{Spe+2025, stoll2025generalizable}. Furthermore, we incorporate RWSE~\citep{Dwivedi2022RWSE} and LPE~\citep{Mueller2024Aligning} as node positional encodings for our GNN and graph transformer baselines.

\paragraph{Decoder}~
Since each task requires task-specific predictions, we provide a decoder for each task. Across all functions, we use the same decoder layout, differing only in the predictions they make. We propose a decoder two-layer multilayer perceptron for our baseline models consisting of two linear layers $\vec{W}_1 \in \mathbb{R}^{d \times d}, \vec{W}_2 \in \mathbb{R}^{d \times o}$ as well as GELU non-linearity, with $o$ denoting the task-specific output dimension, i.e.,
\begin{equation*}
	\vec{W}_2(\textsf{LayerNorm}(\textsf{GELU}(\vec{W}_1 x)).
\end{equation*}
Optionally, a bias term can be added to the decoder.

\paragraph{Pretraining}~
In addition to training models using our encoder-processor-decoder architecture, we provide support for evaluating pretrained models on specific tasks. These are further specified in each subsection concerning additional evaluation benchmarks.

\paragraph{Evaluation protocol}~
Using the aforementioned hyperparameter tuning process, we evaluate all baseline models across three random seeds for each task and report the mean performance and standard deviation. For out-of-distribution generalization tasks, we utilize pretrained models from these baseline evaluations.

\subsection{Hyperparameter selection}
Here, we provide the hyperparameters used in training our baseline models for each task. We note that in most cases, no extensive hyperparameter search was conducted for the baseline models.
We highlight the hyperparameters included in the sweep for each dataset below:

For blueksy datasets we sweeped over learning rates $\{1e^{-3},2e^{-3},5e^{-4}\}$ for all models and for the number of layers $l \in \{2,3,4,6,8\}$ and hidden dimension $\{128,192,384\}$. We further included residual connections and feed-forward MLP layers in all GNNPlus architectures.

Regarding electronic circuits, we provide results obtained from a hyperparameter sweep over learning rates $\{1e^{-3},2e^{-3},5e^{-4}\}$, RWSE encoding dimension $\{4,8,16\}$, and number of layers $\{3,6\}$.

For CO datasets we sweep over the following learning rates $\{1e^{-3},2e^{-3},5e^{-4}\}$ and use the six layers across all models.

Across all algorithmic reasoning tasks, we conducted hyperparameter tuning with the following selection of hyperparameters. For GIN and GT over learning rates $\{1e^{-4},2e^{-4},3e^{-4}\}$ with the exception of steiner tree tasks, where we also considered $\{1e^{-3},2e^{-3}\}$. Furthermore, layer choices are given by the following $\{6,16\}$ and PE encoding dimension and type: $\{4,8,16,32\}$ and $\{RWSE, LPE\}$.
In contrast to this, the GNNPlus methods are tuned with the following hyperparameter budget:
learning rates $\{1e^{-3},2e^{-3},5e^{-4}\}$, PE encoding dimension $\{4,8,16\}$ and layer choices $\{3,4\}$

For SAT solving, we used automatic hyperparameter selection via SMAC3. The process is outlined in \autoref{sec:hpo}.

For weather forecasting, we only swept over the following learning rates: $\{1e-4,2e-4,5e-5\}$.

\begin{table}[htbp]
	\centering
	\caption{List of hyperparameters used for baseline models in BlueSky datasets.}

	\resizebox{1.0\columnwidth}{!}{%
		\label{table:blue_hyperparms}
\begin{tabular}{llccccccccc}
			\toprule
			\textbf{Hyperparam.} & \textbf{Variable} & $\text{GNN}_{\text{quotes}}$ & $\text{MLP}_{\text{quotes}}$ & $\text{Deep Sets}_{\text{quotes}}$ & $\text{GNN}_{\text{replies}}$ & $\text{MLP}_{\text{replies}}$ & $\text{Deep Sets}_{\text{replies}}$ & $\text{GNN}_{\text{reposts}}$ & $\text{MLP}_{\text{reposts}}$ & $\text{Deep Sets}_{\text{reposts}}$ \\

			\midrule
			\multirow{2}{*}{General}
			                     & Learning rate     & 1e-3                         & 1e-3                         & 1e-3                               & 1e-3                          & 1e-3                          & 1e-3                                & 1e-3                          & 1e-3                          & 1e-3                                \\
			                     & Weight decay      & 0.                           & 0.                           & 0.                                 & 0.                            & 0.                            & 0.                                  & 0.                            & 0.                            & 0.                                  \\
			                     & Dropout           & 0.                           & 0.2                          & 0.                                 & 0.                            & 0.                            & 0.                                  & 0.                            & 0.2                           & 0.2                                 \\
			                     & Epochs            & 1\,000                       & 1\,000                       & 1\,000                             & 1\,000                        & 1\,000                        & 1\,000                              & 1\,000                        & 1\,000                        & 1\,000                              \\
\bottomrule
		\end{tabular}}
\end{table}
\begin{table}[htbp]
	\centering
	\caption{Incumbent hyperparameters (AS:LCG). List of hyperparameters used for baseline models in SAT solving tasks. These include parameters for positional encoding (PE), the architecture (GIN), and general learning parameters.}
	\resizebox{\columnwidth}{!}{%
		\label{table:incumbents:as_lcg}
		\begin{tabular}{llcccccc}
			\toprule
			Hyperparam. & Variable      & Gated GCN+ & GCN        & GCN+     & GIN       & GIN+     & GRASS    \\
			\midrule
			            & Learning rate & 1.17e-05   & 0.00183706 & 4.59e-04 & 4.76e-04  & 2.23e-05 & 2.29e-05 \\
			            & Batch size    & 16         & 16         & 16       & 16        & 16       & 16       \\
			            & Weight decay  & 1.86e-08   & 0.0515994  & 3.23e-05 & 1.35e-08  & 1.43e-07 & 3.28e-06 \\
			            & Dropout       & 0.0591372  & 0.0716766  & 0.144238 & 0.0812465 & 0.240447 & 0.237349 \\
			            & Steps         & 100000     & 100000     & 100000   & 100000    & 100000   & 100000   \\
			            & Warmup steps  & 13134      & 9622       & 2134     & 11146     & 9467     & 11215    \\
			            & Scheduler     & cosine     & cosine     & cosine   & cosine    & cosine   & cosine   \\
			            & Optimizer     & AdamW      & AdamW      & AdamW    & AdamW     & AdamW    & AdamW    \\
			            & PE Type       & -          & -          & -        & -         & -        & -        \\
			            & Encoding dim. & -          & -          & -        & -         & -        & -        \\
			            & Hidden dim.   & 256        & 512        & 256      & 384       & 512      & 256      \\
			            & Layers        & 5          & 6          & 7        & 7         & 7        & 9        \\
			            & Activation    & relu       & relu       & relu     & relu      & relu     & relu     \\
			\bottomrule
		\end{tabular}%
	}
\end{table}

\begin{table}[htbp]
	\centering
	\caption{Hyperparameters for baseline models on Bluesky social graph tasks: Quotes (Q), Replies (Re), and Reposts (Rp). Configurations are derived from GNN+ architectural settings for GCN, GIN, and GatedGCN.}
	\resizebox{\columnwidth}{!}{%
		\label{table:BlueskyHyperparametersFinal}
		\begin{tabular}{llccccccc}
			\toprule
			\textbf{Hyperparam.} & \textbf{Variable} & $\text{GCN}_{Q}$                      & $\text{GIN}_{Q}$ & $\text{GatedGCN}_{Q}$ & $\text{GCN}_{Re}$ & $\text{GIN}_{Re}$ & $\text{GCN}_{Rp}$ & $\text{GIN}_{Rp}$ \\
			\midrule
			\multirow{5}{*}{General}
			                     & Learning rate     & 5e-4                                  & 2e-3             & 5e-4                  & 2e-3              & 2e-3              & 2e-3              & 5e-4              \\
			                     & Batch size        & 1                                     & 1                & 1                     & 1                 & 1                 & 1                 & 1                 \\
			                     & Weight decay      & 0.1                                   & 0.1              & 0.1                   & 0.1               & 0.1               & 0.1               & 0.1               \\
			                     & Dropout           & 0.1                                   & 0.1              & 0.1                   & 0.1               & 0.1               & 0.1               & 0.1               \\
			                     & Steps             & 1,000                                 & 1,000            & 1,000                 & 1,000             & 1,000             & 1,000             & 1,000             \\
			\midrule
			\multirow{4}{*}{Optimizing}
			                     & Warmup steps      & 50                                    & 50               & 50                    & 50                & 50                & 50                & 50                \\
			                     & Scheduler         & \multicolumn{7}{c}{CosineAnnealingLR}                                                                                                                            \\
\bottomrule
		\end{tabular}}
\end{table}

\begin{table}[htbp]
	\centering
	\caption{Incumbent hyperparameters (AS:VCG). List of hyperparameters used for baseline models in SAT solving tasks. These include parameters for positional encoding (PE), the architecture (GIN), and general learning parameters.}
	\resizebox{\columnwidth}{!}{%
		\label{table:incumbents:as_vcg}
		\begin{tabular}{llcccccc}
			\toprule
			Hyperparam. & Variable      & Gated GCN+ & GCN        & GCN+       & GIN       & GIN+       & GRASS     \\
			\midrule
			            & Learning rate & 4.59e-04   & 0.00183706 & 0.00325039 & 4.76e-04  & 0.00325039 & 6.34e-05  \\
			            & Batch size    & 16         & 16         & 16         & 16        & 16         & 16        \\
			            & Weight decay  & 3.23e-05   & 0.0515994  & 2.48e-08   & 1.35e-08  & 2.48e-08   & 0.0467242 \\
			            & Dropout       & 0.144238   & 0.0716766  & 0.352207   & 0.0812465 & 0.352207   & 0.308829  \\
			            & Steps         & 100000     & 100000     & 100000     & 100000    & 100000     & 100000    \\
			            & Warmup steps  & 2134       & 9622       & 14173      & 11146     & 14173      & 7011      \\
			            & Scheduler     & cosine     & cosine     & cosine     & cosine    & cosine     & cosine    \\
			            & Optimizer     & AdamW      & AdamW      & AdamW      & AdamW     & AdamW      & AdamW     \\
			            & PE Type       & -          & -          & -          & -         & -          & -         \\
			            & Encoding dim. & -          & -          & -          & -         & -          & -         \\
			            & Hidden dim.   & 256        & 512        & 256        & 384       & 256        & 384       \\
			            & Layers        & 7          & 6          & 9          & 7         & 9          & 7         \\
			            & Activation    & relu       & relu       & relu       & relu      & relu       & relu      \\
			\bottomrule
		\end{tabular}%
	}
\end{table}

\begin{table}[htbp]
	\centering
	\caption{Incumbent hyperparameters (AS:VG). List of hyperparameters used for baseline models in SAT solving tasks. These include parameters for positional encoding (PE), the architecture (GIN), and general learning parameters.}
	\resizebox{\columnwidth}{!}{%
		\label{table:incumbents:as_vg}
		\begin{tabular}{llcccccc}
			\toprule
			Hyperparam. & Variable      & Gated GCN+ & GCN       & GCN+       & GIN        & GIN+     & GRASS      \\
			\midrule
			            & Learning rate & 0.036691   & 2.62e-04  & 0.00126235 & 0.00133172 & 9.74e-05 & 9.04e-04   \\
			            & Batch size    & 16         & 16        & 16         & 16         & 16       & 16         \\
			            & Weight decay  & 0.00279961 & 0.0123051 & 1.70e-05   & 0.0273136  & 3.95e-08 & 0.00392744 \\
			            & Dropout       & 0.274407   & 0.357595  & 0.211827   & 0.150787   & 0.464041 & 0.272442   \\
			            & Steps         & 100000     & 100000    & 100000     & 100000     & 100000   & 100000     \\
			            & Warmup steps  & 7036       & 3163      & 8877       & 14310      & 5652     & 5851       \\
			            & Scheduler     & cosine     & cosine    & cosine     & cosine     & cosine   & cosine     \\
			            & Optimizer     & AdamW      & AdamW     & AdamW      & AdamW      & AdamW    & AdamW      \\
			            & PE Type       & -          & -         & -          & -          & -        & -          \\
			            & Encoding dim. & -          & -         & -          & -          & -        & -          \\
			            & Hidden dim.   & 384        & 512       & 256        & 256        & 256      & 512        \\
			            & Layers        & 9          & 5         & 5          & 6          & 5        & 4          \\
			            & Activation    & relu       & relu      & relu       & relu       & relu     & relu       \\
			\bottomrule
		\end{tabular}%
	}
\end{table}

\begin{table}[htbp]
	\centering
	\caption{Incumbent hyperparameters (EPM:LCG \solver{AMSAT}). List of hyperparameters used for baseline models in SAT solving tasks. These include parameters for positional encoding (PE), the architecture (GIN), and general learning parameters.}
	\resizebox{\columnwidth}{!}{%
		\label{table:incumbents:epm_lcg_solver_solver_amsat}
		\begin{tabular}{llcccccc}
			\toprule
			Hyperparam. & Variable      & Gated GCN+ & GCN        & GCN+     & GIN       & GIN+      & GRASS    \\
			\midrule
			            & Learning rate & 0.00379986 & 0.00153328 & 0.002836 & 4.76e-04  & 0.0157381 & 3.34e-04 \\
			            & Batch size    & 16         & 16         & 16       & 16        & 16        & 16       \\
			            & Weight decay  & 2.51e-07   & 5.03e-08   & 6.49e-04 & 1.35e-08  & 2.60e-05  & 1.76e-04 \\
			            & Dropout       & 0.434063   & 0.0095966  & 0.271249 & 0.0812465 & 0.0159195 & 0.448019 \\
			            & Steps         & 100000     & 100000     & 100000   & 100000    & 100000    & 100000   \\
			            & Warmup steps  & 12042      & 6846       & 9352     & 11146     & 14634     & 2913     \\
			            & Scheduler     & cosine     & cosine     & cosine   & cosine    & cosine    & cosine   \\
			            & Optimizer     & AdamW      & AdamW      & AdamW    & AdamW     & AdamW     & AdamW    \\
			            & PE Type       & -          & -          & -        & -         & -         & -        \\
			            & Encoding dim. & -          & -          & -        & -         & -         & -        \\
			            & Hidden dim.   & 256        & 512        & 256      & 384       & 256       & 512      \\
			            & Layers        & 6          & 7          & 7        & 7         & 8         & 9        \\
			            & Activation    & relu       & relu       & relu     & relu      & relu      & relu     \\
			\bottomrule
		\end{tabular}%
	}
\end{table}

\begin{table}[htbp]
	\centering
	\caption{Incumbent hyperparameters (EPM:LCG \solver{BreakIDKissat}). List of hyperparameters used for baseline models in SAT solving tasks. These include parameters for positional encoding (PE), the architecture (GIN), and general learning parameters.}
	\resizebox{\columnwidth}{!}{%
		\label{table:incumbents:epm_lcg_solver_solver_breakidkissat}
		\begin{tabular}{llcccccc}
			\toprule
			Hyperparam. & Variable      & Gated GCN+ & GCN        & GCN+       & GIN       & GIN+     & GRASS    \\
			\midrule
			            & Learning rate & 0.010885   & 0.00183706 & 0.00379986 & 4.76e-04  & 3.24e-05 & 2.29e-05 \\
			            & Batch size    & 16         & 16         & 16         & 16        & 16       & 16       \\
			            & Weight decay  & 0.0719626  & 0.0515994  & 2.51e-07   & 1.35e-08  & 1.44e-05 & 3.28e-06 \\
			            & Dropout       & 0.207184   & 0.0716766  & 0.434063   & 0.0812465 & 0.317529 & 0.237349 \\
			            & Steps         & 100000     & 100000     & 100000     & 100000    & 100000   & 100000   \\
			            & Warmup steps  & 1973       & 9622       & 12042      & 11146     & 19819    & 11215    \\
			            & Scheduler     & cosine     & cosine     & cosine     & cosine    & cosine   & cosine   \\
			            & Optimizer     & AdamW      & AdamW      & AdamW      & AdamW     & AdamW    & AdamW    \\
			            & PE Type       & -          & -          & -          & -         & -        & -        \\
			            & Encoding dim. & -          & -          & -          & -         & -        & -        \\
			            & Hidden dim.   & 384        & 512        & 256        & 384       & 512      & 256      \\
			            & Layers        & 4          & 6          & 6          & 7         & 6        & 9        \\
			            & Activation    & relu       & relu       & relu       & relu      & relu     & relu     \\
			\bottomrule
		\end{tabular}%
	}
\end{table}

\begin{table}[htbp]
	\centering
	\caption{Incumbent hyperparameters (EPM:LCG \solver{Kissat}). List of hyperparameters used for baseline models in SAT solving tasks. These include parameters for positional encoding (PE), the architecture (GIN), and general learning parameters.}
	\resizebox{\columnwidth}{!}{%
		\label{table:incumbents:epm_lcg_solver_solver_kissat}
		\begin{tabular}{llcccccc}
			\toprule
			Hyperparam. & Variable      & Gated GCN+ & GCN        & GCN+      & GIN       & GIN+       & GRASS      \\
			\midrule
			            & Learning rate & 0.002836   & 0.00153328 & 0.0160222 & 4.76e-04  & 0.00126654 & 0.00598895 \\
			            & Batch size    & 16         & 16         & 16        & 16        & 16         & 16         \\
			            & Weight decay  & 6.49e-04   & 5.03e-08   & 3.69e-07  & 1.35e-08  & 1.89e-06   & 0.0146903  \\
			            & Dropout       & 0.271249   & 0.0095966  & 0.06191   & 0.0812465 & 0.390265   & 0.0823471  \\
			            & Steps         & 100000     & 100000     & 100000    & 100000    & 100000     & 100000     \\
			            & Warmup steps  & 9352       & 6846       & 7177      & 11146     & 15627      & 12605      \\
			            & Scheduler     & cosine     & cosine     & cosine    & cosine    & cosine     & cosine     \\
			            & Optimizer     & AdamW      & AdamW      & AdamW     & AdamW     & AdamW      & AdamW      \\
			            & PE Type       & -          & -          & -         & -         & -          & -          \\
			            & Encoding dim. & -          & -          & -         & -         & -          & -          \\
			            & Hidden dim.   & 256        & 512        & 384       & 384       & 512        & 256        \\
			            & Layers        & 7          & 7          & 9         & 7         & 8          & 9          \\
			            & Activation    & relu       & relu       & relu      & relu      & relu       & relu       \\
			\bottomrule
		\end{tabular}%
	}
\end{table}

\begin{table}[htbp]
	\centering
	\caption{Incumbent hyperparameters (EPM:LCG \solver{KissatMABDC}). List of hyperparameters used for baseline models in SAT solving tasks. These include parameters for positional encoding (PE), the architecture (GIN), and general learning parameters.}
	\resizebox{\columnwidth}{!}{%
		\label{table:incumbents:epm_lcg_solver_solver_kissatmabdc}
		\begin{tabular}{llcccccc}
			\toprule
			Hyperparam. & Variable      & Gated GCN+ & GCN        & GCN+       & GIN       & GIN+       & GRASS     \\
			\midrule
			            & Learning rate & 2.69e-05   & 0.00153328 & 0.00258292 & 4.76e-04  & 0.00379986 & 0.0157381 \\
			            & Batch size    & 16         & 16         & 16         & 16        & 16         & 16        \\
			            & Weight decay  & 3.22e-04   & 5.03e-08   & 0.0981588  & 1.35e-08  & 2.51e-07   & 2.60e-05  \\
			            & Dropout       & 0.378053   & 0.0095966  & 0.145039   & 0.0812465 & 0.434063   & 0.0159195 \\
			            & Steps         & 100000     & 100000     & 100000     & 100000    & 100000     & 100000    \\
			            & Warmup steps  & 10680      & 6846       & 3775       & 11146     & 12042      & 14634     \\
			            & Scheduler     & cosine     & cosine     & cosine     & cosine    & cosine     & cosine    \\
			            & Optimizer     & AdamW      & AdamW      & AdamW      & AdamW     & AdamW      & AdamW     \\
			            & PE Type       & -          & -          & -          & -         & -          & -         \\
			            & Encoding dim. & -          & -          & -          & -         & -          & -         \\
			            & Hidden dim.   & 256        & 512        & 512        & 384       & 256        & 256       \\
			            & Layers        & 4          & 7          & 10         & 7         & 6          & 8         \\
			            & Activation    & relu       & relu       & relu       & relu      & relu       & relu      \\
			\bottomrule
		\end{tabular}%
	}
\end{table}

\begin{table}[htbp]
	\centering
	\caption{Incumbent hyperparameters (EPM:VCG \solver{AMSAT}). List of hyperparameters used for baseline models in SAT solving tasks. These include parameters for positional encoding (PE), the architecture (GIN), and general learning parameters.}
	\resizebox{\columnwidth}{!}{%
		\label{table:incumbents:epm_vcg_solver_solver_amsat}
		\begin{tabular}{llcccccc}
			\toprule
			Hyperparam. & Variable      & Gated GCN+ & GCN        & GCN+       & GIN       & GIN+     & GRASS     \\
			\midrule
			            & Learning rate & 0.010885   & 0.00153328 & 0.00153328 & 4.76e-04  & 1.51e-04 & 2.62e-04  \\
			            & Batch size    & 16         & 16         & 16         & 16        & 16       & 16        \\
			            & Weight decay  & 0.0719626  & 5.03e-08   & 5.03e-08   & 1.35e-08  & 9.71e-05 & 0.0123051 \\
			            & Dropout       & 0.207184   & 0.0095966  & 0.0095966  & 0.0812465 & 0.459118 & 0.357595  \\
			            & Steps         & 100000     & 100000     & 100000     & 100000    & 100000   & 100000    \\
			            & Warmup steps  & 1973       & 6846       & 6846       & 11146     & 14683    & 3163      \\
			            & Scheduler     & cosine     & cosine     & cosine     & cosine    & cosine   & cosine    \\
			            & Optimizer     & AdamW      & AdamW      & AdamW      & AdamW     & AdamW    & AdamW     \\
			            & PE Type       & -          & -          & -          & -         & -        & -         \\
			            & Encoding dim. & -          & -          & -          & -         & -        & -         \\
			            & Hidden dim.   & 384        & 512        & 512        & 384       & 512      & 512       \\
			            & Layers        & 4          & 7          & 7          & 7         & 10       & 5         \\
			            & Activation    & relu       & relu       & relu       & relu      & relu     & relu      \\
			\bottomrule
		\end{tabular}%
	}
\end{table}

\begin{table}[htbp]
	\centering
	\caption{Incumbent hyperparameters (EPM:VCG \solver{BreakIDKissat}). List of hyperparameters used for baseline models in SAT solving tasks. These include parameters for positional encoding (PE), the architecture (GIN), and general learning parameters.}
	\resizebox{\columnwidth}{!}{%
		\label{table:incumbents:epm_vcg_solver_solver_breakidkissat}
		\begin{tabular}{llcccccc}
			\toprule
			Hyperparam. & Variable      & Gated GCN+ & GCN        & GCN+       & GIN      & GIN+       & GRASS    \\
			\midrule
			            & Learning rate & 0.0416586  & 0.00153328 & 0.00183706 & 4.59e-04 & 0.00258292 & 2.29e-05 \\
			            & Batch size    & 16         & 16         & 16         & 16       & 16         & 16       \\
			            & Weight decay  & 0.0492703  & 5.03e-08   & 0.0515994  & 3.23e-05 & 0.0981588  & 3.28e-06 \\
			            & Dropout       & 0.216644   & 0.0095966  & 0.0716766  & 0.144238 & 0.145039   & 0.237349 \\
			            & Steps         & 100000     & 100000     & 100000     & 100000   & 100000     & 100000   \\
			            & Warmup steps  & 17409      & 6846       & 9622       & 2134     & 3775       & 11215    \\
			            & Scheduler     & cosine     & cosine     & cosine     & cosine   & cosine     & cosine   \\
			            & Optimizer     & AdamW      & AdamW      & AdamW      & AdamW    & AdamW      & AdamW    \\
			            & PE Type       & -          & -          & -          & -        & -          & -        \\
			            & Encoding dim. & -          & -          & -          & -        & -          & -        \\
			            & Hidden dim.   & 256        & 512        & 512        & 256      & 512        & 256      \\
			            & Layers        & 4          & 7          & 6          & 7        & 10         & 9        \\
			            & Activation    & relu       & relu       & relu       & relu     & relu       & relu     \\
			\bottomrule
		\end{tabular}%
	}
\end{table}

\begin{table}[htbp]
	\centering
	\caption{Incumbent hyperparameters (EPM:VCG \solver{Kissat}). List of hyperparameters used for baseline models in SAT solving tasks. These include parameters for positional encoding (PE), the architecture (GIN), and general learning parameters.}
	\resizebox{\columnwidth}{!}{%
		\label{table:incumbents:epm_vcg_solver_solver_kissat}
		\begin{tabular}{llcccccc}
			\toprule
			Hyperparam. & Variable      & Gated GCN+ & GCN        & GCN+       & GIN       & GIN+       & GRASS      \\
			\midrule
			            & Learning rate & 0.002836   & 0.00153328 & 0.00153328 & 4.76e-04  & 0.00258292 & 0.00598895 \\
			            & Batch size    & 16         & 16         & 16         & 16        & 16         & 16         \\
			            & Weight decay  & 6.49e-04   & 5.03e-08   & 5.03e-08   & 1.35e-08  & 0.0981588  & 0.0146903  \\
			            & Dropout       & 0.271249   & 0.0095966  & 0.0095966  & 0.0812465 & 0.145039   & 0.0823471  \\
			            & Steps         & 100000     & 100000     & 100000     & 100000    & 100000     & 100000     \\
			            & Warmup steps  & 9352       & 6846       & 6846       & 11146     & 3775       & 12605      \\
			            & Scheduler     & cosine     & cosine     & cosine     & cosine    & cosine     & cosine     \\
			            & Optimizer     & AdamW      & AdamW      & AdamW      & AdamW     & AdamW      & AdamW      \\
			            & PE Type       & -          & -          & -          & -         & -          & -          \\
			            & Encoding dim. & -          & -          & -          & -         & -          & -          \\
			            & Hidden dim.   & 256        & 512        & 512        & 384       & 512        & 256        \\
			            & Layers        & 7          & 7          & 7          & 7         & 10         & 9          \\
			            & Activation    & relu       & relu       & relu       & relu      & relu       & relu       \\
			\bottomrule
		\end{tabular}%
	}
\end{table}

\begin{table}[htbp]
	\centering
	\caption{Incumbent hyperparameters (EPM:VCG \solver{KissatMABDC}). List of hyperparameters used for baseline models in SAT solving tasks. These include parameters for positional encoding (PE), the architecture (GIN), and general learning parameters.}
	\resizebox{\columnwidth}{!}{%
		\label{table:incumbents:epm_vcg_solver_solver_kissatmabdc}
		\begin{tabular}{llcccccc}
			\toprule
			Hyperparam. & Variable      & Gated GCN+ & GCN       & GCN+       & GIN        & GIN+     & GRASS     \\
			\midrule
			            & Learning rate & 4.59e-04   & 4.76e-04  & 0.00258292 & 0.00126235 & 1.51e-04 & 0.0369504 \\
			            & Batch size    & 16         & 16        & 16         & 16         & 16       & 16        \\
			            & Weight decay  & 3.23e-05   & 1.35e-08  & 0.0981588  & 1.70e-05   & 9.71e-05 & 0.0683645 \\
			            & Dropout       & 0.144238   & 0.0812465 & 0.145039   & 0.211827   & 0.459118 & 0.310739  \\
			            & Steps         & 100000     & 100000    & 100000     & 100000     & 100000   & 100000    \\
			            & Warmup steps  & 2134       & 11146     & 3775       & 8877       & 14683    & 1423      \\
			            & Scheduler     & cosine     & cosine    & cosine     & cosine     & cosine   & cosine    \\
			            & Optimizer     & AdamW      & AdamW     & AdamW      & AdamW      & AdamW    & AdamW     \\
			            & PE Type       & -          & -         & -          & -          & -        & -         \\
			            & Encoding dim. & -          & -         & -          & -          & -        & -         \\
			            & Hidden dim.   & 256        & 384       & 512        & 256        & 512      & 384       \\
			            & Layers        & 7          & 7         & 10         & 5          & 10       & 6         \\
			            & Activation    & relu       & relu      & relu       & relu       & relu     & relu      \\
			\bottomrule
		\end{tabular}%
	}
\end{table}

\begin{table}[htbp]
	\centering
	\caption{Incumbent hyperparameters (EPM:VG \solver{AMSAT}). List of hyperparameters used for baseline models in SAT solving tasks. These include parameters for positional encoding (PE), the architecture (GIN), and general learning parameters.}
	\resizebox{\columnwidth}{!}{%
		\label{table:incumbents:epm_vg_solver_solver_amsat}
		\begin{tabular}{llcccccc}
			\toprule
			Hyperparam. & Variable      & Gated GCN+ & GCN        & GCN+      & GIN       & GIN+      & GRASS     \\
			\midrule
			            & Learning rate & 0.036691   & 0.00133172 & 0.0029426 & 0.0029426 & 0.0369504 & 0.010885  \\
			            & Batch size    & 16         & 16         & 16        & 16        & 16        & 16        \\
			            & Weight decay  & 0.00279961 & 0.0273136  & 8.44e-04  & 8.44e-04  & 0.0683645 & 0.0719626 \\
			            & Dropout       & 0.274407   & 0.150787   & 0.108411  & 0.108411  & 0.310739  & 0.207184  \\
			            & Steps         & 100000     & 100000     & 100000    & 100000    & 100000    & 100000    \\
			            & Warmup steps  & 7036       & 14310      & 18592     & 18592     & 1423      & 1973      \\
			            & Scheduler     & cosine     & cosine     & cosine    & cosine    & cosine    & cosine    \\
			            & Optimizer     & AdamW      & AdamW      & AdamW     & AdamW     & AdamW     & AdamW     \\
			            & PE Type       & -          & -          & -         & -         & -         & -         \\
			            & Encoding dim. & -          & -          & -         & -         & -         & -         \\
			            & Hidden dim.   & 384        & 256        & 512       & 512       & 384       & 384       \\
			            & Layers        & 9          & 6          & 4         & 4         & 6         & 4         \\
			            & Activation    & relu       & relu       & relu      & relu      & relu      & relu      \\
			\bottomrule
		\end{tabular}%
	}
\end{table}

\begin{table}[htbp]
	\centering
	\caption{Incumbent hyperparameters (EPM:VG \solver{BreakIDKissat}). List of hyperparameters used for baseline models in SAT solving tasks. These include parameters for positional encoding (PE), the architecture (GIN), and general learning parameters.}
	\resizebox{\columnwidth}{!}{%
		\label{table:incumbents:epm_vg_solver_solver_breakidkissat}
		\begin{tabular}{llcccccc}
			\toprule
			Hyperparam. & Variable      & Gated GCN+ & GCN        & GCN+       & GIN       & GIN+     & GRASS     \\
			\midrule
			            & Learning rate & 0.036691   & 0.00379986 & 0.00260718 & 4.76e-04  & 0.002836 & 0.010885  \\
			            & Batch size    & 16         & 16         & 16         & 16        & 16       & 16        \\
			            & Weight decay  & 0.00279961 & 2.51e-07   & 9.99e-04   & 1.35e-08  & 6.49e-04 & 0.0719626 \\
			            & Dropout       & 0.274407   & 0.434063   & 0.330087   & 0.0812465 & 0.271249 & 0.207184  \\
			            & Steps         & 100000     & 100000     & 100000     & 100000    & 100000   & 100000    \\
			            & Warmup steps  & 7036       & 12042      & 16037      & 11146     & 9352     & 1973      \\
			            & Scheduler     & cosine     & cosine     & cosine     & cosine    & cosine   & cosine    \\
			            & Optimizer     & AdamW      & AdamW      & AdamW      & AdamW     & AdamW    & AdamW     \\
			            & PE Type       & -          & -          & -          & -         & -        & -         \\
			            & Encoding dim. & -          & -          & -          & -         & -        & -         \\
			            & Hidden dim.   & 384        & 256        & 512        & 384       & 256      & 384       \\
			            & Layers        & 9          & 6          & 8          & 7         & 7        & 4         \\
			            & Activation    & relu       & relu       & relu       & relu      & relu     & relu      \\
			\bottomrule
		\end{tabular}%
	}
\end{table}

\begin{table}[htbp]
	\centering
	\caption{Incumbent hyperparameters (EPM:VG \solver{Kissat}). List of hyperparameters used for baseline models in SAT solving tasks. These include parameters for positional encoding (PE), the architecture (GIN), and general learning parameters.}
	\resizebox{\columnwidth}{!}{%
		\label{table:incumbents:epm_vg_solver_solver_kissat}
		\begin{tabular}{llcccccc}
			\toprule
			Hyperparam. & Variable      & Gated GCN+ & GCN        & GCN+      & GIN       & GIN+      & GRASS     \\
			\midrule
			            & Learning rate & 0.036691   & 0.00260718 & 4.76e-04  & 0.0029426 & 0.0369504 & 0.010885  \\
			            & Batch size    & 16         & 16         & 16        & 16        & 16        & 16        \\
			            & Weight decay  & 0.00279961 & 9.99e-04   & 1.35e-08  & 8.44e-04  & 0.0683645 & 0.0719626 \\
			            & Dropout       & 0.274407   & 0.330087   & 0.0812465 & 0.108411  & 0.310739  & 0.207184  \\
			            & Steps         & 100000     & 100000     & 100000    & 100000    & 100000    & 100000    \\
			            & Warmup steps  & 7036       & 16037      & 11146     & 18592     & 1423      & 1973      \\
			            & Scheduler     & cosine     & cosine     & cosine    & cosine    & cosine    & cosine    \\
			            & Optimizer     & AdamW      & AdamW      & AdamW     & AdamW     & AdamW     & AdamW     \\
			            & PE Type       & -          & -          & -         & -         & -         & -         \\
			            & Encoding dim. & -          & -          & -         & -         & -         & -         \\
			            & Hidden dim.   & 384        & 512        & 384       & 512       & 384       & 384       \\
			            & Layers        & 9          & 8          & 7         & 4         & 6         & 4         \\
			            & Activation    & relu       & relu       & relu      & relu      & relu      & relu      \\
			\bottomrule
		\end{tabular}%
	}
\end{table}

\begin{table}[htbp]
	\centering
	\caption{Incumbent hyperparameters (EPM:VG \solver{KissatMABDC}). List of hyperparameters used for baseline models in SAT solving tasks. These include parameters for positional encoding (PE), the architecture (GIN), and general learning parameters.}
	\resizebox{\columnwidth}{!}{%
		\label{table:incumbents:epm_vg_solver_solver_kissatmabdc}
		\begin{tabular}{llcccccc}
			\toprule
			Hyperparam. & Variable      & Gated GCN+ & GCN      & GCN+       & GIN        & GIN+      & GRASS      \\
			\midrule
			            & Learning rate & 0.036691   & 4.59e-04 & 0.00260718 & 0.00183706 & 0.0369504 & 0.00379986 \\
			            & Batch size    & 16         & 16       & 16         & 16         & 16        & 16         \\
			            & Weight decay  & 0.00279961 & 3.23e-05 & 9.99e-04   & 0.0515994  & 0.0683645 & 2.51e-07   \\
			            & Dropout       & 0.274407   & 0.144238 & 0.330087   & 0.0716766  & 0.310739  & 0.434063   \\
			            & Steps         & 100000     & 100000   & 100000     & 100000     & 100000    & 100000     \\
			            & Warmup steps  & 7036       & 2134     & 16037      & 9622       & 1423      & 12042      \\
			            & Scheduler     & cosine     & cosine   & cosine     & cosine     & cosine    & cosine     \\
			            & Optimizer     & AdamW      & AdamW    & AdamW      & AdamW      & AdamW     & AdamW      \\
			            & PE Type       & -          & -        & -          & -          & -         & -          \\
			            & Encoding dim. & -          & -        & -          & -          & -         & -          \\
			            & Hidden dim.   & 384        & 256      & 512        & 512        & 384       & 256        \\
			            & Layers        & 9          & 7        & 8          & 6          & 6         & 6          \\
			            & Activation    & relu       & relu     & relu       & relu       & relu      & relu       \\
			\bottomrule
		\end{tabular}%
	}
\end{table}

\begin{table}[htbp]
	\centering
	\caption{List of hyperparameters used for baseline models in algorithmic reasoning tasks. These include parameters for positional encoding (PE), architecture (GT/GIN), and general learning. The respective abbreviations denote the following tasks: topological sorting (TO), minimum spanning tree (MST), bridges (BR), and Steiner tree (ST).}

	\resizebox{\columnwidth}{!}{%
		\label{table:AlgoResHyperparameters1}
		\begin{tabular}{llccccccccc}
			\toprule
			\textbf{Hyperparam.} & \textbf{Variable}  & $\text{GIN}_{TO}$                            & $\text{GT}_{TO}$ & $\text{GIN}_{MST}$ & $\text{GT}_{MST}$ & $\text{GIN}_{BR}$ & $\text{GT}_{BR}$ & $\text{GIN}_{ST}$ & $\text{GT}_{ST}$ \\

			\midrule
			\multirow{5}{*}{General}
			                     & Learning rate      & 1e-4                                         & 1e-4             & 3e-4               & 3e-4              & 2e-4              & 2e-4             & 2e-3              & 1e-4             \\
			                     & Batch size         & 256                                          & 256              & 256                & 256               & 256               & 256              & 256               & 256              \\
			                     & Weight decay       & 0.1                                          & 0.1              & 0.1                & 0.1               & 0.1               & 0.1              & 0.1               & 0.1              \\
			                     & Dropout            & 0.1                                          & 0.1              & 0.1                & 0.1               & 0.1               & 0.1              & 0.1               & 0.1              \\
			                     & Steps              & 11\,880                                      & 11\,880          & 11\,880            & 11\,880           & 11\,880           & 11\,880          & 11\,880           & 11\,880          \\
			\midrule
			\multirow{4}{*}{Optimizing}
			                     & Warmup steps       & 1\,000                                       & 1\,000           & 1\,000             & 1\,000            & 1\,000            & 1\,000           & 1\,000            & 1\,000           \\
			                     & Scheduler          & \multicolumn{8}{c}{CosineAnnealingLR}                                                                                                                                                  \\
\bottomrule
		\end{tabular}}
\end{table}

\begin{table}[htbp]
	\centering
	\caption{Hyperparameters for GNNPlus baseline models across algorithmic reasoning tasks: topological sorting (TO) and Steiner tree (ST). The model variants are defined by their GNNPlus layer type.}

	\resizebox{\columnwidth}{!}{%
		\label{table:AlgoResHyperparameters_Updated}
		\begin{tabular}{llccccccccc}
			\toprule
			\textbf{Hyperparam.} & \textbf{Variable} & $\text{GIN}_{TO}$                             & $\text{GatedGCN}_{TO}$ & $\text{GCN}_{TO}$ & $\text{GIN}_{ST}$ & $\text{GatedGCN}_{ST}$ & $\text{GCN}_{ST}$ \\

			\midrule
			\multirow{5}{*}{General}
			                     & Learning rate     & 2e-3                                          & 2e-3                   & 2e-3              & 1e-3              & 2e-3                   & 2e-3              \\
			                     & Batch size        & 256                                           & 256                    & 256               & 256               & 256                    & 256               \\
			                     & Weight decay      & 0.1                                           & 0.1                    & 0.1               & 0.1               & 0.1                    & 0.1               \\
			                     & Dropout           & 0.1                                           & 0.1                    & 0.1               & 0.1               & 0.1                    & 0.1               \\
			                     & Steps             & 11\,880                                       & 11\,880                & 11\,880           & 11\,880           & 11\,880                & 11\,880           \\
			\midrule
			\multirow{4}{*}{Optimizing}
			                     & Warmup steps      & 1\,000                                        & 1\,000                 & 1\,000            & 1\,000            & 1\,000                 & 1\,000            \\
			                     & Scheduler         & \multicolumn{6}{c}{CosineAnnealingLR}                                                                                                                       \\
\bottomrule
		\end{tabular}}
\end{table}

\begin{table}[htbp]
	\centering
	\caption{Hyperparameters for GNNPlus baseline models across algorithmic reasoning tasks: bridges (BR) and minimum spanning tree (MST). The model variants are defined by their GNNPlus layer type.}

	\resizebox{\columnwidth}{!}{%
		\label{table:AlgoResHyperparameters_Updated}
		\begin{tabular}{llccccccccc}
			\toprule
			\textbf{Hyperparam.} & \textbf{Variable} & $\text{GIN}_{BR}$                             & $\text{GatedGCN}_{BR}$ & $\text{GCN}_{BR}$ & $\text{GIN}_{MST}$ & $\text{GatedGCN}_{MST}$ & $\text{GCN}_{MST}$ \\

			\midrule
			\multirow{5}{*}{General}
			                     & Learning rate     & 1e-3                                          & 1e-3                   & 1e-3              & 2e-3               & 2e-3                    & 2e-3               \\
			                     & Batch size        & 256                                           & 256                    & 256               & 256                & 256                     & 256                \\
			                     & Weight decay      & 0.1                                           & 0.1                    & 0.1               & 0.1                & 0.1                     & 0.1                \\
			                     & Dropout           & 0.1                                           & 0.1                    & 0.1               & 0.1                & 0.1                     & 0.1                \\
			                     & Steps             & 11\,880                                       & 11\,880                & 11\,880           & 11\,880            & 11\,880                 & 11\,880            \\
			\midrule
			\multirow{4}{*}{Optimizing}
			                     & Warmup steps      & 1\,000                                        & 1\,000                 & 1\,000            & 1\,000             & 1\,000                  & 1\,000             \\
			                     & Scheduler         & \multicolumn{6}{c}{CosineAnnealingLR}                                                                                                                          \\
\bottomrule
		\end{tabular}}
\end{table}

\begin{table}[htbp]
	\centering
	\caption{Hyperparameters for GNNPlus baseline models across algorithmic reasoning tasks: flow (FL), maximum matching (MM) and max clique (MC). The model variants are defined by their GNNPlus layer type.}

	\resizebox{\columnwidth}{!}{%
		\label{table:AlgoResHyperparameters_Updated}
		\begin{tabular}{llcccccccccccc}
			\toprule
			\textbf{Hyperparam.} & \textbf{Variable} & $\text{GIN}_{FL}$                             & $\text{GatedGCN}_{FL}$ & $\text{GCN}_{FL}$ & $\text{GIN}_{MM}$ & $\text{GatedGCN}_{MM}$ & $\text{GCN}_{MM}$ & $\text{GIN}_{MC}$ & $\text{GatedGCN}_{MC}$ & $\text{GCN}_{MC}$ \\

			\midrule
			\multirow{5}{*}{General}
			                     & Learning rate     & 2e-3                                          & 2e-3                   & 2e-3              & 1e-3              & 2e-3                   & 1e-3              & 1e-3              & 1e-3                   & 1e-3              \\
			                     & Batch size        & 256                                           & 256                    & 256               & 256               & 256                    & 256               & 256               & 256                    & 256               \\
			                     & Weight decay      & 0.1                                           & 0.1                    & 0.1               & 0.1               & 0.1                    & 0.1               & 0.1               & 0.1                    & 0.1               \\
			                     & Dropout           & 0.1                                           & 0.1                    & 0.1               & 0.1               & 0.1                    & 0.1               & 0.1               & 0.1                    & 0.1               \\
			                     & Steps             & 11\,880                                       & 11\,880                & 11\,880           & 11\,880           & 11\,880                & 11\,880           & 11\,880           & 11\,880                & 11\,880           \\
\bottomrule
		\end{tabular}}
\end{table}

\begin{table}[htbp]
	\centering
	\caption{List of hyperparameters used for baseline models in algorithmic reasoning and weather forecasting tasks. These include parameters for positional encoding (PE), architecture (GT/GIN), and general learning. The following tasks are denoted by the respective abbreviations: max clique (MC), flow (FL), maximum matching (MM), and weather forecasting (WE).}

	\resizebox{\columnwidth}{!}{%
		\label{table:AlgoResHyperparameters2}
		\begin{tabular}{llcccccccc}
			\toprule
			\textbf{Hyperparam.} & \textbf{Variable}  & $\text{GIN}_{MC}$                           & $\text{GT}_{MC}$ & $\text{GIN}_{FL}$ & $\text{GT}_{FL}$ & $\text{GIN}_{MM}$ & $\text{GT}_{MM}$ & $\text{GT}_{WE}$ \\

			\midrule
			\multirow{5}{*}{General}
			                     & Learning rate      & 1e-4                                        & 1e-4             & 1e-4              & 1e-4             & 1e-4              & 1e-4             & 1e-4             \\
			                     & Batch size         & 256                                         & 256              & 256               & 256              & 256               & 256              & 8                \\
			                     & Weight decay       & 0.1                                         & 0.1              & 0.1               & 0.1              & 0.1               & 0.1              & 0.1              \\
			                     & Dropout            & 0.1                                         & 0.1              & 0.1               & 0.1              & 0.1               & 0.1              & 0.1              \\
			                     & Steps              & 11\,880                                     & 11\,880          & 11\,880           & 11\,880          & 11\,880           & 11\,880          & 400\,000         \\
			\midrule
			\multirow{4}{*}{Optimizing}
			                     & Warmup steps       & 1\,000                                      & 1\,000           & 1\,000            & 1\,000           & 1\,000            & 1\,000           & 20000            \\
			                     & Scheduler          & \multicolumn{7}{c}{CosineAnnealingLR}                                                                                                                           \\
\bottomrule
		\end{tabular}}
\end{table}

\begin{table}[htbp]
	\centering
	\caption{List of hyperparameters used for baseline models in CO and electronic circuits (EC) tasks. These include parameters for positional encoding (PE), architecture (GT/GIN), and general learning.}

	\label{table:co_hyperparms}
	\begin{tabular}{llcccccc}
		\toprule
		\textbf{Hyperparam.} & \textbf{Variable}  & $\text{GIN}_{CO}$ & $\text{GT}_{CO}$ & $\text{MLP}$ & $\text{DeepSet}$ & $\text{GT}_{EC}$ & $\text{GIN}_{EC}$ \\

		\midrule
		\multirow{6}{*}{General}
		                     & Learning rate      & 1e-3              & 1e-3             & 1e-3         & 1e-3             & 5e-4             & 5e-4              \\
		                     & Batch size         & 256               & 256              & 256          & 256              & 128              & 128               \\
		                     & Weight decay       & 0.                & 0.               & 0.           & 0.               & 0.1              & 0.1               \\
		                     & Dropout            & 0.                & 0.               & 0.           & 0.               & 0.1              & 0.1               \\
		                     & Epochs             & 1\,000            & 1\,000           & 1\,000       & 1\,000           & -                & -                 \\
		                     & Steps              & -                 & -                & -            & -                & 20\,000          & 20\,000           \\
		                     & Patience           & 100               & 100              & 100          & 100              & -                & -                 \\
\bottomrule
	\end{tabular}
\end{table}
\begin{table}[htbp]
	\centering
	\caption{Hyperparameters for GNNPlus baseline models (GINE and GCN) for CO datasets.}
	\label{table:GNN_Hyperparameters_New}
	\begin{tabular}{llcc}
		\toprule
		\textbf{Hyperparam.} & \textbf{Variable} & GIN+                                   & GCN+   \\
		\midrule
		\multirow{5}{*}{General}
		                     & Learning rate     & 1e-3                                   & 1e-3   \\
		                     & Batch size        & 32                                     & 32     \\
		                     & Weight decay      & 1e-5                                   & 1e-5   \\
		                     & Dropout           & 0.0                                    & 0.05   \\
		                     & Max Epochs        & 1\,000                                 & 1\,000 \\
		\midrule
		\multirow{4}{*}{Optimizing}
		                     & Warmup epochs     & 50                                     & 50     \\
		                     & Scheduler         & \multicolumn{2}{c}{Cosine with Warmup}          \\
\bottomrule
	\end{tabular}
\end{table}
\begin{table}[htbp]
	\centering
	\caption{Hyperparameters for GNNPlus baseline models (GatedGCN, GCN, GIN) applied to Electronic Circuit tasks (Efficiency and Output Voltage).}

	\resizebox{\columnwidth}{!}{%
		\label{table:ElectronicCircuitsHyperparameters}
		\begin{tabular}{llcccccccc}
			\toprule
			                     &                   & \multicolumn{2}{c}{\textbf{GatedGCN}}         & \multicolumn{2}{c}{\textbf{GCN}} & \multicolumn{2}{c}{\textbf{GIN}}                               \\
			\cmidrule(lr){3-4} \cmidrule(lr){5-6} \cmidrule(lr){7-8}
			\textbf{Hyperparam.} & \textbf{Variable} & EFF                                           & VOUT                             & EFF                              & VOUT    & EFF     & VOUT    \\

			\midrule
			\multirow{5}{*}{General}
			                     & Learning rate     & 1e-3                                          & 2e-3                             & 1e-3                             & 2e-3    & 1e-3    & 2e-3    \\
			                     & Batch size        & 128                                           & 128                              & 128                              & 128     & 128     & 128     \\
			                     & Weight decay      & 0.1                                           & 0.1                              & 0.1                              & 0.1     & 0.1     & 0.1     \\
			                     & Dropout           & 0.1                                           & 0.1                              & 0.1                              & 0.1     & 0.1     & 0.1     \\
			                     & Steps             & 20\,000                                       & 20\,000                          & 20\,000                          & 20\,000 & 20\,000 & 20\,000 \\
			\midrule
			\multirow{4}{*}{Optimizing}
			                     & Warmup steps      & 100                                           & 100                              & 100                              & 100     & 100     & 100     \\
\bottomrule
		\end{tabular}}
\end{table}

\subsection{Baseline Architectures}
In the following, we provide implementation details on the baselines used for the datasets in \gb{}. For dataset-specific design choices, we provide detailed information in \Cref{appendix:datasetsarchchoices}.

\paragraph{Graph transformer architecture}
As described in \Cref{Section:ExpSetup} we use an encoder-processor-decoder baseline across tasks. In this case, we consider a graph transformer as the processor architecture following the implementation outlined in \citet{Spe+2025} and \citet{stoll2025generalizable}.
For most tasks, we use node-level tokenization, where each graph node is treated as a single token input to the graph transformer. However, for edge-level tasks, we use the transformation outlined for algorithmic reasoning tasks, allowing edge-level tokens to be used without changes to the processor architecture.
Additionally, absolute PEs, such as RWSE or LPE, are added to the node embeddings before the first graph transformer layer.
Then, the graph transformer layer computes full multi-head scaled-dot-product attention, adding an attention bias $\vec{B}$ to the unnormalized attention matrix and applying softmax to it. Let $\vec{Q}, \vec{K}, \vec{V} \in \mathbb{R}^{L \times d}$ and $\vec{B} \in \mathbb{R}^{L \times L}$ with $L$ denoting the number of tokens and $d$ the embedding dimension. Then attention and a graph transformer layer take the form
\begin{equation*}
	\textsf{Attention}(\vec{Q}, \vec{K}, \vec{V}, \vec{B}) \coloneqq \textsf{softmax}\big(d^{-\frac{1}{2}} \cdot \vec{Q}\vec{K}^T + \vec{B} \big) \vec{V},
\end{equation*}
\begin{equation*}
	\vec{X}^{t+1} \coloneqq \mathsf{MLP}\big(\textsf{Attention}(\vec{X}^t\vec{W}_Q, \vec{X}^t\vec{W}_K, \vec{X}^t\vec{W}_V, \vec{B})\big),
\end{equation*}
where $\vec{W}_Q, \vec{W}_K, \vec{W}_V \in \mathbb{R}^{d \times d}$ are learnable weight matrices and $\mathsf{MLP}$ denotes a two layer-MLP.  In practice, we compute attention over multiple heads, allowing for different attention biases to be added to the attention matrix. With $\phi$ given by the multihead attention computation, a processor layer is provided by \Cref{eq:processor_eq}.
In the processor, multiple layers are stacked, enabling the pipeline shown in \Cref{Section:ExpSetup}.

\paragraph{GINE architecture}~
Throughout this work, we use a GINE-based GNN baseline as the processor within the framework outlined in \Cref{Section:ExpSetup}. Following, \citet{HuStrategiesPretraining} the GINE layer updates the node representations $h_v^{(t)}$ at iteration $t$ as follows:
\begin{equation*}
	\boldsymbol{h}_v^{(t+1)} = \mathsf{N}\Big((1+\epsilon) \boldsymbol{h}_v^{(t)} + \sum_{u \in \mathcal{N}(v)}\mathsf{ReLU}(\boldsymbol{h}_v^{(t)} + e_{uv})\Big)
\end{equation*}
where $\mathcal{N}(v)$ denotes the neighborhood of a node $v$ and $\mathsf{N}$ a neural network such as an MLP.

We apply the GINE processor layer in a similar way to the graph transformer baseline design outlined previously by replacing $\phi$ with a GINE layer where $\mathsf{N}$ is given by a dropout layer.
First, the node embeddings, optionally with added PEs, are passed to the GINE layer, where layer normalization is applied. Then, the output of the GINE message-passing layer is forwarded to a two-layer MLP. We use the same residual connection as seen in the GIN implementation from \citet{Spe+2025}.
We then stack multiple layers to form the processor component of our baseline architecture.
Unless otherwise specified, we use mean pooling for graph-level tasks at the end of the processor step.

\paragraph{GNN+ Baseline methods}
In addition to the graph transformer and GINE baseline methods, we use current state-of-the-art GNN methods originally proposed by \citet{} as an extension to existing GNN baselines. Following the approach from their paper, we include the GIN+, GatedGCN+, and GCN+ baselines throughout our experiments. Specifically, layer updates are given by the following, with node representations $h_v^{(l)}$ updated. \\
GIN+:
\begin{equation*}
	\boldsymbol{h}_v^l = \text{FFN}(\text{Dropout}(\sigma(\text{BN}( \text{MLP}^l((1 + \epsilon) \cdot \boldsymbol{h}_v^{l-1} + \sum_{u \in \mathcal{N}(v)} \boldsymbol{h}_u^{l-1}) + \boldsymbol{e}_{uv} \boldsymbol{W}_e^l)))) + \boldsymbol{h}_v^{l-1})
\end{equation*} \\
GatedGCN+:
\begin{equation*}
	\boldsymbol{h}_v^l = \text{FFN}(\text{Dropout}(\sigma(\text{BN}(\sum_{u \in \mathcal{N}(v) \cup \{v\}} \frac{1}{\sqrt{\hat{d}_u \hat{d}_v}} \boldsymbol{h}_u^{l-1} \boldsymbol{W}^l + \boldsymbol{e}_{uv} \boldsymbol{W}_e^l)))) + \boldsymbol{h}_v^{l-1})
\end{equation*} \\
GCN+
\begin{equation*}
	\boldsymbol{h}_v^l = \text{FFN}(\text{Dropout}(\sigma(\text{BN}( \boldsymbol{h}_v^{l-1}\boldsymbol{W}_1^l + \sum_{u \in \mathcal{N}(v)} \eta_{v,u} \odot \boldsymbol{h}_u^{l-1} \boldsymbol{W}_2^l + \boldsymbol{e}_{uv} \boldsymbol{W}_e^l)))) + \boldsymbol{h}_v^{l-1})
\end{equation*}

where $\sigma$ denotes an activation function, FFN and MLP denote a fully connected linear layer and an $l$ layer MLP, respectively. Furthermore, $\eta= \sigma(\boldsymbol{h}_{u}^{l-1}\boldsymbol{W}_3 + \boldsymbol{h}_{v}^{l-1}\boldsymbol{W}_4)$, $\hat{d}_u$ denotes the degree of node $u$ and $\boldsymbol{W}$ denote weight matrices. Edge features are included via $e_{uv}$
\subsection{Architecture choices for datasets}\label{appendix:datasetsarchchoices}

\paragraph{Social Networks}~
On the proposed BlueSky datasets, we opted for a different GNN architecture. The high(er) number of nodes, along with high average node degree and variance, pushed us towards resorting to a message-passing architecture aggregating messages over neighborhoods via averaging (instead of summation as in GIN). We opted, in particular, for a variant of GraphConv with mean aggregation. Node representations $h_v^{(t)}$ are, namely, updated as follows:
\begin{equation}
	h_v^{(t+1)} = \sigma \Big( W^{(t)}_1 h_v^{(t)} + \frac{1}{\text{deg}_{in}(v)} \sum_{u \colon \ u \to v}  W^{(t)}_2 h_u^{(t+1)} \Big),
\end{equation}
\noindent where $\sigma$ is set to ReLU, dropout is applied before its application, and, finally, no normalization layers are interleaved.

\paragraph{Weather forecasting}~

For the proposed weather forecasting dataset, we provide a hybrid GNN+Graph Transformer architecture closely based on Graphcast \cite{lam2023learning}. Since the data contains both grid and mesh structures, our architecture contains the following modules. A GNN-based encoder, a Graph Transformer processor, and a GNN decoder, along with a prediction head.
The encoder consists of a single GNN layer and maps the grid points onto the mesh. It receives in addition to the node features (the current and previous time step, with associated variables) the spherical coordinates of the grid node and corresponding mesh nodes. Furthermore, edge features are given by the relative 3d positions of mesh and grid points in the global mapping as well as an L2 normalized distance. The encoder then computes the mesh features.

Let $\mathcal{G} = (\mathcal{V}_g, \mathcal{V}_m, \mathcal{E}_{g2m}, \mathcal{E}_{m}, \mathcal{E}_{m2g})$
be the heterogeneous graph, where $\mathcal{V}_g$ are grid nodes (lat-lon),
$\mathcal{V}_m$ are mesh nodes, $\mathcal{E}_{g2m}$ grid-to-mesh edges,
$\mathcal{E}_{m}$ mesh-internal edges, and $\mathcal{E}_{m2g}$ mesh-to-grid edges.
Each grid node $i \in \mathcal{V}_g$ carries state features
$\mathbf{u}_i^{(t)} \in \mathbb{R}^{F}$ at timestep $t$
and spatial coordinates $(\phi_i, \theta_i)$ (longitude, latitude).

The raw two-timestep grid state is stacked and normalized channel-wise:
\[
	\tilde{\mathbf{u}}_i
	= \frac{\left[\mathbf{u}_i^{(t)},\, \mathbf{u}_i^{(t-1)}\right] - \boldsymbol{\mu}}
	{\boldsymbol{\sigma}}
	\;\in\; \mathbb{R}^{2F},
\]
where $\boldsymbol{\mu}, \boldsymbol{\sigma} \in \mathbb{R}^{2F}$ are pre-computed
training-set statistics (tiled for both timesteps), and NaN/Inf entries are replaced by 0.

Further, we compute the spherical to Cartesian coordinate transformation:
For a node at latitude $\lambda$ and longitude $\ell$ (in degrees):
\[
	\phi = \frac{\pi}{180}\ell, \qquad \theta = \frac{\pi}{180}(90 - \lambda), \qquad
	\mathbf{p} = \begin{pmatrix} \cos\phi\sin\theta \\ \sin\phi\sin\theta \\ \cos\theta \end{pmatrix}
	\in \mathbb{R}^3.
\]

Afterward, we can compute edge features used in the encoder and decoder:
\[
	\mathbf{s}_i^{\text{grid}},\mathbf{s}_i^{\text{mesh}}  = \bigl[\cos\theta_i,\; \cos\phi_i,\; \sin\phi_i\bigr]
	\;\in\; \mathbb{R}^3.
\]

For an edge $e_{ij}$ the relative displacement and L2 distance in 3D are:
\[
	\Delta\mathbf{p}_{ij} = \mathbf{p}_j - \mathbf{p}_i, \qquad
	d_{ij} = \|\Delta\mathbf{p}_{ij}\|_2,
\]
normalized by the maximum distance $d_{\max}$ of same type edges:
\[
	\mathbf{s}_{ij}^{\text{edge}}
	= \left[\frac{d_{ij}}{d_{\max}},\;
		\frac{\Delta\mathbf{p}_{ij}}{d_{\max}}\right]
	\;\in\; \mathbb{R}^{4}.
\]
We now compute the initial grid and mesh features:
\[
	\mathbf{h}_i^g = W_g\,\bigl[\tilde{\mathbf{u}}_i \;\|\; \mathbf{s}_i^{\text{grid}}\bigr]
	\;\in\; \mathbb{R}^{D}, \qquad i \in \mathcal{V}_g,
\]
\[
	\mathbf{h}_j^m = W_m\,\mathbf{s}_j^{\text{mesh}}
	\;\in\; \mathbb{R}^{D}, \qquad j \in \mathcal{V}_m,
\]
where $W_g \in \mathbb{R}^{D \times (2F+3)}$ and $W_m \in \mathbb{R}^{D \times 3}$ are linear embeddings.

Each grid-to-mesh edge $(i,j) \in \mathcal{E}_{g2m}$ is embedded via a MLP layer and then processed using a neighborhood aggregating GNN:
\[
	\mathbf{e}_{ij}^{g2m} = W_{g2m}\,\mathbf{s}_{ij}^{\text{edge}} \;\in\; \mathbb{R}^{D},
\]
\[
	\tilde{\mathbf{e}}_{ij}^{g2m}
	= \mathrm{MLP}_{g2m}\!\left(\mathbf{e}_{ij}^{g2m} \;\|\; \mathbf{h}_i^g \;\|\; \mathbf{h}_j^m\right)
	\;\in\; \mathbb{R}^{D},
\]
with residual: $\hat{\mathbf{e}}_{ij}^{g2m} = \mathbf{e}_{ij}^{g2m} + \tilde{\mathbf{e}}_{ij}^{g2m}$.

\[
	\mathbf{a}_j^m = \sum_{i:\,(i,j)\in\mathcal{E}_{g2m}} \tilde{\mathbf{e}}_{ij}^{g2m},
\]
\[
	\tilde{\mathbf{h}}_j^m
	= \mathrm{MLP}_{m}\!\left(\mathbf{h}_j^m \;\|\; \mathbf{a}_j^m\right),
\]
\[
	\hat{\mathbf{h}}_j^m = \mathbf{h}_j^m + \tilde{\mathbf{h}}_j^m.
\]

All encoder/decoder MLPs use the SiLU activation:
\[
	\mathrm{MLP}(\mathbf{x})
	= W_2\,\mathrm{SiLU}(W_1\mathbf{x} + \mathbf{b}_1) + \mathbf{b}_2,
	\quad \mathrm{SiLU}(z) = z \cdot \sigma(z).
\]

In a second step, the processor then computes the node updates for mesh nodes only. For this, the same graph transformer from \citet{stoll2025generalizable} is used as in previous tasks.

Mesh-internal edge features are embedded and projected to per-head attention biases:
\[
	\mathbf{b}_{jk}^{\text{mesh}}
	= W_{\text{enc}}\Bigl(W_{\text{mesh}}\,\mathbf{s}_{jk}^{\text{mesh-edge}}\Bigr)
	\;\in\; \mathbb{R}^{H}, \quad (j,k) \in \mathcal{E}_m,
\]
where $H$ is the number of attention heads. These biases are added to the attention logits.
The processor runs $L$ standard Transformer encoder layers:
\[
	\mathbf{Z}^{(\ell)} = \mathrm{TransformerLayer}_\ell\!\left(\mathbf{Z}^{(\ell-1)},\, A^{\text{mesh}}\right),
	\quad \ell = 1,\ldots,L,
\]
where $A^{\text{mesh}} \in \mathbb{R}^{(N_m+1)\times(N_m+1)\times H}$ contains the
edge-derived attention biases ($-\infty$ for non-edges). The output mesh embeddings
are $\bar{\mathbf{h}}_j^m = \mathbf{Z}^{(L)}_{j+1}$.

As a final step, the decoder maps mesh features back to the underlying grid. Afterward, computed predictions are denormalized, and the residual between time steps $t$ and $t+2$ is computed as the result using an MLP prediction head.

\[
	\tilde{\mathbf{e}}_{ji}^{m2g}
	= \mathrm{MLP}_{m2g}\!\left(\mathbf{e}_{ji}^{m2g} \;\|\; \bar{\mathbf{h}}_j^m \;\|\; \mathbf{h}_i^g\right)
	\;\in\; \mathbb{R}^{D},
\]
where $\mathbf{e}_{ji}^{m2g} = W_{m2g}\,\mathbf{s}_{ji}^{\text{edge}}$.

\[
	\mathbf{a}_i^g = \sum_{j:\,(j,i)\in\mathcal{E}_{m2g}} \tilde{\mathbf{e}}_{ji}^{m2g}.
\]

\[
	\hat{\boldsymbol{\delta}}_i
	= \mathrm{MLP}_{\text{out}}\!\left(\mathbf{h}_i^g \;\|\; \mathbf{a}_i^g\right)
	\;\in\; \mathbb{R}^F.
\]

\[
	\boldsymbol{\delta}_i = \hat{\boldsymbol{\delta}}_i \odot \boldsymbol{\sigma}_{\Delta},
\]
\[
	\hat{\mathbf{u}}_i^{(t+2)} = \mathbf{u}_i^{(t)} + \boldsymbol{\delta}_i,
\]
where $\boldsymbol{\sigma}_{\Delta} \in \mathbb{R}^F$ is the pre-computed standard deviation
of the per-variable one-step differences (training set) and $\odot$ denotes the elementwise product.

\subsection{Dataset preprocessing}\label{app:data_preproc}
To align the proposed datasets with baseline methods, we applied the following preprocessing steps to the original datasets.

\paragraph{Algorithmic reasoning}~
For edge-level tasks requiring edge classification on a graph, we transform the problem statement as follows to transform the task into a node-level task:
We leverage the following edge transform which allows us to provide node level tokenization/targets on a modified graph $G'$ obtained from the original graph $G$, with $V(G') \coloneqq \{(v, v) \mid v \in V(G)\} \cup E(G)$ and $E(G') \coloneqq \{ ((u, v), (w, z)) \mid u = w \vee u = z \vee v = w \vee v = z\}$. We use this transform for all baselines evaluated for algorithmic reasoning datasets.

Furthermore, since networkX outputs only directed edges for each of the edge level tasks we further transform the target tensor $y$ as follows:
We assign the same label to the reverse edge of each outgoing edge. This is done by comparing existing edges and the target tensor.
Therefore, we transform the directed target into an undirected one.

\paragraph{Weather forecasting}~

Since features are provided as unnormalized values, we follow the preprocessing pipeline from \citet{lam2023learning} and \citet{price2023gencast}.
We outline the process of preprocessing for weather forecasting information in the following:

First, the raw ERA5 data, provided as a .zarr file, is loaded, and the empty mesh is created once. Mesh creation depends on the number of refinement levels selected; for our case, we set it to four. Furthermore, grid points are connected to corresponding nearby mesh points, thereby creating node and edge indices for the graphs outlined in \autoref{appendix:datasetsarchchoices}. Since these are the same across all time steps, they are saved as static components to reduce storage requirements.

Afterward, the node features are attached to the grid nodes, and variable count information is saved for later use in the loss function. This constitutes the raw data included in our dataset.

To reduce local preprocessing time, we also provide the dataset used in our evaluation. For this, the following additional steps have been applied.

Each timestep receives the previous timestep as additional node features, including all weather variables but not location information. Furthermore, static information is saved separately, and mesh node features are initialized with zero vectors.

During training, the features are additionally normalized and enhanced with distance information, as outlined in \autoref{appendix:datasetsarchchoices}. All additional information is saved as metadata in the dataset or provided as additional files.

Detailed descriptions of each feature and the preprocessing pipeline can be found in \cite{lam2023learning}.

\section{Automated hyperparameter optimization}
\label{sec:hpo}

\gb{} also integrates automated hyperparameter optimization (HPO), which aims to make it easier to tune new models, improve performance, and enhance reproducibility. Deep graph learning architectures are notoriously sensitive to their hyperparameter configurations, ranging from learning rates and weight decay to structural choices such as the number of layers and hidden dimensions. To address this, the hyperparameter optimization in \gb{} is based on the \textsc{SMAC3}~\citep{LinEtAl22} package, a versatile framework that leverages Bayesian Optimization (BO) to efficiently explore the configuration space.

At its core, standard Bayesian Optimization constructs a probabilistic surrogate model that maps a hyperparameter configuration to its observed performance. The BO loop proceeds iteratively: first, an initial set of configurations is evaluated. Then, the surrogate model is fitted to this historical data. To decide which configuration to evaluate next, BO employs an acquisition function, such as Expected Improvement (EI)~\citep{Mockus2012}. The acquisition function evaluates the utility of candidate configurations by balancing \emph{exploitation} (proposing configurations in regions where the surrogate model predicts high performance) and \emph{exploration} (proposing configurations in regions where the surrogate model is highly uncertain). Once the acquisition function is maximized, the best candidate configuration is evaluated on the target task, and the resulting performance is used to update the surrogate model.
In \textsc{SMAC3}, the default surrogate model is a random forest (RF). Random Forests are particularly well-suited for HPO because they natively handle mixed continuous, categorical, and conditional hyperparameter spaces, and they can estimate predictive variance (uncertainty) empirically from the variance across the individual regression trees.

While standard BO is vastly more efficient than random search, evaluating every proposed configuration to completion (e.g., training a GNN for 100\,000 steps) remains too costly. To mitigate this, we specifically utilize the \new{multi-fidelity facade} provided by \textsc{SMAC3}. Multi-fidelity optimization accelerates the search by evaluating configurations on cheaper, approximate versions of the target task, known as \emph{fidelities}. Fidelities can be defined by the number of training steps, the number of epochs, or the size of the dataset subset. A well-known algorithm for multi-fidelity optimization is Hyperband~\citep{hyperband}. Hyperband samples configurations randomly, evaluates them on a low fidelity, and prunes unpromising configurations using a tournament-like strategy. Hyperbands repeats this process multiple times, with increasing minimum fidelities, to give slow-learning configurations a chance to show their performance.

The underlying scheduling algorithm used by the \textsc{SMAC3} multi-fidelity facade is based on BOHB~\citep{FalEtAl18}, which combines the strengths of Hyperband for early-stage elimination of unpromising configurations with those of BO for late-stage convergence toward the global optimum. In short, BOHB samples new configurations using BO and prunes unpromising configurations using a Hyperband-style strategy.

Due to limited computational resources, we were unable to run HPO on all datasets and domains. However, we conducted HPO for all SAT datasets and for all baselines. For this purpose, we used the \textsc{SMAC3} multi-fidelity facade with 60 trials per run. The chosen fidelity metric was the number of training steps: we defined the minimum budget as 1\,000 steps and the maximum budget as 100\,000 steps. This multi-fidelity approach allowed us to explore a vast configuration space---evaluating many architectures and hyperparameter combinations at 1\,000 steps---while only training the very best candidates to the full 100\,000 steps. The complete configuration space used for this process is shown in \autoref{table:hpo_cs}.

\begin{table}[htbp]
	\centering
	\caption{Configuration space used in the hyperparameter optimization.}

	\label{table:hpo_cs}
	\begin{tabular}{lc}
		\toprule
		\textbf{Hyperparameter} & \textbf{Range}    \\

		\midrule
		Learning rate           & [1e-5, 5e-2]      \\
		Weight decay            & [1e-8, 1e-1]      \\
		Warmup iters            & [1\,000, 20\,000] \\
		Dropout                 & [0.0, 0.5]        \\
		Num layers              & [4, 10]           \\
		Hidden dim              & \{256, 384, 512\} \\

		\bottomrule
	\end{tabular}
\end{table}

\end{document}